\pdfoutput=1
\documentclass[final,12pt]{colt2021} 
 
\usepackage[utf8]{inputenc} 
\usepackage[T1]{fontenc}    
\usepackage{hyperref}       
\usepackage{url}            
\usepackage{booktabs}       
\usepackage{amsfonts}       
\usepackage{nicefrac}       
\usepackage{caption}
\usepackage{lipsum} 

\usepackage{pifont}
\usepackage{microtype}
\usepackage{graphicx}
\usepackage{textcomp}
\usepackage{makecell}
\usepackage{amssymb,dsfont,empheq}
\usepackage{algorithm,algorithmicx,setspace}
\usepackage{algpseudocode}
\usepackage{syntonly,bm,wrapfig}
\usepackage{xcolor,cases,colortbl}
\usepackage{amsmath}
\usepackage{mathtools}
\usepackage[toc,page,header]{appendix}
\usepackage{minitoc}
\renewcommand \thepart{}

\definecolor{ao}{rgb}{0.0, 0.5, 0.0}

\newcommand{\cmark}{\textcolor{ao}{\ding{51}}}
\newcommand{\xmark}{\textcolor{purple}{\ding{55}}}
\input{mysymbol.sty}

\DeclareMathOperator*{\argmin}{arg\,min}

\newcommand{\EE}{\mathbb{E}}

\def \bbeta {{\boldsymbol \beta}}

\newcommand{\blue}[1]{{\color{blue}#1}}
\newcommand{\red}[1]{{\color{red}#1}}

\newcommand\numberthis{\addtocounter{equation}{1}\tag{\theequation}}

\newtheorem{theorem}{Theorem}
\newtheorem{lemma}{Lemma}

\newtheorem{corollary}{Corollary}
\newtheorem{definition}{Definition}
\newtheorem{innercustomthm}{Restatement of Theorem}
\newenvironment{customthm}[1]
  {\renewcommand\theinnercustomthm{#1}\innercustomthm}
  {\endinnercustomthm}

\newtheorem{innercustomlm}{Restatement of Lemma}
\newenvironment{customlm}[1]
  {\renewcommand\theinnercustomlm{#1}\innercustomlm}
  {\endinnercustomlm}

\definecolor{LightCyan}{rgb}{0.88,1,1} 
\definecolor{Lightpurple}{rgb}{0.9,0.9,1} 

\makeatletter
\def\thanks#1{\protected@xdef\@thanks{\@thanks
        \protect\footnotetext{#1}}}
\makeatother

\algrenewcommand{\algorithmiccomment}[1]{\hskip0em$\triangleright$ #1}
\newcommand{\FullTitle}{Alternating Implicit  Projected SGD and Its Efficient Variants for  Equality-constrained Bilevel Optimization}

\title[Alternating Implicit Projected SGD and Its Efficient Variants]{\FullTitle}

\usepackage{times}

\coltauthor{%
  \Name{Quan Xiao} \Email{xiaoq5@rpi.edu}\\
  \addr Rensselaer Polytechnic Institute\\
  \Name{Han Shen} \Email{shenh5@rpi.edu}\\
  \addr   Rensselaer Polytechnic Institute \\
  \Name{Wotao Yin} \Email{wotao.yin@alibaba-inc.com}\\
  \addr   DAMO Academy, Alibaba US \\
    \Name{Tianyi Chen} \Email{chent18@rpi.edu}\\
  \addr   Rensselaer Polytechnic Institute
 \thanks{
 The work of Q. Xiao, H. Shen and T. Chen was   supported by National
Science Foundation MoDL-SCALE project   2134168, the RPI-IBM Artificial Intelligence Research Collaboration (AIRC) and an Amazon Research Award.}
}

\begin{document}

\maketitle
\doparttoc 
\faketableofcontents 

\begin{abstract}
Stochastic bilevel optimization, which captures the inherent nested structure of machine learning problems, is gaining popularity in many recent applications. Existing works on bilevel optimization mostly consider either unconstrained problems or constrained upper-level problems. This paper considers the stochastic bilevel optimization problems with equality constraints both in the upper and lower levels. By leveraging the special structure of the equality constraints problem, the paper first presents an alternating implicit projected SGD approach and establishes the $\tilde{\cal O}(\epsilon^{-2})$ sample complexity that matches the state-of-the-art complexity of ALSET \citep{chen2021closing} for  unconstrained bilevel problems. To further save the cost of projection, the paper presents two  alternating implicit projection-efficient SGD approaches, where one algorithm enjoys the $\tilde{\cal O}(\epsilon^{-2}/T)$  upper-level and $\tilde{\cal O}(\epsilon^{-1.5}/T^{\frac{3}{4}})$ lower-level projection complexity with ${\cal O}(T)$ lower-level batch size, and the other one enjoys $\tilde{\cal O}(\epsilon^{-1.5})$ upper-level and lower-level projection complexity with ${\cal O}(1)$ batch size. Application to federated bilevel optimization has been presented to showcase the empirical performance of our algorithms. 
Our results
demonstrate that equality-constrained bilevel optimization with strongly-convex lower-level problems can be solved as efficiently as stochastic single-level optimization problems.

\end{abstract}

\section{Introduction}
Projected stochastic gradient descent (SGD) is a fundamental approach to solving large-scale constrained single-level machine learning problems. Specifically, to minimize $\EE_\xi\left[\mathcal{L}(x;\xi)\right]$ over a given convex set $\mathcal{X}$, it generates the sequence $x^{k+1}= \operatorname{Proj}_{\mathcal{X}}(x^k-\alpha\nabla \mathcal{L}(x^k;\xi^k))$, where $\alpha>0$ is the stepsize and $\nabla \mathcal{L}(x^k;\xi^k)$ is a stochastic gradient estimate of $\EE_\xi\left[\mathcal{L}(x^k;\xi)\right]$. If $\EE_\xi\left[\mathcal{L}(x;\xi)\right]$ is nonconvex, projected SGD requires a sample complexity of ${\cal O}(\epsilon^{-2})$ with ${\cal O}(1/\epsilon)$ batch size \citep{ghadimi2016mini}. The requirement of ${\cal O}(1/\epsilon)$ batch size  has been later relaxed \citep{davis2019stochastic} using the Moreau envelope technique, and its convergence rate matches that of vanilla SGD.

However, recent machine learning applications often go beyond the single-level structure, including hyperparameter optimization \citep{maclaurin2015gradient,franceschi2017forward}, meta-learning, \citep{finn2017model} reinforcement learning, \citep{sutton2018reinforcement} and neural architecture search \citep{liu2018darts}. While the nonasymptotic analysis of the  alternating implicit SGD for unconstrained bilevel optimization with strongly convex and smooth lower-level problems was well-understood \citep{ghadimi2018approximation,hong2020two,ji2021bilevel,chen2021closing,li2022fully}, to the best of our knowledge, the finite-time guarantee of  alternating implicit projected SGD on bilevel problems with both upper-level (UL) and lower-level (LL) constraints have not been investigated yet. 
In this context, a natural but important question is 
\begin{center}
\emph{Can we establish the $\tilde{\cal O}(\epsilon^{-2})$ sample complexity of  alternating implicit projected SGD for  a family of bilevel problems with both UL and LL constraints?} 
\end{center}

We give an affirmative answer to this question for the following \emph{stochastic bilevel optimization problems} with both UL and LL constraints, given by 
\begin{subequations}\label{opt0}
	\begin{align}
	&\min_{x\in \mathcal{X}}~~~F(x)\triangleq \mathbb{E}_\xi[f(x, y^*(x);\xi)] ~~~~~~~~~~~~~~~~~~~\, {\rm (upper)}\\
	&~{\rm s.  t.}~~~~~y^*(x)\triangleq \argmin_{y\in {\cal Y}(x)}~\mathbb{E}_\phi[g(x, y;\phi)] ~~~~~~~~~~~{\rm (lower)}\label{opt0-0-low}
\end{align}
\end{subequations}
where $\xi$ and $\phi$ are random variables,  $\mathcal{X}=\{x~|~Bx=e\}\subset\mathbb{R}^{d_x}$ and ${\cal Y}(x)=\{y~|~Ay+h(x)=c\}\subset\mathbb{R}^{d_y}$ are closed convex set; $A\in\mathbb{R}^{m_y\times d_y}, B\in\mathbb{R}^{m_x\times d_x},c\in\mathbb{R}^{m_y}, e\in\mathbb{R}^{m_x}, h:\mathbb{R}^{d_x}\rightarrow\mathbb{R}^{m_y}$; $A$ and $B$ are not necessarily full row or column rank and the coupling function $h$ can be nonlinear. In \eqref{opt0}, the UL optimization problem depends on the solution of the LL optimization over $y$, and both the LL function and constraint set depend on the UL variable $x$. The equality-constrained bilevel problem \eqref{opt0} covers a wider class of applications than unconstrained bilevel optimization, such as distributed bilevel optimization \citep{tarzanagh2022fednest,yang2022decentralized}, hyperparameter optimization for optimal transport \citep{luise2018differential,gould2022exploiting}, and the design of transportation networks \citep{marcotte1986network,alizadeh2013two}. When $A=0,B=0,h=0,c=0,e=0$, the problem \eqref{opt0} reduces to the unconstrained stochastic bilevel problem \citep{ghadimi2018approximation,hong2020two,ji2021bilevel,khanduri2021near,chen2021closing,chen2022single,li2022fully}. 

Generically speaking, to solve \eqref{opt0}, alternating implicit projected SGD performs
\begin{align}
    y^{k+1}=\operatorname{Proj}_{\mathcal{Y}(x^k)}\left(y^k-\beta h_g^k\right) ~~~{\rm and}~~~ x^{k+1}=\operatorname{Proj}_{\mathcal{X}}\left(x^k-\beta h_f^k\right),\label{dynamic-aipod}
\end{align}
where $h_g^k$ is an unbiased stochastic gradient estimator of $\mathbb{E}_\phi[g(x^k, y^k;\phi)]$, $h_f^k$ is a (possibly biased) stochastic gradient estimator of $F(x^k)$, and, $\alpha$ and $\beta$ are stepsizes. An immediate difficulty in analyzing the stochastic methods for \eqref{opt0} is that $h_f^k$ is usually biased due to the inaccessibility of $y^*(x)$. Moreover, the bias is roughly proportional to the LL accuracy $\|y^{k+1}-y^*(x^k)\|$, but the latter is not ensured to be small enough after finite LL steps. Therefore, if we directly apply the existing analysis for nonconvex constrained single-level problems \citep{davis2019stochastic} to even merely UL constrained bilevel problems, it either leads to a suboptimal rate \citep{hong2020two} or requires additional LL corrections \citep{chen2022single}, let alone coupled with LL constraints. Leveraging the smoothness of the projection to linear equality constraints, we establish the convergence of alternating implicit projected SGD comparable to the unconstrained case. 
 
Despite its popularity, alternating implicit projected SGD is not suitable for scenarios where evaluating projections is expensive since it calls projections at each step. For the linear equality constraints in \eqref{opt0}, although the projection onto the set of constraints has the analytical solution (See \citep{boyd2004convex} or Appendix \ref{sec:app-pre}), assessing it involves calculating the product of the projection matrix with a vector which is computationally costly when the projection matrix is high-rank. Even if the projection matrix is low-rank, it would also suffer from other bottlenecks. For example, in federated bilevel learning \citep{tarzanagh2022fednest}, projection is low-rank and simple as it amounts to averaging the gradients of all clients \citep{parikh2014proximal}. However, as the data is stored on the client side and clients are only able to communicate with the server, obtaining the projection suffers from extreme communication costs. Therefore, this inspires us to design a provable projection efficient algorithm for \eqref{opt0} beyond alternating implicit projected SGD.

\subsection{Contributions}
In this context, we consider bilevel optimization with equality constraints. We analyze the convergence rate for alternating implicit projected SGD, {\sf AiPOD} for short, propose two projection efficient variants of {\sf AiPOD}, and apply them to federated bilevel optimization. We summarize our contributions below. 
\begin{itemize}\setlength\itemsep{-0.15em}
    \item[C1)] We provide the first nonasymptotic analysis of {\sf AiPOD} for bilevel optimization with both UL and LL constraints and attain the $\tilde{\cal O}(\epsilon^{-2})$ sample and iteration complexity to achieve $\epsilon$ stationary point of \eqref{opt0}, which matches the complexity of alternating implicit SGD algorithm for the unconstrained bilevel problem \citep{chen2021closing}. 
    \item[C2)] Leveraging the recent work Proxskip \cite{mishchenko2022proxskip}, we propose two efficient variants of {\sf AiPOD} termed {\sf E-AiPOD} and {\sf E2-AiPOD} tailored to the setting where evaluating projection is costly. In the setting \eqref{opt0}, {\sf E-AiPOD} reduces the UL and LL projection complexity from $\tilde{\cal O}(\epsilon^{-2})$ to $\tilde{\cal O}(\epsilon^{-2}/T)$ and $\tilde{\cal O}(\epsilon^{-1.5}/T^{\frac{3}{4}})$ with ${\cal O}(T)$ LL batch size. In the setting \eqref{opt0} with $h(x)=0$, {\sf E2-AiPOD} reduces both the UL and LL projection complexity to $\tilde{\cal O}(\epsilon^{-1.5})$ with ${\cal O}(1)$ batch size. 
    \item[C3)] We show the implication of the proposed method in federated bilevel learning and provide improved communication complexity over the state-of-the-art work \citep{tarzanagh2022fednest}. Experiments on numerical   and federated examples are provided to verify our theoretical findings. 
\end{itemize}

\subsection{Technical challenges}
We highlight the technical challenges for the theoretical analysis. 
 \begin{itemize}\setlength\itemsep{-0.15em}
    \item[T1)] The state-of-the-art analysis of unconstrained bilevel optimization \citep{ghadimi2018approximation,hong2020two,ji2021bilevel,chen2021closing} relies on the smoothness of the implicit solution mapping $y^*(x)$. However, the well-known formula of $\nabla y^*(x)$ does not hold when LL problem has coupling constraints so that the smoothness of $y^*(x)$ is unexplored. 
    \item[T2)] The update of UL can be viewed as biased projected SGD but the bias of the gradient estimator leads to suboptimal rates in the general analysis of projected SGD \citep{davis2019stochastic} since we can not separate out a negative term to mitigate the LL bias. 
    \item[T3)] The Lyapunov function that is critical in analyzing the single-loop unconstrained bilevel optimization \citep{chen2021closing} is insufficient for the analysis of our projection-efficient variant {\sf E-AiPOD} and {\sf E2-AiPOD} due to the additional errors caused by skipping projection steps. 
    \item[T4)] Analysis for {\sf E2-AiPOD} relies on the convergence of multiple coupling sequences, but the new implicit solution mapping generated by {\sf E2-AiPOD} is not Lipschitz smooth, which prevents us from leveraging the recent advance in multi-sequence stochastic approximation \citep{shen2022single}. 
\end{itemize}
  
  \begin{table}[tbp]
\fontsize{9}{9}\selectfont
    \begin{tabular}{>{\columncolor{gray!10}}c|>{\columncolor{Lightpurple!70}}c|>{\columncolor{Lightpurple!70}} c|>{\columncolor{Lightpurple!70}}c|  c| c|c| c| c}
    \hline\hline
           &AiPOD &E-AiPOD&E2-AiPOD &BSA &stocBiO&ALSET&TTSA&IG-AL\\
        \hline
        stochasticity&\cmark&\cmark&\cmark&\cmark&\cmark&\cmark&\cmark&\xmark\\
        \hline
        UL constraint&\cmark&\cmark&\cmark&\xmark&\xmark&\xmark&\cmark&\xmark\\
        \hline
        LL constraint&\cmark&\cmark&\cmark&\xmark&\xmark&\xmark&\xmark&\cmark\\
        \hline
        LL batch size &${\cal O}(1)$&${\cal O}(T)$&${\cal O}(1)$&${\cal O}(1)$&${\cal O}(\epsilon^{-1})$&${\cal O}(1)$&${\cal O}(1)$&/\\
        \hline
        $y$-update&PSGD&Proxskip&Proxskip&SGD&SGD&SGD&SGD&ALM\\
        \hline
        $w$-update&NS&delayed NS&Proxskip&NS&CG&NS&NS&exact\\
        \hline
        sample complexity&$\tilde{\cal O}(\epsilon^{-2})$&$\tilde{\cal O}(\epsilon^{-2})$&$\tilde{\cal O}(\epsilon^{-2})$&${\cal O}(\epsilon^{-3})$&$\tilde{\cal O}(\epsilon^{-2})$&$\tilde{\cal O}(\epsilon^{-2})$&${\cal O}(\epsilon^{-2.5})$&/\\
        \hline
        UL projection&$\tilde{\cal O}(\epsilon^{-2})$&$\tilde{\cal O}(\epsilon^{-2}/T)$&$\tilde{\cal O}(\epsilon^{-1.5})$&/&/&/&${\cal O}(\epsilon^{-2.5})$&/\\
        \hline
        LL projection&$\tilde{\cal O}(\epsilon^{-2})$&$\tilde{\cal O}(\epsilon^{-1.5}/T^{\frac{3}{4}})$&${\cal O}(\epsilon^{-1.5})$&/&/&/&/&/\\
        \hline\hline
    \end{tabular}
    \caption{Sample complexity and projection complexity of our methods {\sf AiPOD}, {\sf E-AiPOD}, {\sf E2-AiPOD} and the state-of-the-art works on the bilevel problem (BSA in \citep{ghadimi2018approximation}, ALSET in \citep{chen2021closing}, stoBiO in \citep{ji2021bilevel}, TTSA in \citep{hong2020two}), IG-AL in \citep{tsaknakis2022implicit}) to achieve an $\epsilon$ stationary point, where PSGD denotes projected SGD, Proxskip is in \citep{mishchenko2022proxskip}, ALM denotes Augmented Lagrangian method, NS denotes Neumann series approximation, and CG denotes conjugate gradient method. The $w$-update of IG-AL includes computing the exact Hessian inverse, so we denote it as `exact'. The notation $\tilde{\cal O}$ omits the polynomial dependency on $\log(\epsilon^{-1})$ terms.  } 
    \label{tab:table1}
     \vspace{-0.2cm}
\end{table}

\subsection{Related works}
To put our work in context, we review prior art from the following two categories. We summarize the comparison of our work with the closely related prior art in Table \ref{tab:table1}.  
 
\paragraph{Unconstrained bilevel optimization.} 
Bilevel optimization has a long history back to  \citep{bracken1973mathematical} and has inspired rich literature, e.g., \citep{ye1995optimality,vicente1994bilevel,colson2007overview,sinha2017review}. Later on, spurred by the advancement of hyperparameter optimization \citep{maclaurin2015gradient,franceschi2018bilevel} and meta-learning \citep{finn2017model}, bilevel optimization received more attention as a unified tool for problems with nested structures. With more use cases in large-scale machine learning, developing stochastic methods with finite-time guarantees has become the recent focus in the area of bilevel optimization. The interest in the nonasymptotic analysis of the stochastic bilevel optimization has been stimulated since a recent work \citep{ghadimi2018approximation} that tackles the bilevel setting where the LL objective is strongly convex and Lipschitz smooth. As $\nabla F(x)$ contains the Hessian inverse of the LL objective which is computationally expensive, it has emerged various numeric approximation methods including Neumann series approximation \citep{ghadimi2018approximation}, unrolling differentiation \citep{grazzi2020iteration}, and conjugate gradient \citep{ji2021bilevel}; see  comparisons in \citep{ji2022will,lorraine2020optimizing}. In terms of the alternating implicit SGD algorithm, \citep{chen2021closing} achieved the $\tilde{\cal O}(\epsilon^{-2})$ sample complexity, which matches the results for the single-level case.  Beyond the alternating implicit SGD framework, \citep{khanduri2021near,yang2021provably} incorporated variance reduction techniques to further accelerate the convergence, \citep{li2022fully} has put forward a fully single-loop algorithm with UL variance reduction and achieved ${\cal O}(\epsilon^{-2})$ sample complexity, and later on, \citep{dagreou2022framework} generalized this framework to allow global variance reduction for each level and further enhanced the convergence rate; see a recent survey for bilevel optimization \citep{liu2021investigating}. Nevertheless, none of these attempts can solve \eqref{opt0} in the presence of both UL and LL constraints.

\paragraph{Constrained bilevel optimization.} 
While the nonasymptotic convergence for various approaches in the unconstrained bilevel setting has been extensively studied in the literature, the nonasymptotic analysis of stochastic algorithms for constrained bilevel optimization problems is very limited. Some recent efforts have been devoted to tackling the constrained UL setting. \citep{hong2020two} has established ${\cal O}(\epsilon^{-2.5})$ rate of TTSA which applied SGD in LL update and projected SGD in UL update; \citep{chen2022single} has achieved ${\cal O}(\epsilon^{-2})$ convergence rate by adding additional corrections on LL update for the stochastic setting;
\citep{chen2022fast} has proved the ${\cal O}(\epsilon^{-1})$ convergence rate of the proximal accelerated gradient-based method for deterministic constrained UL problem under the Kurdyka-{\L}ojasiewicz geometry. As for constrained LL problem, the vast majority of works focus on either asymptotic analysis, e.g.,  initialization auxiliary method \citep{liu2021towards}, value function based approach \citep{gao2022value}; or design aspects, e.g. optimality of bilevel problem \citep{dempe2007new,ye2010new}, reformulation \citep{dempe2013bilevel,brotcorne2013one}, and implicit differential properties \citep{gould2016differentiating,bolte2021nonsmooth,bertrand2022implicit}. The notable exception is a recent work \citep{tsaknakis2022implicit}, which solved linearly inequality constrained LL problem in a double-loop manner, i.e. update UL variable after attaining a sufficiently accurate LL solution. However, the overall iteration complexity has not been established therein.

\section{AiPOD: Alternating Implicit Projected SGD for Bilevel Problems}
In this section, we first introduce notations, present the   algorithm and establish its convergence. 
 
\subsection{Preliminaries}
For convenience, we define $g(x,y):=\EE_{\phi}\left[g(x,y;\phi)\right]$ and $f(x,y):=\EE_{\xi}\left[f(x,y;\xi)\right]$. We also define $\nabla_{yy}g(x,y)$ as the Hessian of $g$ with respsect to $y$ and denote
\begin{align*}
    \nabla_{xy}g(x,y)=\left[\begin{array}{ccc}
\frac{\partial^2 }{\partial x_1\partial y_1} g(x,y)& \cdots & \frac{\partial^2 }{\partial x_1\partial y_{d_y}} g(x,y)\\
\vdots & \ddots & \vdots \\
\frac{\partial^2 }{\partial x_{d_x}\partial y_1} g(x,y) & \cdots & \frac{\partial^2 }{\partial x_{d_x}\partial y_{d_y}} g(x,y)
\end{array}\right].
\end{align*}
We use $\|\cdot\|$ to denote the $\ell_2$ norm for vectors and Frobenius norm for matrix. We also denote $A^\dagger$ and $B^\dagger$ as the the Moore-Penrose inverse of $A$ and $B$ \citep{james1978generalised}. Moreover, we define $P_x:=I-B^\dagger B$ and $P_y:=I-A^\dagger A$ as the projection matrix over $x$ and $y$, and denote $\|x\|_{P_x}:=\sqrt{x^\top P_x x}$ and $ \|y\|_{P_y}:=\sqrt{y^\top P_y y}$ as the $P_x$ and $P_y$ weighted Euclidean norm, respectively. We also let $V_1$ be the orthogonal basis of $\operatorname{Ran}(A^\top):=\{A^\top y\}$ and $V_2$ be the orthogonal basis of $\operatorname{Ker}(A):=\{y\mid Ay=0\}$.

The common convergence metric for constrained optimization is  \citep{ghadimi2016mini}  
\begin{align}\label{davis-measure}
    \EE[\|\lambda^{-1}( x-\operatorname{Proj}_{\mathcal{X}}(x-\lambda\nabla F(x)))\|^2]
\end{align}
for some $\lambda>0$. 
In \eqref{opt0}, since $\mathcal{X}$ contains only linear equality constraints, \eqref{davis-measure} can be simplified according to the following lemma, the proof of which will be deferred to Appendix \ref{sec:app-pre}. 
\begin{lemma}\label{eq:measure}
For any $x\in\mathcal{X}:=\{x\mid Bx=e\}$ and any $\lambda> 0$, we have that
\begin{align*}
    \|\lambda^{-1}( x-\operatorname{Proj}_{\mathcal{X}}( x-\lambda\nabla F( x)))\|^2=\|\nabla F(x)\|^2_{P_x}.
\end{align*}
\end{lemma}

Therefore, we have the following definition of the $\epsilon$ stationary point.

\paragraph{The $\epsilon$ stationary point.} 
We define the $\epsilon$ stationary point $x$ for \eqref{opt0} as
\begin{align}\label{stationary}
    \EE\left[\|\nabla F( x)\|_{P_x}^2\right]\leq\epsilon.
\end{align}

If $B=0$,  \eqref{stationary} is reduced to $\EE\left[\|\nabla F( x)\|^2\right]\leq\epsilon$, which is the standard stationary measure for unconstrained stochastic bilevel optimization settings \citep{ghadimi2018approximation,ji2021bilevel, chen2021closing}. 

 \begin{wrapfigure}{R}{0.52\textwidth}
\vspace{-0.4cm}
 \begin{minipage}{0.53\textwidth} 
\begin{algorithm}[H]
\caption{AiPOD for constrained bilevel problem}
\begin{algorithmic}[1]
\State Initialization: $x^0,y^0$, stepsizes $\{\alpha,\beta\}$, the number of LL and UL rounds $\{S,K\}$, the number of $\phi$ samples $N$
\For{$k=0$ {\bfseries to} $K-1$}
\For{$s=0$ {\bfseries to} $S-1$}~~~~~~~~~\Comment{Set $y^{k,0}=y^{k}$}
\State update $y^{k,s+1}$ by \eqref{update-ys}. 
\EndFor~~~~~~~~~~~~~~~~~~~~~~~~~~~~~~\Comment{Set $y^{k+1}=y^{k,S}$}
\State evaluate $w^k$ in \eqref{NS-wk} and $h_f^k$ in \eqref{NS-hf}
\State update $x^{k+1}=\operatorname{Proj}_{\mathcal{X}} (x^k-\alpha h_f^k)$ 
\EndFor
\end{algorithmic}
\label{stoc-alg}
\end{algorithm}
\end{minipage}
\end{wrapfigure}

\subsection{The basic algorithm}
In this section, we will introduce the basic version of {\sf AiPOD} algorithm for \eqref{opt0}, which updates $x$ and $y$ in an alternating implicit projected SGD manner. 

At a given UL iteration $k$, we update $y^{k+1}$ by the output of the $S$ projected SGD steps for $g(x^k,y)$. With initialization $y^{k,0}=y^k$, we update
\begin{align}\label{update-ys}
    y^{k,s+1}=\operatorname{Proj}_{\mathcal{Y}(x^k)}(y^{k,s}-\beta\nabla g(x^k,y^{k,s};\phi^{k,s}))
\end{align}
and set $y^{k+1}=y^{k,S}$. For UL, the gradient $\nabla F(x)$ can be calculated by the chain rule 
\begin{align}\label{eq.F_grad}\small
     \nabla F(x)&=\nabla_x f(x,y^*(x))+\nabla^\top y^*(x)\nabla_y f(x,y^*(x)). 
\end{align}
Therefore, the implicit mapping $\nabla y^*(x)$ is essential to the UL update. 

Without LL constraint, $\nabla y^*(x)$ can be derived from the LL optimality condition $\nabla_y g(x,y^*(x))=0$ as \citep{ghadimi2018approximation}
\begin{align}\label{grad-y*}\small
    \nabla y^*(x)=-\nabla_{yy}^{-1}g(x,y^*(x))\nabla_{yx}g(x,y^*(x)).
\end{align}
We generalize it to the constrained LL setting and establish the implicit gradient for constrained LL in the following lemma, the proof of which is deferred to Appendix \ref{sec:app-pre}. 

\begin{lemma}\label{y-imgrad}
Define $V_2$ as the orthogonal basis of $\operatorname{Ker}(A):=\{y\mid Ay=0\}$. 
When $g(x,y)$ is twice differentiable and strongly convex over $y$, the implicit gradient $\nabla y^*(x)$ can be written as
\begin{equation}\label{eq.y_grad}
 \!\!   \nabla y^*(x)\!=\!-\underbrace{V_2(V_2^\top \nabla_{yy}g(x,y^*(x))V_2)^{-1}V_2^\top}_{P_1}\underbrace{\left(\nabla_{yx}g(x,y^*(x))-\nabla_{yy}g(x,y^*(x)) A^\dagger \nabla h(x)\right)-A^\dagger\nabla h(x)}_{P_2}\!\!
\end{equation}
where $A^\dagger$ is the Moore-Penrose inverse of $A$. 
\end{lemma}

Compared with \eqref{grad-y*}, the term $P_1$ in \eqref{eq.y_grad} can be roughly viewed as projecting $\nabla_{yy}^{-1}g(x,y^*(x))$ to $\operatorname{Ker}(A)$; while $P_2$ accounts for the coupling constraints $Ay+h(x)=c$. 

With similar spirits with the existing works \citep{ghadimi2018approximation,hong2020two,chen2021closing}, we obtain the UL gradient estimator $h_f^k$ at UL iteration $k$ by setting $x=x^k$, approximating $y^*(x^k)$ by $y^{k+1}$ in \eqref{eq.F_grad} and estimating $P_1$ by Neumann series. Formally, $h_f^k$ is defined as
\begin{subequations}
\begin{align} \small
&~~~~~~~~~~~~~~~~~~~~~~~~~~~~~~~~~~~~h_f^k:= \nabla_{x} f(x^{k}, y^{k+1} ; \xi^{k}) + w^k\label{NS-hf}\\
\text{with }w&^k:=\left\{-\nabla h(x^k)^\top A^{\dagger\top}+\left(\nabla h(x^k)^\top A^{\dagger\top}\nabla_{yy} g(x^{k}, y^{k+1} ; \phi_{(0)}^{k})-\nabla_{x y} g(x^{k}, y^{k+1} ; \phi_{(0)}^{k})\right)\right.\nonumber\\
&~~\left.\times V_2\left[\frac{\tilde cN}{\ell_{g, 1}} \prod_{n=1}^{N^{\prime}}\left(I-\frac{\tilde c}{\ell_{g, 1}} V_2^\top\nabla_{y y} g\left(x^{k}, y^{k+1} ; \phi_{(n)}^{k}\right)V_2\right)\right] V_2^\top\right\}\nabla_{y} f(x^{k}, y^{k+1} ; \xi^{k})\label{NS-wk}
\end{align}
\end{subequations}
where $\tilde c\in(0,1]$ is a given constant, $N^\prime$ is drawn uniformly at random from $\{0,\cdots N-1\}$, and $\{\phi_{(0)}^k,\cdots,\phi_{(N^\prime)}^k\}$ are i.i.d samples. 

Afterward, we can update $x^{k+1}$ by projected SGD with estimator $h_f^k$ in \eqref{NS-hf}. The full {\sf AiPOD} algorithm is summarized in Algorithm \ref{stoc-alg}.

\subsection{Theoretical analysis}

For the subsequent analysis, we make the following assumptions. 

\begin{assumption}\label{as1}
Assume that $f,\nabla f,\nabla g,\nabla_{xy} g,\nabla_{yy} g, h$ and $\nabla h$ are Lipschitz continuous with $\ell_{f,0},$
$\ell_{f,1},\ell_{g,1},\ell_{g,2},\ell_{g,2},\ell_{h,0}$ and $\ell_{h,1}$, respectively. 
\end{assumption}

\begin{assumption}\label{as2}
For any fixed $x$, assume that $g(x,y)$ is $\mu_g$-strongly convex with respect to $y\in\mathbb{R}^{d_y}$.
\end{assumption}

\begin{assumption}\label{as-stoc}
The stochastic estimators $\nabla f(x,y;\xi),\nabla g(x,y;\phi), \nabla_{xy} g(x,y;\phi)$ and $\nabla_{yy} g(x,y;\phi)$ are unbiased estimators of $\nabla f(x,y),\nabla g(x,y),\nabla_{xy} g(x,y)$ and $\nabla_{yy} g(x,y)$, and their variance are bounded by $\sigma_f^2,\sigma_{g,1}^2,\sigma_{g,2}^2$ and $\sigma_{g,2}^2$, respectively.
\end{assumption}

\begin{assumption}\label{ash}
The set $\mathcal{X}$ is nonempty. For any $x$, the set $\mathcal{Y}(x)$ is nonempty. 
\end{assumption}

Assumption \ref{as1}--\ref{as-stoc} are standard for stochastic bilevel optimization \citep{ghadimi2018approximation,hong2020two,ji2021bilevel,khanduri2021near,chen2021closing,chen2022single,li2022fully}.  Assumption \ref{ash} is to ensure the feasibility of the problem \eqref{opt0}. Since $Ay=b$ has solution $y$ if and only if $AA^\dagger b=b$ according to \citep{james1978generalised}, Assumption \ref{ash} is equivalent to $AA^\dagger(c-h(x))=c-h(x)$. One sufficient but not necessary condition for Assumption \ref{ash} is that $A$ is full row rank, which does not impose any additional requirement on $h(x)$. Another sufficient but not necessary condition for Assumption \ref{ash} is $\forall x$, $c-h(x)\in\operatorname{Ran}(A)$, e.g., $h=0, c\in\operatorname{Ran}(A)$, which does not assume $A$ is full row rank. 



One of the keys to establishing $\tilde{\cal O}(\epsilon^{-2})$ sample complexity of unconstrained bilevel optimization \citep{chen2021closing,li2022fully} is to utilize the smoothness of $ y^*(x)$. Thanks to the singular value decomposition, we can obtain the smoothness of $ y^*(x)$ for linearly equality constrained LL \eqref{opt0-0-low} in the following lemma, the proof of which is deferred to Appendix \ref{sec:app-pre2}. 

\begin{lemma}\label{y-smooth}
Under Assumption \ref{as1}--\ref{as2} and \ref{ash}, $y^*(x)$ is $L_y$- Lipschitz continuous and $L_{yx}$- smooth, where the constants $L_y$ and $L_{yx}$ are specified in Appendix \ref{sec:app-pre2}.
\end{lemma}

However, due to the UL constraint, the proof for unconstrained stochastic bilevel optimization \citep{chen2021closing} cannot be applied even with the smoothness of $y^*(x)$ in Lemma \ref{y-smooth}. For constrained UL with LL unconstrained bilevel problem, recent works \citep{hong2020two,chen2022single} leveraged the Moreau envelope technique of projected SGD in \citep{davis2019stochastic}. However, when the stochastic gradient estimator is biased, the gradient bias term can not be mitigated by the framework in \citep{davis2019stochastic}. Therefore, \citep{hong2020two} ended up with suboptimal sample complexity ${\cal O}(\epsilon^{-2.5})$, while \citep{chen2022single} added an additional correction in the LL update. Owing to the special property of linear equality constrained UL presented in Lemma \ref{eq:measure}, we establish the $\tilde{\mathcal O}(\epsilon^{-2})$ sample complexity of Algorithm \ref{stoc-alg} in the next lemma without resorting to the Moreau envelope technique. The proof is deferred to Appendix \ref{sec:thm-alg1-conv}.

\begin{theorem}[\bf Convergence rate of {\sf AiPOD}]\label{thm}
Under Assumption \ref{as1}--\ref{ash}, if we choose $$\alpha=\min\left(\bar\alpha_1,\bar\alpha_2,\frac{\bar\alpha}{\sqrt{K}}\right),\qquad \beta=\frac{5L_fL_y+  L_{yx}\tilde C_f^2}{\mu_g}\alpha, \qquad N={\cal O}(\log K)$$
where $\bar\alpha_1,\bar\alpha_2$ are defined in Appendix \ref{sec:thm-alg1-conv},
then for any $S={\cal O}(1)$ in Algorithm \ref{stoc-alg}, we have
\begin{align*}
    \frac{1}{K}\sum_{k=0}^{K-1} \EE\left[\|\nabla F(x^k)\|_{P_x}^2\right]=\tilde{\cal O}\left(\frac{1}{\sqrt{K}}\right).
\end{align*}
\end{theorem}
Theorem \ref{thm} shows that Algorithm \ref{stoc-alg} achieves $\epsilon$ stationary point by $\tilde{\cal O}(\epsilon^{-2})$ iterations, which matches the iteration complexity of single-level projected SGD method \citep{davis2019stochastic}. Moreover, the sample complexity of AiPOD is $\tilde{\cal O}(K)=\tilde{\cal O}(\epsilon^{-2})$, which matches the unconstrained bilevel SGD method \citep{chen2021closing}.

\begin{wrapfigure}{R}{0.57\textwidth}
\vspace{- .6cm}
\begin{minipage}{0.57\textwidth} 
\begin{algorithm}[H]
\caption{{\sf E-AiPOD} for constrained bilevel problem} 
\begin{algorithmic}[1]
\State Initialization: $x^0,y^0$, stepsizes $\{\alpha,\beta,\delta\}$, skipping probability $p$, the number of rounds $\{S,K\}$, UL projection frequency $T$ and the number of $\phi$ samples $N$
\For{$k=0$ {\bfseries to} $K-1$}
\State $\{y^{k+1},r^{k+1}\}=${\sf E-AiPOD}$_{\rm{low}}(x^k,y^k,r^k,\beta,p,S)$
\State compute $w^k$ defined in \eqref{NS-wk}. 
\State initialize $x^{k,0}=x^{k}$
\For{$t=0$ {\bfseries to} $T-1$}
\State calculate $h_f^{k,t}$ in \eqref{NS-hf-t}
\State update $x^{k,t+1}$ by \eqref{update-xkt} 
\EndFor
\State update $x^{k+1}$ by \eqref{delta-x}
\EndFor
\end{algorithmic}
\label{stoc-alg-skip}
\end{algorithm}
\end{minipage}
\vspace{-1.cm}
\end{wrapfigure} 

\section{E-AiPOD: A Projection-efficient Variant  of {\sf AiPOD}}\label{sec:skip}
In this section, we focus on the case when evaluating projection is expensive and propose a projection efficient variant of {\sf AiPOD} that we term {\sf E-AiPOD} to avoid frequent projection steps. 

\subsection{Algorithm development}
Besides the explicit projections for $x$- and $y$- updates in Algorithm \ref{stoc-alg}, calculating $w^k$ in \eqref{NS-wk} also requires projecting $\nabla_{yy}g(x^k,y^{k+1};\phi_{(n)}^k)$ onto the null space of the LL problem, e.g., calculating $V_2^\top\nabla_{yy}g(x^k,y^{k+1};\phi_{(n)}^k)V_2$. The explicit projections in \eqref{update-ys} are referred as LL projections, while both the explicit projections in $x$-update and implicit projections contained in $w^k$ are regarded as UL projections. We use the following ways to save projections. 

\paragraph{Skip LL projections.} We leverage a recent variant of projected SGD called Proxskip \citep{mishchenko2022proxskip} in the LL update which evaluates projection lazily with probability $0<p<1$. Fixing UL iteration $k$ and at each LL iteration $s$, we first perform an SGD update corrected by the residual $r^{k,s}$ as
\begin{subequations}
\begin{align}\label{update-z}
    \hat{y}^{k,s+1}=y^{k,s}-\beta(\nabla_y g(x^k,y^{k,s};\phi^{k,s})-r^{k,s}).
\end{align}
With probability $1-p$, we skip the projection and keep $y^{k,s+1}=\hat{y}^{k,s+1},r^{k,s+1}=r^{k,s}$; with probability $p$, we update LL parameter $y^{k,s}$ and the residual $r^{k,s}$ as
\begin{align}
    &y^{k,s+1}=\operatorname{Proj}_{\mathcal{Y}(x^k)} \left(\hat{y}^{k,s+1}-\beta r^{k,s}/p\right)\label{update-yz1}\\
    &r^{k,s+1}=r^{k,s}+p(y^{k,s+1}-\hat{y}^{k,s+1})/\beta\label{update-yz2}. 
\end{align}
\end{subequations}
In \eqref{update-z}, $r^{k,s}$ compensates the error of skipping projection which will be updated every $1/p$ rounds in expectation so that the corrected stochastic gradient update  in \eqref{update-z} will approximate the projected SGD update in \eqref{update-ys}. The update of residual sequences  in \eqref{update-yz2} ensures that $r^{k,s}$ converges to the projection skipping errors at the optimal point $y^*(x^k)$, i.e. $\lim_{s\rightarrow\infty}r^{k,s}=\nabla_y g(x^k,y^*(x^k))=:r^*(x)$, so that the asymptotic convergence for {\sf E-AiPOD} is similar to {\sf AiPOD}. With $K$ steps of UL updates and $S$ steps of LL updates, the expected  number of projection evaluations   reduces from $KS$ in {\sf AiPOD} to $pKS$ in {\sf E-AiPOD}. We summarize the LL update of {\sf E-AiPOD} in Algorithm \ref{stoc-alg-skip-LL}. 

\paragraph{Use delayed $w^k$ and reduce UL projections.} 
At upper iteration $k$, we calculate $w^k$ by the Neumann series \eqref{NS-wk} and initialize $x^{k,0}=x^k$. Subsequently, for $t=0$ to $T-1$, we update $x^{k,t+1}$ via
\begin{subequations}\label{11ab}
\begin{align}
    x^{k,t+1}=&\,x^{k,t}-\alpha h_f^{k,t}\label{update-xkt}\\
{\rm where}~~~~h_f^{k,t}:= & \, \nabla_{x} f(x^{k,t}, y^{k+1} ; \xi^{k,t})+w^k\label{NS-hf-t}. 
\end{align}
\end{subequations}
Compared with $h_f^{k}$ in \eqref{NS-hf}, $h_f^{k,t}$ can be regarded as the UL gradient estimator at $x^{k,t}$ obtained by delayed Hessian inverse vector approximation $w^k$. After $T$ rounds, we update $x^{k+1}$ by 
\begin{align}\label{delta-x}
    x^{k+1}=(1-\delta)x^k+\delta\operatorname{Proj}_{\mathcal{X}}(x^{k,T}) 
\end{align}
where $\delta\geq 1$ positively correlates to $T$ so that scales the projected descent stepsizes. In this way, we only project the Hessian estimator to the null space to evaluate $w^k$ at $t=0$ and project $x$ sequence at the end of the $T$-loop. The error resulting from the  delayed $w^k$ will be shown to be bounded by ${\cal O}(\alpha^2)$, which is not the dominating term in the analysis. We summarize {\sf E-AiPOD} in Algorithm \ref{stoc-alg-skip}. 



\subsection{Theoretical analysis}
We characterize its convergence rate by using a new Lyapunov function as 
 \begin{align}
 \mathbb{V}_1^k\:= F(x^k)+\frac{ L_f}{L_r}\left(\|y^*(x^k)-y^k\|^2+\frac{\beta^2}{p^2}\|r^{k}-r^*(x^k)\|^2\right)\label{lyapunov-skip}
\end{align}
where $L_f$ and $L_r$ are constants defined in Lemma \ref{smooth-barf} and \ref{smooth-r} in Appendix and $r^*(x^k)=\nabla_y g(x,y^*(x))$. The complexity bound of {\sf E-AiPOD} is stated in the following theorem.  





\begin{figure*}
\begin{minipage}{0.49\textwidth}
\begin{algorithm}[H]
\caption{{\sf E-AiPOD}$_{\rm{low}}(x^k,y^k,r^k,\beta,p,S)$} 
\begin{algorithmic}[1]
\State Inputs: $x^k,y^k,r^k$, stepsize $\beta$, skipping probability $p$, the number of LL rounds $S$
\State Initialization $y^{k,0}=y^{k}, r^{k,0}=r^{k}$
\For{$s=0$ {\bfseries to} $S-1$}
\State update $\hat{y}^{k,s+1}$ by \eqref{update-z}
\State draw $\theta^{k,s}$ with $\mathbb{P}(\theta^{k,s}=1)=p$
\If{$\theta^{k,s}=1$}
\State update $y^{k,s+1}$ by \eqref{update-yz1}
\Else
\State $y^{k,s+1}=\hat{y}^{k,s+1}$
\EndIf
\State update $r^{k,s+1}$ by \eqref{update-yz2} 
\EndFor
\State Outputs: $y^{k+1}=y^{k,S},r^{k+1}=r^{k,S}$
\end{algorithmic}
\label{stoc-alg-skip-LL}
\end{algorithm}
\end{minipage}
\hfill
\begin{minipage}{0.49\textwidth}
\begin{algorithm}[H]
\caption{{\sf E2-AiPOD}$_{\rm{med}}(x^k,y^{k+1},\rho,q,N)$} 
\begin{algorithmic}[1]
\State Inputs: $x^k,y^{k+1}$, stepsize $\rho$, skipping probability $q$, the number of rounds $N$
\State Initialization $u^{k,0}=0, e^{k,0}=0$
\For{$n=0$ {\bfseries to} $N-1$}
\State update $\hat{u}^{k,n+1}$ by \eqref{update-hatlambda-WS}
\State draw $\tilde\theta^{k,n}$ with $\mathbb{P}(\tilde\theta^{k,n}=1)=q$
\If{$\tilde\theta^{k,n}=1$}
\State update $u^{k,n+1}$ by \eqref{update-lz1-WS}
\Else
\State set $u^{k,n+1}=\hat{u}^{k,n+1}$
\EndIf
\State update $e^{k,n+1}$ by \eqref{update-lz2-WS} 
\EndFor
\State Outputs: $u^{k+1}=u^{k,N}$
\end{algorithmic}
\label{stoc-alg-skip-hessian}
\end{algorithm}
\end{minipage}
\end{figure*}

\begin{theorem}[\bf Convergence of {\sf E-AiPOD}]\label{alskip-conv}
Under Assumption \ref{as1}--\ref{ash}, if we choose stepsizes such that $\alpha\delta T<\bar\alpha$, where $\bar\alpha<1$ is a constant formally defined in \eqref{baralpha}, and let $\beta={\cal O}(\alpha T), p={\cal O}(\sqrt{\beta}), S={\cal O}(1), N={\cal O}(\log(\alpha^{-1}))$, then the sequences generated by Algorithm \ref{stoc-alg-skip} satisfies
\begin{align}\label{alskip}
    \frac{1}{K}\sum_{k=0}^{K-1}\EE\left[\|\nabla F(x^k)\|_{P_x}^2\right]\leq \frac{2(\mathbb{V}_1^{0}-F^*)}{\alpha \delta T K}+2c_1\sigma_{g,1}^2\alpha\delta T+2c_2\tilde\sigma_f^2\alpha\delta+{\cal O}(\alpha^2\delta^2 T)
\end{align}
where $c_1$ and $c_2$ are constant formally defined in Appendix \ref{sec:proxskip-conv}, $F^*$ is the lower bound of $F(x)$, and $\sigma_{g,1}^2, \tilde\sigma_{f,2}^2$ are the variance of LL and UL updates.
\end{theorem}
With proper choices of $\alpha$ and $\delta$, the four terms in \eqref{alskip} could vanish simultaneously. The following corollary shows the results for vanilla periodical projections when $\delta=1$. 

\begin{corollary}[\bf Reduction of LL projection]\label{redu-LL}
Under the same condition of Theorem \ref{alskip-conv}, if we choose $\alpha=\frac{\bar\alpha}{T\sqrt{K}}$, $\delta=1$, the convergence rate of Algorithm \ref{stoc-alg-skip} is $$\frac{1}{K}\sum_{k=0}^{K-1}\EE\left[\|\nabla F(x^k)\|_{P_x}^2\right]=\tilde{\cal O}\left(\frac{1}{\sqrt{K}}\right).$$
Since $K=\tilde{\cal O}(\epsilon^{-2})$ and $p={\cal O}(\sqrt{\beta})={\cal O}(K^{-\frac{1}{4}})$, the total number of LL projections   reduces to
\begin{align*}
    pKS={\cal O}(K^{\frac{3}{4}})=\tilde{\cal O}(\epsilon^{-1.5}).
\end{align*}
\end{corollary}
Corollary \ref{redu-LL} implies that the convergence rate of {\sf E-AiPOD} is the same as that of {\sf AiPOD} when $\delta=1$, but the LL projection complexity can be reduced to $\tilde{\cal O}(\epsilon^{-1.5})$. Compared with Proxskip \citep{mishchenko2022proxskip} which improves the projection complexity on $\kappa=\ell_{g,1}/\mu_g$, we can further achieve the reduction on $\epsilon$ owing to the smaller LL stepsize $\beta={\cal O}(1/\sqrt{K})$. Besides, the next corollary shows the benefit for enlarging $\delta$, which further reduces the iteration and projection complexity of {\sf E-AiPOD}.

\begin{corollary}[\bf Reduction of UL projection]\label{redu-UL}
Under the same condition of Theorem \ref{alskip-conv}, if we choose $\alpha=\frac{\bar\alpha}{T\sqrt{K}}$, $\delta=\sqrt{T}$ with $T<K$, and select LL batch size as ${\cal O}(T)$ such that $\sigma_g^2={\cal O}(1/T)$, then the convergence rate of Algorithm \ref{stoc-alg-skip} is 
\begin{align*}
\frac{1}{K}\sum_{k=0}^{K-1}\EE\left[\|\nabla F(x^k)\|_{P_x}^2\right]=\tilde{\cal O}\left(\frac{1}{\sqrt{TK}}\right).
\end{align*}
As a result, the sample complexity of E-AiPOD for both UL and LL are $\tilde{\cal O}(TK)=\tilde{\cal O}(\epsilon^{-2})$, while the total number of the UL and LL projection are respectively, reduced to
\begin{align*}
    \textrm{UL}: K+KN=\tilde{\cal O}(\epsilon^{-2}/T), ~~~~~{\rm and}~~~~~\textrm{LL}: pKS={\cal O}(K^{\frac{3}{4}})=\tilde{\cal O}(\epsilon^{-1.5}/T^{\frac{3}{4}}).
\end{align*}
\end{corollary}
Corollary \ref{redu-UL} implies with larger $\delta$, increasing $T$ can accelerate the convergence rate and improve the projection complexity without degrading the sample complexity for both levels. Compared with the single-level case, $\sigma_g^2$ can be seen as the additional error caused by the LL stochasticity. The idea behind reducing the UL projection complexity is to reduce the variance of the averaged gradient estimator by $T$ gradient descent steps and use larger $\delta$ to balance the projection frequency. However, this can not reduce the variance of LL; see the different terms for $\tilde\sigma_f^2$ and  $\sigma_g^2$ over $T$ in \eqref{alskip}. Therefore, we need to increase the LL batch size correspondingly to reduce $\sigma_g^2$ in the bilevel problem. By virtue of the faster rate, the sample complexity for LL is $\tilde{\cal O}(K)$, which is still $\tilde{\cal O}(\epsilon^{-2})$. 

\section{E2-AiPOD: A Projection-efficient Variant of {\sf AiPOD} without Coupling Constraints}\label{sec.e2-aipod}
In Section \ref{sec:skip}, we have developed {\sf E-AiPOD} that reduces both UL and LL projection complexity with an increasing LL batch size ${\cal O}(T)$. 
The essential obstacle for preventing {\sf E-AiPOD} from using constant LL batch size is that the delayed $w^k$ in \eqref{NS-hf-t} would bring error to the UL gradient estimator. 
This section overcomes this obstacle in the setting where 
the LL constraints are independent of UL variable $x$, i.e., $h(x)=0$, and introduces {\sf E2-AiPOD} - another projection-efficient variant of {\sf AiPOD} that reduces both UL and LL projection complexity with a constant LL batch size.  
\subsection{Algorithm development}

\textbf{Intuition.} We first provide some intuition of the proposed {\sf E2-AiPOD} method via the deterministic version of \eqref{opt0}.  Instead of calculating $w^k$ in \eqref{NS-wk} of {\sf E-AiPOD} which requires expected projections $N$ times, we can approximate $w^k$ by a sequence and then use the idea of skipping projections to reduce the number of projections. 
According to \eqref{eq.F_grad} and \eqref{eq.y_grad} and letting $h(x)=0$, it follows that the UL gradient of \eqref{opt0} can also be given by
\begin{align*}
   &\nabla F(x)= \nabla_x f(x,y^*(x))+\nabla_{xy}g(x,y^*(x))u^*(x,y^*(x))\\
    &\text{where }  u^*(x,y):=-V_2\left(V_2^\top\nabla_{yy}g(x,y)V_2\right)^{-1}V_2^\top\nabla_y f(x,y).\numberthis\label{lbda*}
\end{align*}
In {\sf E-AiPOD}, $u^*(x^k,y^{k+1})$ is estimated via Neumann series to approximate the Hessian inverse; see  \eqref{NS-hf}. Instead, we can view $u^*(x,y)$ as an optimal solution of a linear constrained quadratic programming and approximate it using iterative methods. Formally, we have the following lemma. 
\begin{lemma}[Equivalent characterization of $u^*(x,y)$]\label{lm:u*-strong}
The Hessian inverse estimator $u^*(x,y)$ is the minimizer of a $\mu_g$- strongly convex function over a linear space, that is
\begin{equation}\label{eq.lemma4}
    u^*(x,y)=\argmin_{u\in\{u|V_1^\top u=0\}}\frac{1}{2}\left\|\nabla_{yy}^{-\frac{1}{2}}g(x,y)\nabla_y f(x,y)+\nabla_{yy}^{\frac{1}{2}}g(x,y)u\right\|^2.
\end{equation}
where $V_1$ is the orthogonal basis of $\operatorname{Ran}(A^\top)$. 
\end{lemma}
Thanks to the above lemma, we can treat $u^*(x^k,y^{k+1})$ as an optimal solution of an optimization problem and iteratively update another sequence $\{u^{k,n}\}$ to approximate it. As a consequence, the solution of the original problem \eqref{opt0} can be obtained by three coupled sequences.  
For simplicity, we call the optimization with respect to $u^*(x,y)$ the medium-level (ML) problem. Assuming that $\{\phi^{k},\phi^{k,0},\cdots\phi^{k,S-1},\cdots,\phi_{(0)}^k,\cdots,\phi_{({N-1})}^k\}$ and $\{\xi^{k},\xi_{(0)}^k,\cdots,\xi_{({N-1})}^k\}$ are i.i.d samples, we then elaborate the procedure of {\sf E2-AiPOD}. 
\begin{wrapfigure}{R}{0.58\textwidth}
\vspace{-.8cm}
\begin{minipage}{0.58\textwidth} 
\begin{algorithm}[H]
\caption{E2-AiPOD for constrained bilevel problem} 
\begin{algorithmic}[1]
\State Initialization: $x^0,y^0$, stepsizes $\{\alpha,\beta,\rho\}$, the number of rounds $\{S,N,K\}$ and projection frequency $\{p,q, T\}$. 
\For{$k=0$ {\bfseries to} $K-1$}
\State $\{y^{k+1},r^{k+1}\}=$~{\sf E-AiPOD}$_{\rm{low}}(x^k,y^k,r^k,\beta,p,S)$
\State update $u^{k+1}=$~{\sf E2-AiPOD}$_{\rm{med}}(x^k,y^{k+1},\rho,q,N)$
\State compute $d_f^k$ defined in \eqref{WS-df}. 
\If{$k\operatorname{mod}T=0$}
\State update  $x^{k+1}=\operatorname{Proj}_{\mathcal{X}}(x^{k}-\alpha d_f^k)$ in \eqref{WS-x}
\Else
\State update $x^{k+1}=x^{k}-\alpha d_f^k$
\EndIf
\EndFor
\end{algorithmic}
\label{stoc-alg-skip-WS}
\end{algorithm}
\end{minipage}
\end{wrapfigure}

\paragraph{Skip LL projections.} The LL update for {\sf E2-AiPOD} is identical to {\sf E-AiPOD}. 

\paragraph{Skip UL implicit projections.} The implicit UL projections correspond to the ML projections. We update $u^k$ by $N$- step lazily projected SGD with corrections, which is in the same spirit of the LL $y$-update. Specifically, for any UL iteration $k$, we first initialize $u^{k,0}=e^{k,0}=0$. For a given ML iteration $n$, instead of using a projected SGD iteration, we update $u^{k,n+1}$ via the SGD iteration for \eqref{eq.lemma4} corrected by $e^{k,n}$ that compensates the projection error, that is
\begin{subequations}
\begin{align}\label{update-hatlambda-WS}
    \hat{u}^{k,n+1}&=u^{k,n}-\rho\left(\nabla_y f(x^k,y^{k+1};\xi^{k}_{(n)})+\nabla_{yy}g(x^k,y^{k+1};\phi^{k}_{(n)})u^{k,n}-e^{k,n}\right)
\end{align}
where $\rho>0$ is the stepsize of the ML update. 
Subsequently, with probability $1-q$, we keep $u^{k,n+1}=\hat{u}^{k,n+1},e^{k,n+1}=e^{k,n+1}$; with probability $q$, we update $u^{k,n+1}$ and the $e^{k,n+1}$ as
\begin{align}
    &u^{k,n+1}=V_2V_2^\top \left(\hat{u}^{k,n+1}-\rho e^{k,n}/q\right)\label{update-lz1-WS}\\
    &e^{k,n+1}=e^{k,n}+q(u^{k,n+1}-\hat{u}^{k,n+1})/\rho\label{update-lz2-WS}. 
\end{align}
\end{subequations}
After $N$ rounds, we set $u^{k+1}=u^{k,N}$ and $e^{k+1}=e^{k,N}$. 

\paragraph{Lazy UL explicit projections.} Similar to {\sf E-AiPOD}, we perform the UL projection every $T\geq 1$ rounds but we formulate it in a different scheme. Specifically, if $k\operatorname{mod}T=0$, we perform the projected SGD in \eqref{WS-x}; otherwise, we update $x^{k+1}$ by SGD via $d_f^k$ defined in \eqref{WS-df}. 
\begin{subequations}\label{17ab}
\begin{align}
    &d_f^k:=\nabla_x f(x^k,y^{k+1};\xi^k)+\nabla_{xy}g(x^k,y^{k+1};\phi^{k})u^{k+1}\label{WS-df}\\
    &x^{k+1}=\operatorname{Proj}_{\mathcal{X}}(x^{k}-\alpha d_f^k)\label{WS-x}
\end{align}
\end{subequations}
The benefits of viewing the projected SGD in this way other than describing it by another loop as in {\sf E-AiPOD} are that: 1) the total number of UL projections is ${\cal O}(K/T)$ rather than ${\cal O}(K)$; and, 2) it is more convenient to incorporate the ML sequences of {\sf E2-AiPOD} in this UL framework to obtain the improved projection complexity, which will be elaborated more in detail in Section \ref{sec:theory-E2}. 

We summarize the {\sf E2-AiPOD} algorithm in Algorithm \ref{stoc-alg-skip-WS} that encompasses \textsf{E-AiPOD}$_{\rm{low}}$ and \textsf{E2-AiPOD}$_{\rm{med}}$ as subroutines.

\subsection{Theoretical analysis}\label{sec:theory-E2}
The recent advances in stochastic approximation with multiple coupled sequences have revealed that one of the keys to establishing the $\tilde{\cal O}(\epsilon^{-2})$ overall sample complexity for multi-sequences is 
the Lipschitz smoothness of the fixed point \citep{shen2022single}. However, this property does not hold for $u^*(x,y)$ unless one makes extra assumption that $\nabla_{yyy}g(x,y)$ and $\nabla_{yyx}g(x,y)$ are Lipschitz continuous like in \citep{dagreou2022framework}. 
Alternatively, we establish $\tilde{\cal O}(\epsilon^{-2})$ sample complexity  for {\sf E2-AiPOD} by the virtue of boundedness of $u^*(x,y)$, which is established in Lemma \ref{ue-bound}. 


To characterize the convergence of {\sf E2-AiPOD}, we use another Lyapunov function defined as
\begin{align}
\mathbb{V}_2^k:= &F(\bar x^k)+\frac{ L_f}{L_r}\left(\|y^k-y^*(\bar x^k)\|^2+\frac{\beta^2}{p^2}\|r^{k}-r^*(\bar x^k)\|^2\right)\label{lyapunov-skip-WS}
\end{align}
where $\bar x^k=\operatorname{Proj}_{\mathcal{X}}(x^k)$ is a virtual   sequence and   $L_f$, $L_r$ are   defined in Lemma \ref{smooth-barf} and \ref{smooth-r} in Appendix.

We also treat the UL $x$-update of {\sf E2-AiPOD} as the biased SGD, but the reference point of $x^k$ has been changed to the virtual point $\bar x^k$ rather than the most-recent projection point $x^k$ in {\sf E-AiPOD} with $k\mod T=0$. The reasons lie in three folds: 1) $\bar x^k$ can be served as a reference point as we can prove that it is close to $x^k$ in the sense $\EE[\|x^k-\bar x^k\|^2]={\cal O}(\alpha^2)$; 2) the virtual point $\bar x^k$ can measure error due to the one-step projection-free deviation at $x^k$; and, 3) the corrections at ML compensate the one-step projection-free errors so that the convergence is provable by using the reference point $\bar x^k$. 

The convergence of {\sf E2-AiPOD} is stated in Theorem \ref{thm:conv-E2} with its proof deferred in Appendix \ref{sec:thm:conv-E2}. 
\begin{theorem}[Convergence of {\sf E2-AiPOD}]\label{thm:conv-E2}
Under Assumption \ref{as1}--\ref{ash}, if we choose stepsizes such that $\beta={\cal O}(\alpha)$, $\beta\leq\frac{1}{\ell_{g,1}}$ and $\rho\leq\min\left\{\frac{1}{\ell_{g,1}},\frac{\mu_g}{4\sigma_{g,2}^2}\right\}$, the probabilities $p={\cal O}(\sqrt{\beta}), q={\cal O}(\sqrt{\rho})$ and the number of ML and LL steps $N={\cal O}(\log(\alpha^{-1})), S={\cal O}(1)$, then the sequences of Algorithm \ref{stoc-alg-skip-WS} satisfy
\begin{align}\label{convergence-E2}
    \frac{1}{K}\sum_{k=0}^{K-1}\EE\left[\|\nabla F(\bar x^k)\|_{P_x}^2\right]\leq \frac{2(\mathbb{V}_2^{0}-F^*)}{\alpha  K}+{\cal O}(\alpha\sigma_{g,1}^2+\alpha^2T^2+\alpha\tilde\sigma_{f,2}^2+\rho\sigma_u^2)
\end{align}
where $\bar x^k=\operatorname{Proj}_{\mathcal{X}}(x^k)$, $P_x=I-B^\dagger B$, $F^*$ is the lower bound of $F(x)$ and $\sigma_{g,1}^2,\sigma_u^2,\tilde\sigma_{f,2}^2$ are the variance of LL, ML, and UL updates. 
\end{theorem}
We can choose $T, \alpha$ and $\rho$ to balance the right hand side of \eqref{convergence-E2},  which leads to the following corollary. 
\begin{corollary}[Sample and projection complexity of {\sf E2-AiPOD}]\label{cor-E2}
Under Assumption \ref{as1}--\ref{ash}, if we choose stepsizes  $\alpha={\cal O}(1/\sqrt{K}), \rho=\beta={\cal O}(1/\sqrt{K})$, probabilities $p={\cal O}(K^{-1/4}), q={\cal O}(K^{-1/4}), T={\cal O}(K^{1/4})$ and the number of steps $N={\cal O}(\log(K)), S={\cal O}(1)$,   the sequences of {\sf E2-AiPOD} satisfy
\begin{align*}
    \frac{1}{K}\sum_{k=0}^{K-1}\EE\left[\|\nabla F(\bar x^k)\|_{P_x}^2\right]=
     {\cal O}\left(\frac{1}{\sqrt{K}}\right).
\end{align*}
As a result, the sample complexity of Algorithm \ref{stoc-alg-skip-WS} is $KNS=\tilde{\cal O}(\epsilon^{-2})$, and the expected number of projections of UL and LL are respectively
\begin{align*}
    \textrm{UL projections}: K/T+qKN=\tilde{\cal O}(K^{\frac{3}{4}})=\tilde{\cal O}(\epsilon^{-1.5}),~~\textrm{LL projections}: pKS={\cal O}(K^{\frac{3}{4}})={\cal O}(\epsilon^{-1.5}). 
\end{align*}
\end{corollary}

Compared with {\sf E-AiPOD}, {\sf E2-AiPOD} can reduce the UL projection complexity with respect to $\epsilon$ with constant ${\cal O}(1)$ LL batch size owing to the corrections in Hessian skipping stages, which makes it more compatible for large-scale bilevel problems.

\section{Applications to Federated Bilevel Learning}\label{sec:fedbi}

Consider the bilevel federated learning problem \citep{tarzanagh2022fednest} in the following form 
\begin{align}\label{eq:fed bilevel}
&\min _{x \in \mathcal{X}} F(x)=\frac{1}{M} \sum_{m=1}^{M} f_{m}\left(x_m, y_m^{*}(x_m)\right) ~~~~\textrm{s.t. }~~~~ y^{*}(x) = \underset{y \in \mathcal{Y}}{\argmin}~ \frac{1}{M} \sum_{m=1}^{M} g_{m}(x_m, y_m)
\end{align}
where each client $m\in[M]:=\{1,\cdots M\}$ maintains its local model $x_m,y_m$ and is only accessible to  its individual function $(f_m,g_m)$. 
Let $x=[x_1,\cdots,x_M]^\top$ and $y=[y_1,\cdots,y_M]^\top$ denote the collection of individual models;  $y^*(x)=[y_1^*(x_1),\cdots,y_M^*(x_M)]^\top$ is the optimal LL model; and let $\mathcal{X}=\left\{x\mid x_1=\cdots=x_M\right\}$ and $\mathcal{Y}=\left\{y\mid y_1=\cdots=y_M\right\}$ denote the consensus set. 

With $\mathbf{I}_{d}\in\mathbb{R}^{d\times d}$ denoting as the identity matrix and $\mathbf{1}_{M}\in\mathbb{R}^{M}$ denoting as the all-$1$ vector, we can define the consensus matrix $A$ and calculate the orthogonal basis of its kernel as
\begin{equation}\label{A-V2}
A:=\left[\begin{array}{cccc}
1&-1&&\\
&\ddots &\ddots& \\
& &1&-1 \\
\end{array}\right]\otimes \mathbf{I}_d,~~~~~{\rm and}~~~~~ V_2:=\frac{\mathbf{1}_M}{\sqrt{M}}\otimes\mathbf{I}_d
\end{equation}
where $\otimes$ is the Kronecker product, $A\in\mathbb{R}^{d(M-1)\times dM}$ and $V_2\in\mathbb{R}^{dM\times d}$. 
We can define $B$ the same as $A$. In the federated bilevel learning setting, with $e=c=h(x)=0$, the UL and LL constraint sets become $\mathcal{X}=\{x\mid Bx=0\}$ and $\mathcal{Y}=\{y\mid Ay=0\}$.

Therefore, in the federated bilevel setting \eqref{eq:fed bilevel}, the UL gradient in \eqref{eq.F_grad} can be specialized as $\nabla F(x)=[\nabla_{x_1}F(x),\cdots,\nabla_{x_M}F(x)]^\top$, where each block $\nabla_{x_m} F(x)$ is defined as
\begin{subequations}
\begin{align}
&\nabla_{x_m} F(x)=\nabla_{x_m} f_m(x_m,y_m^*(x_m))+\frac{\nabla y_m^{*\top}(x_m)}{M}\sum_{m=1}^M\nabla_{y_m}f_m(x_m,y_m^*(x_m))\label{upp-grad}\\
 &{\rm with}~~   \nabla y_m^*(x_m)=-\left(\frac{1}{M}\sum_{m=1}^M\nabla_{yy}g_m(x_m,y_m^*(x_m))\right)^{-1}\nabla_{yx}g_m\left(x_m,y_m^*(x_m)\right).\label{y*-grad}
\end{align}
\end{subequations}
Besides, from \eqref{A-V2}, the projections amount to averaging $x_m$ and $y_m$, i.e.,
\begin{subequations}
\begin{align}
    &\operatorname{Proj}_{\mathcal{X}=\{x\mid Bx=0\}}(x)=(\bar x,\cdots,\bar x), ~~~\textrm{ with }\bar x=\frac{1}{M}\sum_{m=1}^M x_m\label{proj_x_fed}\\
    &\operatorname{Proj}_{\mathcal{Y}=\{y\mid Ay=0\}}(y)=(\bar y,\cdots,\bar y), ~~~\textrm{ with }\bar y=\frac{1}{M}\sum_{m=1}^M y_m\label{proj_y_fed}.
\end{align}
\end{subequations}
Therefore, in the federated bilevel learning setting, the UL and LL projections in {\sf E-AiPOD} and {\sf E2-AiPOD} correspond to communicating the UL and LL variables between clients and the server.  
With the above facts, we are able to apply {\sf E-AiPOD} and {\sf E2-AiPOD} to the federated bilevel setting, which are summarized in Algorithm \ref{stoc-alg-fed} together in Appendix.

\begin{table}[tbp]
\small
    \centering
    \vspace{-0.4cm}
    \begin{tabular}{>{\columncolor{gray!10}}c|>{\columncolor{Lightpurple!60}}c |>{\columncolor{Lightpurple!60}} c|  c}
    \hline\hline
           &E-AiPOD&E2-AiPOD &FedNest\\
        \hline
        $y$-update& Proxskip& Proxskip&SVRG\\
        \hline
        $x$-update& FedAvg& FedAvg &SVRG\\
        \hline
       \! \! allow data heterogeneity & \cmark& \cmark &\cmark\\
        \hline
       \!\!  UL communication&$\tilde{\cal O}(\epsilon^{-2}/T)$&$\tilde{\cal O}(\epsilon^{-1.5})$&$\tilde{\cal O}(\epsilon^{-2})$\\
        \hline
       \!\!  LL communication&$\tilde{\cal O}(\epsilon^{-1.5}/T^{\frac{3}{4}})$&${\cal O}(\epsilon^{-1.5})$&$\tilde{\cal O}(\epsilon^{-2})$\\
        \hline\hline
    \end{tabular}
    \caption{Communication comparison of {\sf E-AiPOD}, {\sf E2-AiPOD} and the state-of-the-art works FedNest in \citep{tarzanagh2022fednest} on stochastic federated bilevel learning to achieve $\epsilon$ stationary point.  } 
    \label{tab:table2}
\end{table}

\subsection{Communication complexity in federated bilevel optimization}
The weighted norm measure \eqref{stationary} in our analysis coincides with the optimality measure in the federated bilevel learning \citep{tarzanagh2022fednest}; see the proof in Appendix \ref{measure-fed}. Moreover, since Assumption \ref{ash} directly holds in federated bilevel setting, our theoretical results are based on the standard assumptions in unconstrained bilevel optimization \citep{hong2020two,ghadimi2018approximation,ji2021bilevel,chen2021closing}. Inheriting from the projection complexity of {\sf E-AiPOD} and {\sf E2-AiPOD}, their communication complexity is stated in the next corollary. 
 
\begin{corollary}[\bf Communication complexity]
Under Assumption \ref{as1}--\ref{as-stoc} and the same condition of Corollary \ref{redu-UL}, the total number of the UL communication of Algorithm \ref{stoc-alg-fed} with option {\sf E-AiPOD} is reduced to $\tilde{\cal O}(\epsilon^{-2}/T)$, while the LL communication is reduced to $\tilde{\cal O}(\epsilon^{-1.5}/T^{\frac{3}{4}})$. Under the same condition of Corollary \ref{cor-E2}, the total number of the UL communication of Algorithm \ref{stoc-alg-fed} with option {\sf E2-AiPOD} is reduced to $\tilde{\cal O}(\epsilon^{-1.5})$, while the LL communication is reduced to ${\cal O}(\epsilon^{-1.5})$. 
\end{corollary}

Compared with the state-of-the-art work FedNest \citep{tarzanagh2022fednest}, the communication complexity of {\sf E-AiPOD} and {\sf E2-AiPOD} for federated learning in Algorithm \ref{stoc-alg-fed}   reduces from $\tilde{\cal O}(\epsilon^{-2})$ to $\tilde{\cal O}(\epsilon^{-1.5})$ even if we use SGD-type updates instead of SVRG-type in FedNest. See a detailed summary in Table \ref{tab:table2}. 

\subsection{Additional related works on bilevel federated learning} 
There is a rich literature in federated learning and recently bilevel federated learning. 
Federated learning and the Federated average (FedAvg) algorithm were first introduced by \citep{mcmahan2017communication}. The convergence rate of FedAvg has been thoroughly investigated by \citep{stich2018local,yu2019parallel,woodworth2020local,yang2020achieving}; see a survey \citep{kairouz2021advances}. Later on, \citep{mitra2021linear} applied variance reduction techniques to tackle the heterogeneous data and obtain the linear convergence for strongly convex objectives. Recently, \citep{mishchenko2022proxskip} has first theoretically achieved the optimal complexity for strongly convex objectives without assuming any data similarity. 

The federated bilevel learning has been first studied in  \citep{tarzanagh2022fednest} that aims to tackle federated learning problems to  with the nested structure.  
The FedNest algorithm has been developed in \citep{tarzanagh2022fednest} that achieves both $\tilde{\cal O}(\epsilon^{-2})$ sample  and communication complexity. The complexity of FedNest has been improved to $\tilde{\cal O}(\epsilon^{-1.5})$ by using momentum-based variance-reduction technique in LocalBSGVRM \citep{gao2022convergence} and FedBiOAcc \citep{li2022local}, but these two works consider different settings than FedNest and our paper. Specifically, FedBiOAcc \citep{li2022local} only considered the UL federated setting. More importantly, the additional bounded data similarity assumption is enforced in  \citep{gao2022convergence,li2022local} that allows to bypass the key challenge of estimating the global Hessian. 
From the algorithmic perspective, LocalBSGVRM  and FedBiOAcc achieve the improved communication complexity partially thanks to the accelerated $\tilde{\cal O}(\epsilon^{-1.5})$ iteration complexity of the non-federated momentum-based bilevel algorithms \citep{khanduri2021near,yang2021provably}, while E-AiPOD and E2-AiPOD are based on SGD-based bilevel algorithms \citep{chen2021closing,ji2021bilevel} with $\tilde{\cal O}(\epsilon^{-2})$ iteration complexity. Therefore, it would be interesting to see in future work if  incorporating momentum acceleration in {\sf E-AiPOD} and {\sf E2-AiPOD} can further reduce communication complexity.

Recent advances in this line have also considered different settings such as decentralized, deterministic or finite-sum settings; see e.g., \citep{lu2022decentralized,yang2022decentralized,gao2022stochastic,chen2022decentralized,lu2022stochastic,huang2022federated}. While it is not the focus here, it is also promising to extend {\sf E2-AiPOD} to the decentralized bilevel setting with the same projection (communication) complexity guarantee by leveraging the decentralized version of Proxskip in \citep{mishchenko2022proxskip} to the LL and ML, and replacing the periodical SGD with decentralized SGD in the UL level.

\begin{figure*}[t]
\vspace*{-0.2cm}
\centering
    \includegraphics[width=0.4\textwidth]{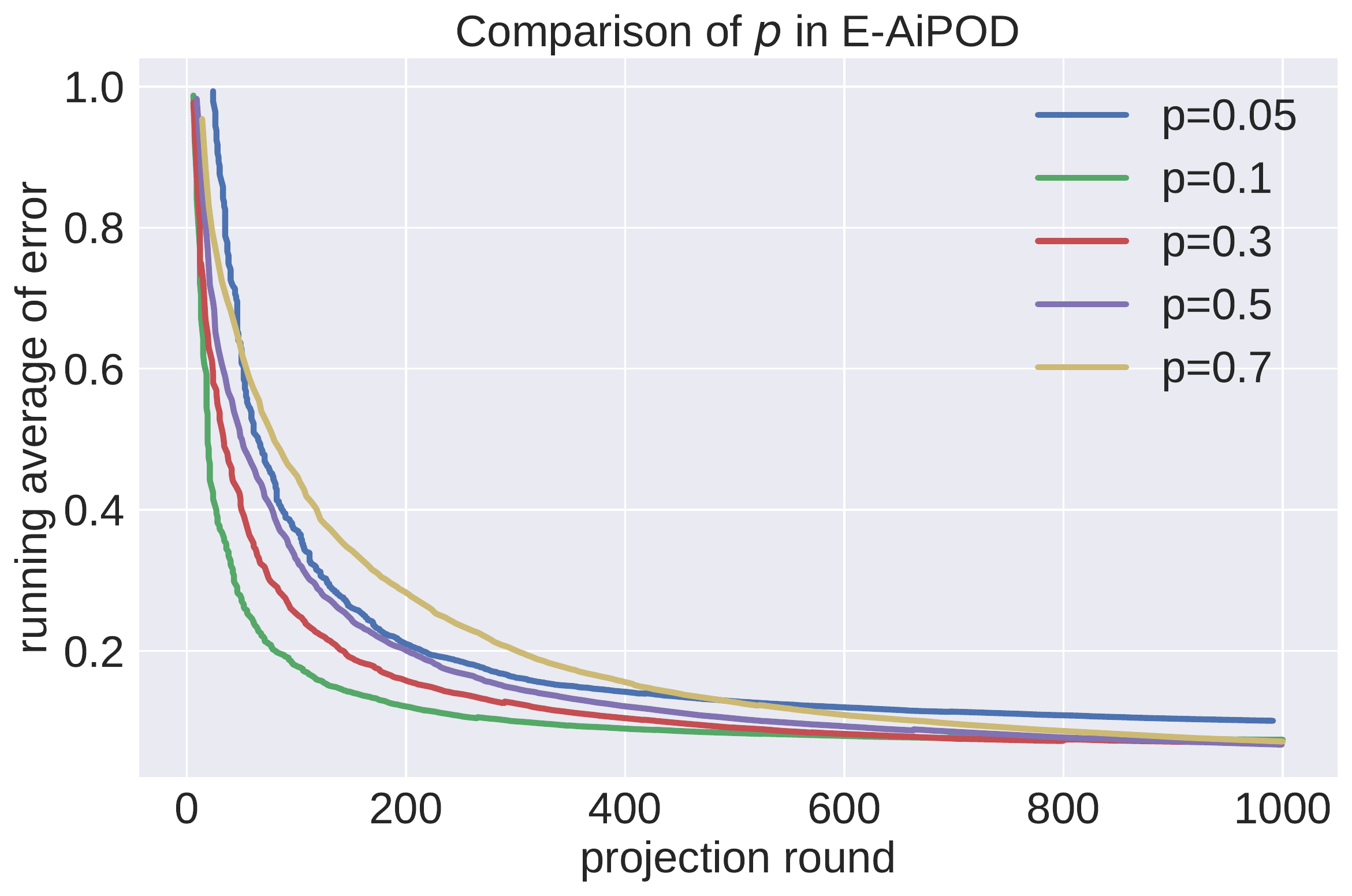}\hspace{0.4cm}
    \includegraphics[width=0.4\textwidth]{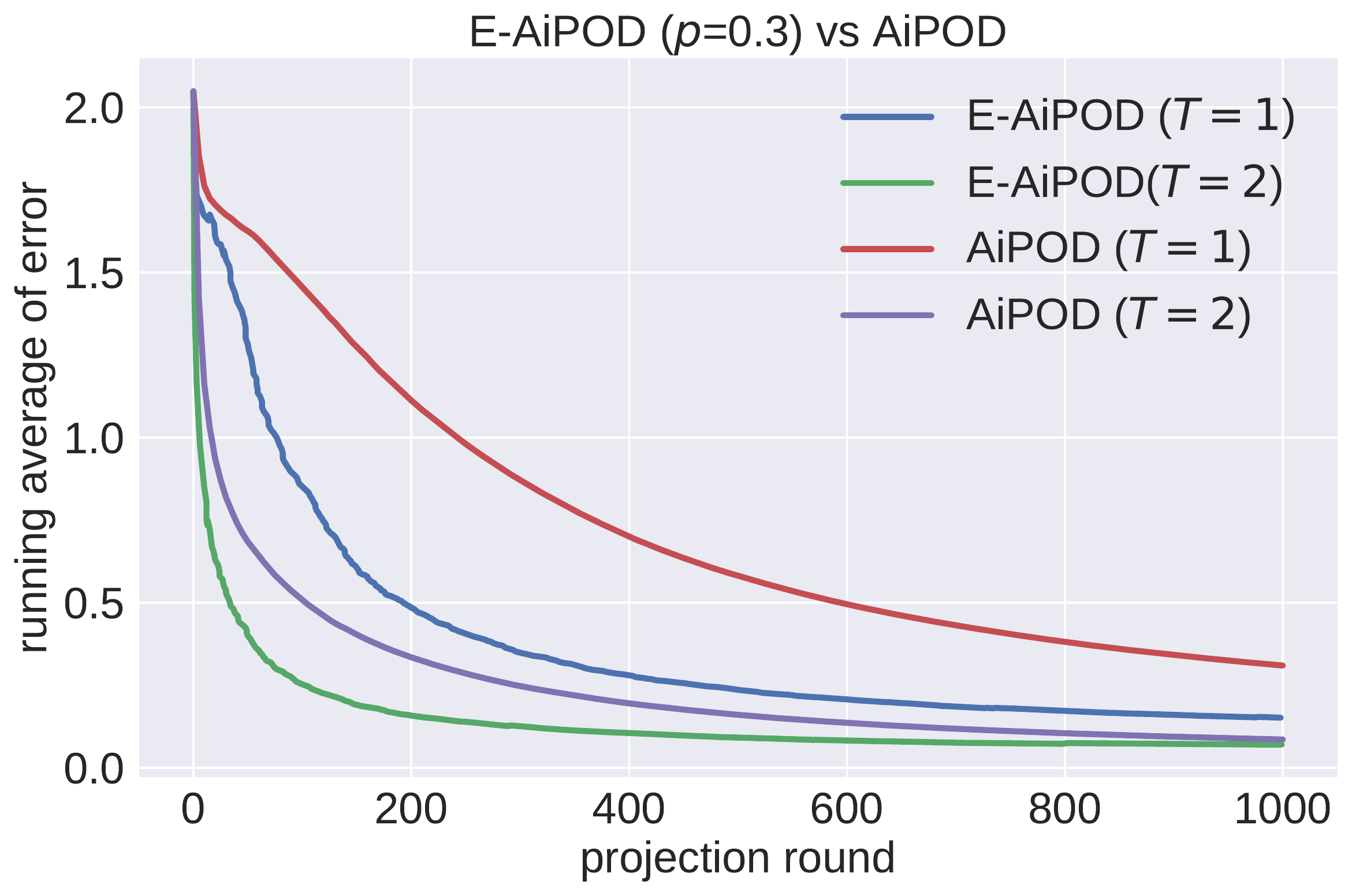}
    \includegraphics[width=0.4\textwidth]{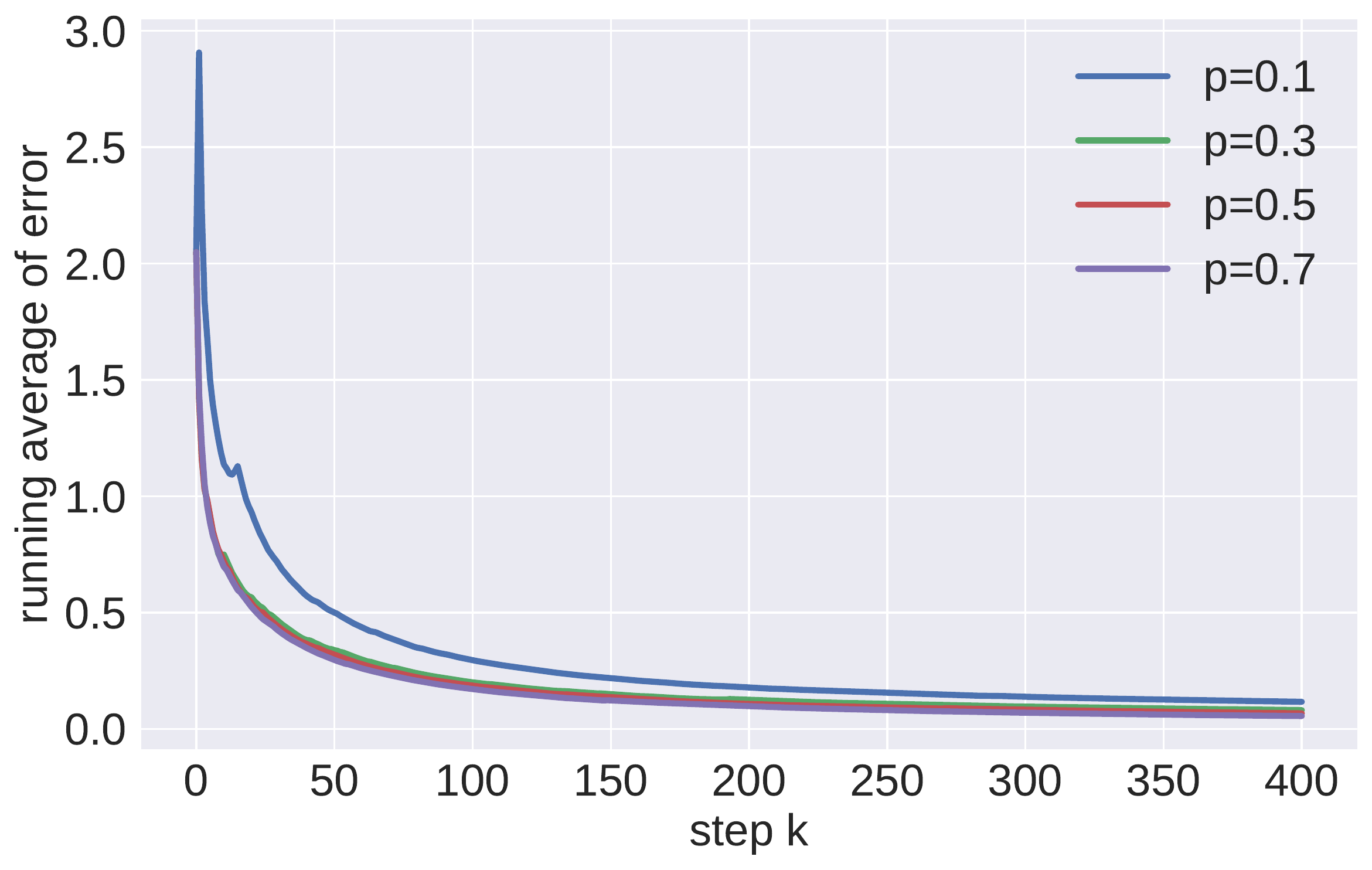}\hspace{0.4cm}
    \includegraphics[width=0.4\textwidth]{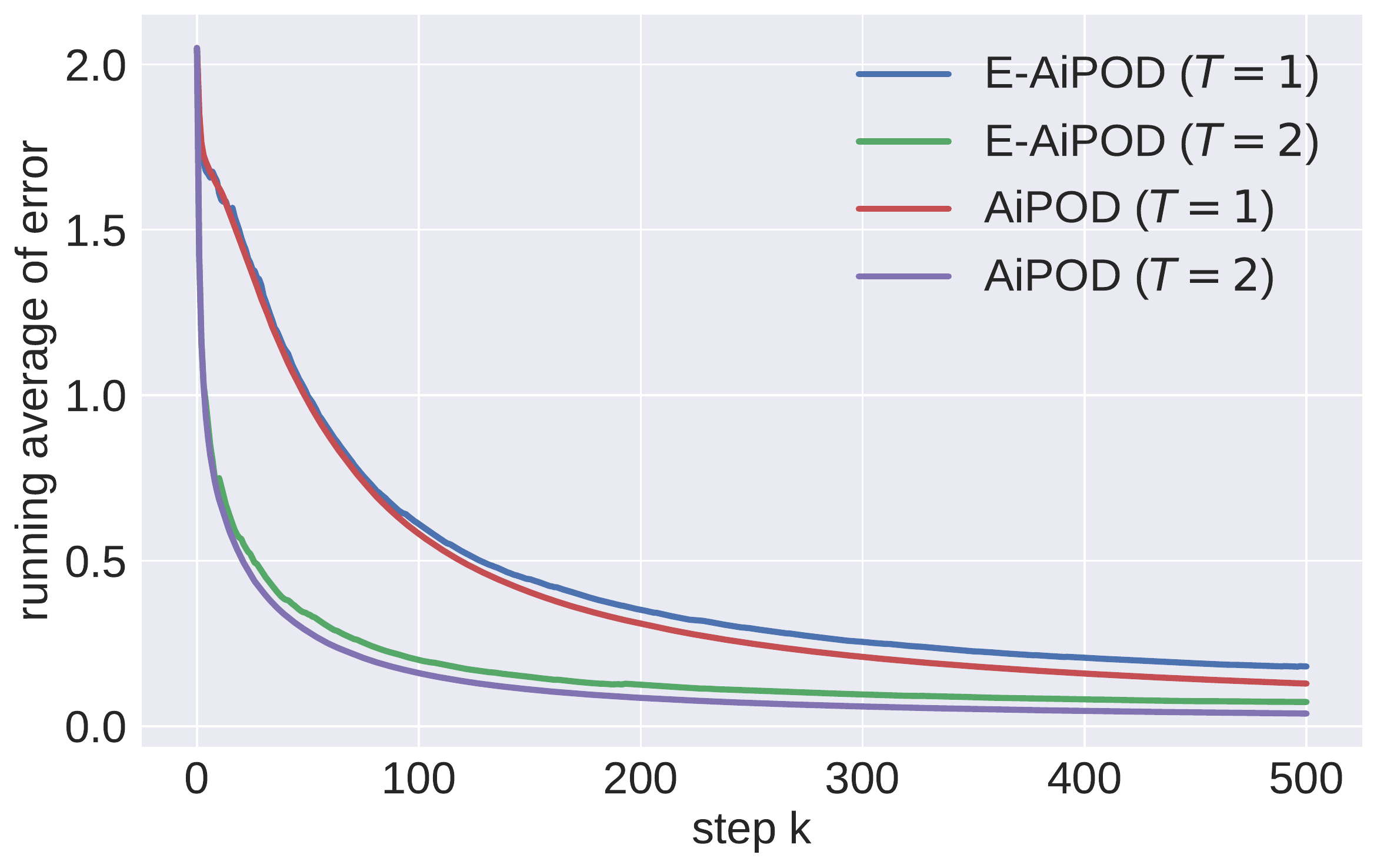}
     \caption{Impact of $p$ in {\sf E-AiPOD} (left) and comparison between  {\sf {\sf AiPOD}} and {\sf E-AiPOD} (right). Here the running average of \textit{error} is defined as $\frac{1}{K}\sum_{k=1}^K(\|\nabla F(x^k)\|_{P_x}^2+\|y^{k+1}-y^*(x^k)\|^2)$.}
    \label{fig:name1}
    \vspace{-0.2cm}
\end{figure*}

\section{Experiments}\label{sec:experiment}
To validate the theoretical results and evaluate the empirical performance of our proposed methods, this section conducts experiments in both synthetic tests and federated bilevel learning tasks including representation learning and learning from imbalanced data. 

\subsection{Synthetic experiments}
We first consider a special case of the equality-constrained bilevel problem \eqref{opt0}, given by 
\begin{align*}
    \min_{x \in \mathcal{X}}F(x)=\sin{(c^\top x+d^\top y^*(x))}+\ln{(\|x+y^*(x)\|^2+1)},~{\rm s. t.}~y^*(x)=\argmin_{y\in\mathcal{Y}(x)}\frac{1}{2}\|x-y\|^2
\end{align*}
where $\mathcal{X}=\{x~|~Bx=0\}\subset\mathbb{R}^{100}$, ${\cal Y}(x)=\{y~|~Ay+H x=0\}\subset\mathbb{R}^{100}$, and $A,B,H,c,d$ are randomly generated non-zero matrices or vectors that satisfy Assumption \ref{ash}. To guarantee that $\mathcal{Y}(x)$ and $\mathcal{X}$ are not singleton, the matrixes $A$ and $B$ are rank-deficient matrices. 
In the simulation, we use the noisy versions of the gradients where a Gaussian noise with zero mean and a standard deviation of $0.1$ is added. It can then be checked that in this setting, Assumptions \ref{as1}--\ref{ash} are satisfied.

The test results are reported in Figure \ref{fig:name1}. In the left figure, we test the impact of the probability $p$ on the projection complexity and the iteration complexity. It can be observed that {\sf E-AiPOD} with relatively small $p$ has almost the same iteration complexity (as indicated in the lower left figure) while it significantly saves projection rounds (see upper left figure). We also compare {\sf AiPOD} and {\sf E-AiPOD} in the right figure. It can be observed that {\sf E-AiPOD} is able to save projection while maintaining the same iteration complexity as that of {\sf AiPOD}, which is consistent with our theoretical result.

\subsection{Federated representation learning}
In this section, we apply {\sf E-AiPOD} in Algorithm \ref{stoc-alg-fed} to the federated representation learning task. The classic machine learning approach learns a data representation and a downstream header jointly on the training data set. While the bilevel representation learning \citep{franceschi2018bilevel} seeks to learn a data representation on the validation set and a header on the training data set, the procedure can then be formulated as a bilevel problem. In a federated representation learning setting with $M=50$ clients, the validation and training data sets are distributed among clients, and the goal is to learn a representation and header respectively on the joint validation and training data set while protecting data privacy. 

Formally, the problem can be formulated as an instance of \eqref{eq:fed bilevel}, given by
\begin{align}
\min_{x\in\mathcal{X}} \frac{1}{M}\sum_{m=1}^M f_{\rm ce}(x_m,y_m^*(x);\mathcal{D}_{\rm val}^m),~~{\rm s. t.}~~y^*(x)=\arg\min_{y \in \mathcal{Y}}\frac{1}{M}\sum_{m=1}^M f_{\rm ce}(x,y_m;\mathcal{D}_{\rm tr}^m)+0.05\|y_m\|^2,\nonumber
\end{align}
where $x$ is the parameters of the representation layer; $y$ is the parameter of the classifier layer; $\mathcal{D}_{\rm tr}^m$ and $\mathcal{D}_{\rm val}^m$ are, respectively, the training and validation set of client $m$; $\mathcal{X}$ and $\mathcal{Y}$ are the consensus sets defined in \eqref{eq:fed bilevel}. The cross-entropy loss $f_{\rm ce}$ is defined as
\begin{align*}
    f_{\rm ce}(x,y; \mathcal{D})\coloneqq -\frac{1}{|\mathcal{D}|}\sum_{d_n\in \mathcal{D}}\log \frac{\exp{\big(h_{l_n}(x,y;d_n)}\big)}{\sum_{c=1}^C \exp{(h_c\big(x,y;d_n)\big)} }
\end{align*}
where $C$ is the number of classes, $d_n$ is the $n$-th data from class $l_n$ in data set $\mathcal{D}$ and $h(x,y;d_n)=[h_1(x,y;d_n),...,h_C(x,y;d_n)]^\top\!\in\!\mathbb{R}^{C}$ is the output of the model with parameter $(x,y)$ and input $d_n$.

\begin{figure*}[t]
\centering
    \includegraphics[width=0.4\textwidth]{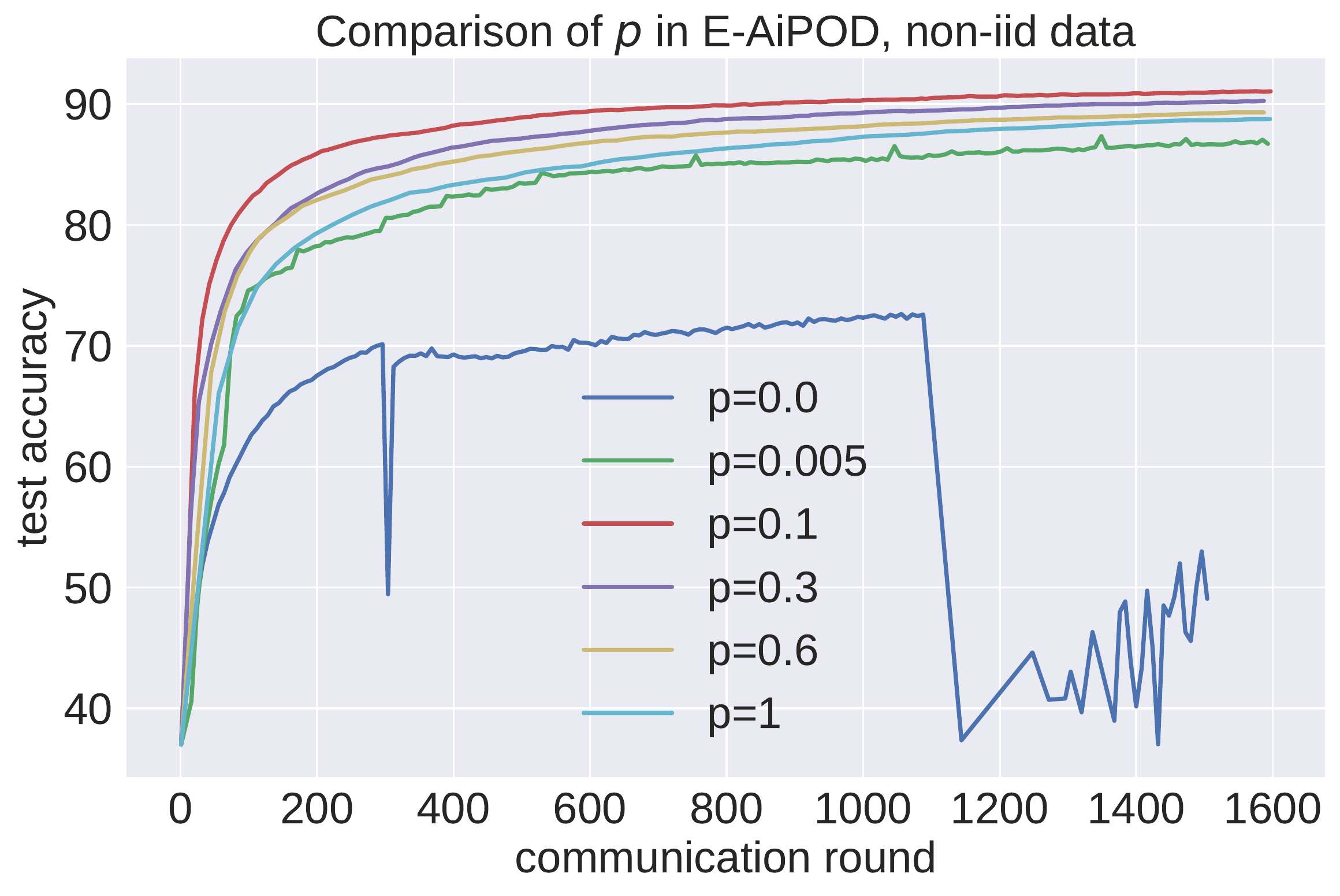}\hspace{0.4cm}
    \includegraphics[width=0.4\textwidth]{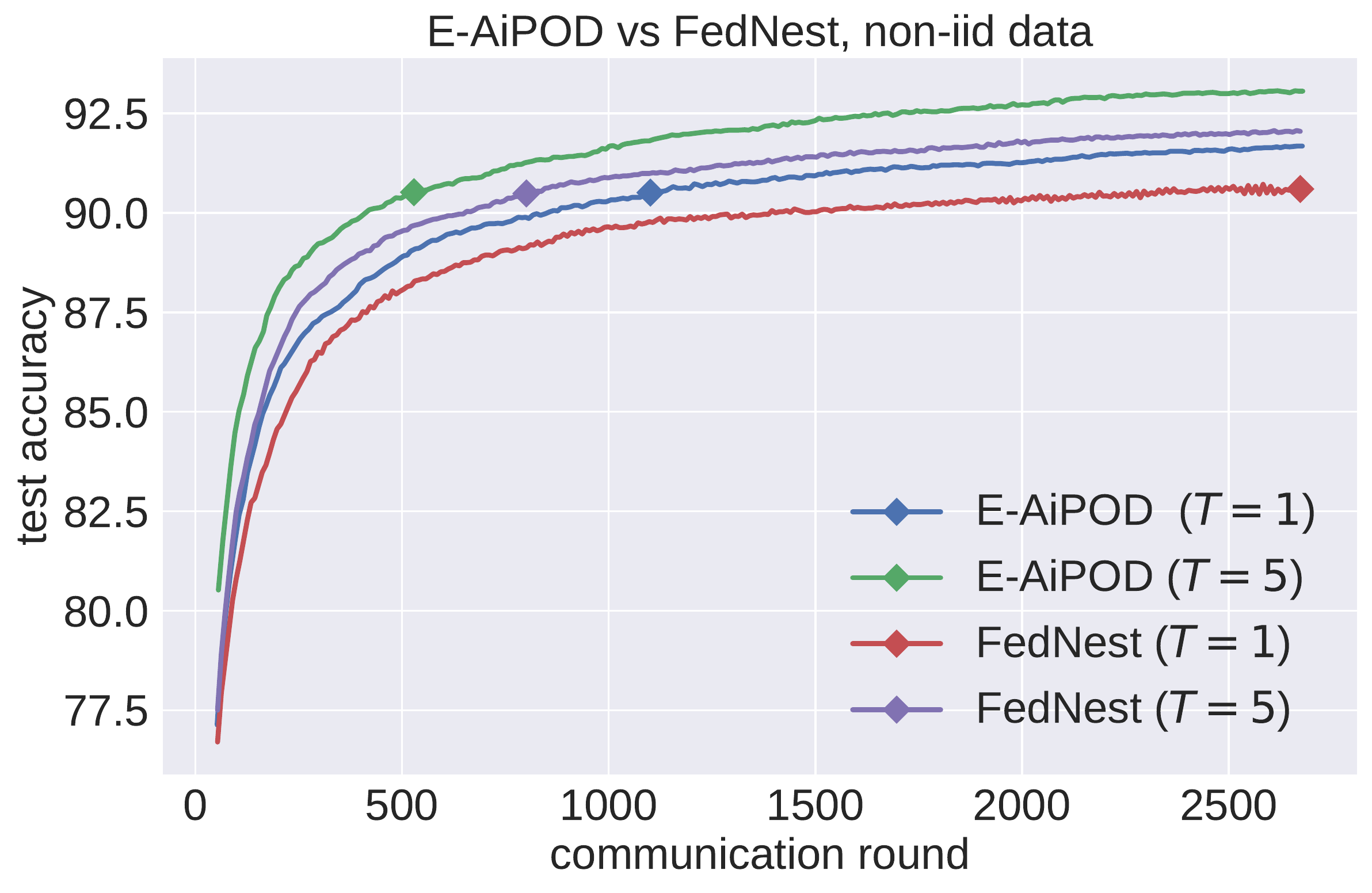}
     \caption{Federated hyper-representation learning: Impact of communication probability $p$ (left) and comparison of our algorithm with FedNest \citep{tarzanagh2022fednest} (right). The experiments are run on the MNIST dataset distributed among clients in a non-i.i.d. fashion.}
    \label{fig:name2}
\end{figure*}

The experimental results are reported in Figure \ref{fig:name2}. From the left figure of Figure \ref{fig:name2}, a relatively small value of $p$ helps save communication rounds while a too small value of $p$ might degrade performance. With a properly chosen $p=0.1$, it can be observed from the right figure that {\sf E-AiPOD} outperforms FedNest \citep{tarzanagh2022fednest} in terms of communication complexity.

\begin{figure*}[t]
\centering
    \includegraphics[width=0.4\textwidth]{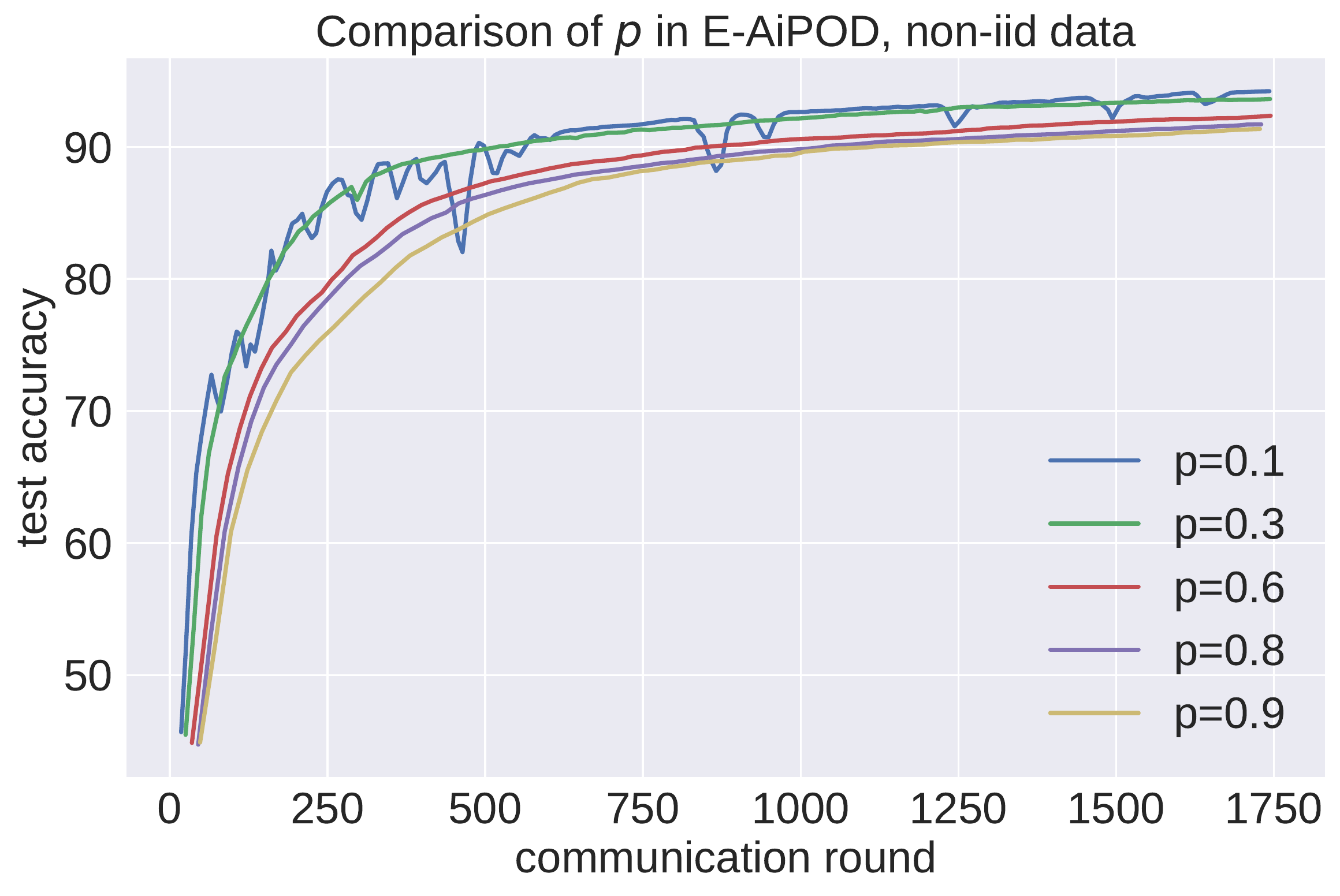}\hspace{0.4cm}
    \includegraphics[width=0.4\textwidth]{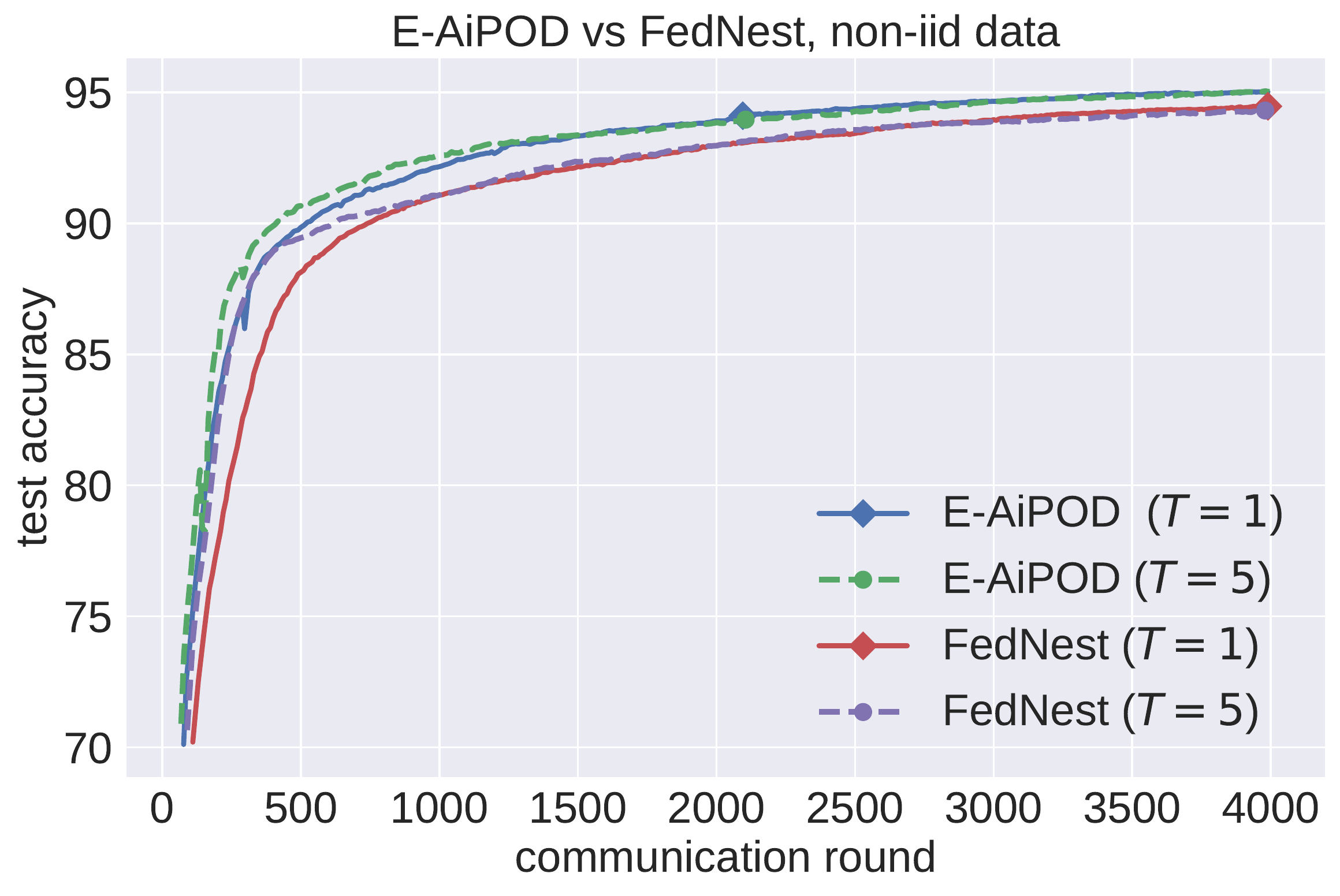}
     \caption{Federated learning from imbalanced data: Impact of communication probability $p$ (left) and comparison of our algorithm with FedNest \citep{tarzanagh2022fednest} (right). The experiments are run on an imbalanced MNIST dataset distributed among clients in a non-i.i.d. fashion.}
    \label{fig:name3}
\end{figure*}

\subsection{Federated learning from imbalanced data}
In this subsection, we apply {\sf E-AiPOD} to the federated learning from the imbalanced data task, where the goal is to learn a good model that  guarantees both fairness and generalization  from datasets with under-represented classes \citep{Li2021autobalance}. 
In the UL, the loss-tuning parameters are trained to improve generalization and fairness, while the model parameters are trained on a possibly imbalanced data-set in the LL. The method was later extended to the federated setting in \citep{tarzanagh2022fednest}. Formally, the problem can be written as a case of \eqref{eq:fed bilevel}, given by
\begin{align}
    \min_{x\in\mathcal{X}} \frac{1}{M}\sum_{m=1}^M f^{\rm up}_{\rm vs}(x_m,y_m^*(x);\mathcal{D}_{\rm val}^m),~~{\rm s. t.}~~y^*(x)=\arg\min_{y \in \mathcal{Y}}\frac{1}{M}\sum_{m=1}^M f^{\rm low}_{\rm vs}(x,y_m;\mathcal{D}_{\rm tr}^m),\nonumber
\end{align}
where the number of clients is $M=50$, $x$ is the loss-tuning parameters and $y$ is  the neural network parameter. Here $\mathcal{D}_{\rm tr}^m$ and $\mathcal{D}_{\rm val}^m$ are respectively the training and validation set of client $m$ and $\mathcal{X},\mathcal{Y}$ are the consensus sets defined in \eqref{eq:fed bilevel}. From \citep{kini2021labelimba}, the so-called vector-scaling loss $f^{\rm low}_{\rm vs}$ is defined as 
\begin{align*}
    f^{\rm low}_{\rm vs}(x,y;\mathcal{D})\coloneqq -\frac{1}{|\mathcal{D}|}\sum_{d_n\in \mathcal{D}} \omega_{l_n}\log \frac{\exp{(\delta_{l_n} h_{l_n}(y;d_n)+\tau_{l_n}})}{\sum_{c=1}^C \exp{(\delta_{c} h_c(y;d_n)}+\tau_{c} )}
\end{align*}
where $N$ is the data set size, $C$ is the number of classes, $d_n$ is the $n$-th data with label class $l_n$ in data set $\mathcal{D}$ and $h(y;d_n)=[h_1(y;d_n),...,h_C(y;d_n)]^\top \in \mathbb{R}^{C}$ is the logit output of the neural network with parameter $y$ and input $d_n$. Define $x=(\omega,\delta,\tau)$ where $\omega \coloneqq [\omega_1,...,\omega_C]^\top \in \mathbb{R}^C$ and $\delta,\tau$ can be defined similarly. The upper-level loss $f^{\rm up}_{\rm vs}$ is a special case of $f^{\rm low}_{\rm vs}$ with $\delta=1$ and $\tau=0$. 

The experimental results are reported in Figure \ref{fig:name3}. From the left figure of Figure \ref{fig:name3}, a relatively small value of $p$ helps save communication rounds. With $p=0.3$, it can be observed from the right figure that {\sf E-AiPOD} outperforms FedNest in terms of communication complexity. A larger $T$ results in faster convergence at the start, but is ultimately equaled by $T=1$.

\section{Conclusions}
In this paper, we established the first finite-time convergence of alternating implicit projected SGD algorithm (AiPOD) for equality-constrained bilevel problems which matches the state-of-the-art result in unconstrained bilevel setting and also its single-level projected SGD. Besides, we propose two  projection-efficient variants {\sf E-AiPOD} and {\sf E2-AiPOD} for the setting where evaluating projection is costly. 
E-AiPOD enjoys the $\tilde{\cal O}(\epsilon^{-2}/T)$  upper-level and $\tilde{\cal O}(\epsilon^{-1.5}/T^{\frac{3}{4}})$ lower-level projection complexity with ${\cal O}(T)$ lower-level batch size, and {\sf E2-AiPOD} enjoys $\tilde{\cal O}(\epsilon^{-1.5})$ upper-level and lower-level projection complexity with ${\cal O}(1)$ batch size. We apply {\sf E-AiPOD} and {\sf E2-AiPOD} on federated bilevel settings and achieve the reduction of communication complexity over the state-of-the-art work. Extensive experiments on numerical examples, federated representative learning, and federated learning from imbalanced data verify our theoretical results and demonstrate the effectiveness of our methods.

\bibliography{bilevel}
 \clearpage

\appendix
\begin{center}
{\Large \bf Appendix for
``\FullTitle"}
\end{center}
\vspace{-1.5cm}
\addcontentsline{toc}{section}{} 
\part{} 
\parttoc 

\section{Preliminaries}\label{sec:app-pre}
\subsection{Proof of Lemmas \ref{eq:measure}--\ref{y-smooth}}\label{sec:app-pre2}
\begin{definition}
Suppose $\mathcal{L}:\mathbb{R}^{d_1}\rightarrow\mathbb{R}^{d_2}$ such that each of its first-order partial derivatives exist on $\mathbb{R}^d$, then its Jacobian is defined as
\begin{align}
    \nabla\mathcal{L}=\left[\begin{array}{ccc}
\frac{\partial \mathcal{L}_{1}}{\partial x_{1}} & \cdots & \frac{\partial \mathcal{L}_{1}}{\partial x_{d_1}} \\
\vdots & \ddots & \vdots \\
\frac{\partial \mathcal{L}_{d_2}}{\partial x_{1}} & \cdots & \frac{\partial \mathcal{L}_{d_2}}{\partial x_{d_1}}
\end{array}\right].
\end{align}
Therefore, $\nabla h(x)$ and $\nabla y^*(x)$ can be written as
\begin{align}\label{hy*}
    \nabla h(x)=\left[\begin{array}{ccc}
\frac{\partial h_{1}}{\partial x_{1}} & \cdots & \frac{\partial \mathcal{L}_{1}}{\partial x_{d_x}} \\
\vdots & \ddots & \vdots \\
\frac{\partial \mathcal{L}_{m_y}}{\partial x_{1}} & \cdots & \frac{\partial \mathcal{L}_{m_y}}{\partial x_{d_x}}
\end{array}\right],~~~~~\nabla y^*(x)=\left[\begin{array}{ccc}
\frac{\partial y_1^*(x)}{\partial x_{1}} & \cdots & \frac{\partial y_1^*(x)}{\partial x_{d_x}} \\
\vdots & \ddots & \vdots \\
\frac{\partial y_{d_y}^*(x)}{\partial x_{1}} & \cdots & \frac{\partial y_{d_y}^*(x)}{\partial x_{d_x}}
\end{array}\right].
\end{align}
\end{definition}
\begin{definition}
The Bregman divergence of a differentiable function $\mathcal{L}:\mathbb{R}^d\rightarrow\mathbb{R}$ is defined as 
\begin{align*}
    D_{\mathcal{L}}(u,v):=\mathcal{L}(u)-\mathcal{L}(v)-\langle\nabla \mathcal{L}(v), u-v\rangle.
\end{align*}
\end{definition}
For an $L$-smooth and $\mu$-strongly convex function $\mathcal{L}$, we have
\begin{align}\label{point_diver}
    \frac{\mu}{2}\|u-v\|^2\leq D_{\mathcal{L}}(u,v)\leq \frac{L}{2}\|u-v\|^2
\end{align}
and 
\begin{align}\label{grad_diver}
    \frac{1}{2L}\|\nabla\mathcal{L}(u)-\nabla\mathcal{L}(v)\|^2\leq D_{\mathcal{L}}(u,v)\leq \frac{1}{2\mu}\|\nabla\mathcal{L}(u)-\nabla\mathcal{L}(v)\|^2.
\end{align}
Moreover, for any $u,v$, we have $\langle\nabla \mathcal{L}(u)-\nabla \mathcal{L}(v), u-v\rangle=D_{\mathcal{L}}(u,v)+D_{\mathcal{L}}(v,u)$. 

\begin{lemma}[{\citep[Proposition 4.3]{de2020matrix}}]\label{lm:matrix} Consider random matrices $X$ and $Y$ of the same size that satisfies $\EE[Y|X]=0$, then it holds
$\EE[\|X+Y\|^2]\leq\EE[\|X\|^2]+\EE[\|Y\|^2]$. 
\end{lemma}


\begin{lemma}[\bf Closed form linear operator of projection]\label{lm:proj_closed}
For any nonempty linear space $C=\{z\mid Az+b=0\}$, the projection operator has the following closed form
\begin{align}\label{closed-form-proj}
    \operatorname{Proj}_C(x)=(I-A^\dagger A)x-A^\dagger b
\end{align}
where $A^\dagger$ is the Moore-Penrose inverse of $A$. 
\end{lemma}
\begin{proof}
\textbf{Case 1.} We first consider the case where $b=0$.

We denote $P=I-A^\dagger A$, then $C=\operatorname{Ker}(A)$. According to Proposition 3.3. in \citep{barata2012moore}, we know that
$P$ is an orthogonal projection and
\begin{align*}
    C=\operatorname{Ker}(A)=\operatorname{Ran}(P), ~~C^\perp=\operatorname{Ker}(A)^\perp=\operatorname{Ran}(A^\dagger)\stackrel{(a)}{=}\operatorname{Ran}(A^\dagger A)=\operatorname{Ran}(I-P)
\end{align*}
where $(a)$ holds since $A^\dagger=A^\dagger A A^\dagger$ \citep{barata2012moore} and 
\begin{align*}
    &\forall z=A^\dagger Aw, z=A^\dagger (Aw)\Rightarrow \operatorname{Ran}(A^\dagger A)\subset\operatorname{Ran}(A^\dagger),\\
    &\forall z=A^\dagger w, z=A^\dagger A(A^\dagger w)\Rightarrow\operatorname{Ran}(A^\dagger)\subset\operatorname{Ran}(A^\dagger A).
\end{align*}
Thus, for any $x$, we can write $x=Px+(I-P)x$ and $Px\perp (I-P)x$, which means $Px$ is the orthogonal projection. Then due to the uniqueness of the orthogonal decomposition, $\operatorname{Proj}_C(x)=Px$. 

\textbf{Case 2. } We then consider the case when $b\neq 0$. 

For any $z\in C$, we have $z+A^\dagger b\in\operatorname{Ker}(A)$ since
\begin{align*}
    A(z+A^\dagger b)=Az+AA^\dagger b=-b+AA^\dagger b\stackrel{(a)}{=}0
\end{align*}
where $(a)$ holds since $C$ is nonempty if and only if $AA^\dagger b=b$ \citep{james1978generalised}. Similarly, we can prove for any $z\in\operatorname{Ker}(A)$, $z-A^\dagger b\in C$. Thus, $\operatorname{Ker}(A)=C+A^\dagger b$. 

Moreover, since projection operator minimizes the distance to set $C$, we have
\begin{align*}
    \operatorname{Proj}_C(x)&=\argmin_{z\in C}\|z-x\|^2=\argmin_{z\in C}\|z+A^\dagger b-(x+A^\dagger b)\|^2\\
    &\stackrel{(a)}{=}\left\{\argmin_{w\in \operatorname{Ker}(A)}\|w-(x+A^\dagger b)\|^2\right\}-A^\dagger b\\
    &=\operatorname{Proj}_{\operatorname{Ker}(A)}(x+A^\dagger b)-A^\dagger b\\
    &\stackrel{(b)}{=}P(x+A^\dagger b)-A^\dagger b\\
    &\stackrel{(c)}{=}(I-A^\dagger A)x-A^\dagger b
\end{align*}
where $(a)$ is due to $\operatorname{Ker}(A)=C+A^\dagger b$, $(b)$ comes from Case 1, and $(c)$ is derived from the definition of $P$ and $(I-A^\dagger A)A^\dagger b=(A^\dagger-A^\dagger AA^\dagger)b=0$. 
\end{proof}

By using the special property of projection on linear space, we prove Lemma \ref{eq:measure} next. 
\begin{customlm}{1}
[\bf Stationarity measure under linear equality constraints]
For any $x\in\mathcal{X}$ and any $\lambda>0$, it holds that
\begin{align*}
    \|\nabla F(x)\|^2_{P_x}=\|\lambda^{-1}( x-\operatorname{Proj}_{\mathcal{X}}( x-\lambda\nabla F( x)))\|^2
\end{align*}
where $P_x=I-B^\dagger B$. 
\end{customlm}
\begin{proof}
For any $\lambda$, according to \eqref{closed-form-proj}, we have
\begin{align*}
    \lambda^{-1}( x-\operatorname{Proj}_{\mathcal{X}}( x-\lambda\nabla F( x)))&=\lambda^{-1}( x-(I-B^\dagger B)( x-\lambda\nabla F( x))-B^\dagger e)\\
    &=\lambda^{-1}( x+(I-B^\dagger B)\lambda\nabla F( x)-[(I-B^\dagger B)x+B^\dagger e])\\
    &=\lambda^{-1}( x-\operatorname{Proj}_{\mathcal X}(x))+(I-B^\dagger B)\nabla F(x)\\
    &=(I-B^\dagger B)\nabla F(x)
\end{align*}
where the first equality is due to $\mathcal{X}=\{x|Bx=e\}$ and \eqref{closed-form-proj} the last two equality holds since $x\in\mathcal{X}$ and \eqref{closed-form-proj}. Then with $(I-B^\dagger B)^2=I-B^\dagger B$ and the definition of $\|\cdot\|_{P_x}$, the proof is complete. 
\end{proof}

We restate Lemma \ref{y-imgrad} and Lemma \ref{y-smooth} together in a more formal way. 
\begin{customlm}{2-3}\label{y-smooth-app}
Under Assumption \ref{as1}--\ref{as2} and \ref{ash}, the gradient of $y^*(x)$ can be expressed as
\begin{equation}\small
    \nabla y^*(x)=-V_2(V_2^\top \nabla_{yy}g(x,y^*(x))V_2)^{-1}V_2^\top\left(\nabla_{yx}g(x,y^*(x))-\nabla_{yy}g(x,y^*(x)) A^\dagger \nabla h(x)\right)-A^\dagger\nabla h(x)
\end{equation}
where $V_2$ is the orthogonal basis of $\operatorname{Ker}(A):=\{y\mid Ay=0\}$. Moreover, $y^*(x)$ is $L_y$ Lipschitz continuous and $L_{yx}$ smooth with
\begin{align*}
    &L_y:=\frac{\ell_{g,1}+(\ell_{g,1}+\mu_g)\|A^\dagger\|\ell_{h,0}}{\mu_g},\\
    &L_{yx}:=\frac{\ell_{g,2}\left(1+\|A^\dagger\|\ell_{h,0}\right)(1+L_y)(1+\frac{\ell_{g,1}}{\mu_g})+(\ell_{g,1}+\mu_g)\|A^\dagger\|\ell_{h,1}}{\mu_g}.\numberthis\label{constant-app}
\end{align*}
\end{customlm}
\begin{proof}\allowdisplaybreaks
By singular value decomposition, we can decompose $A=U\Sigma V^\top $ with $\Sigma=\left[\begin{array}{ll}
\Sigma_1 & 0 \\
0 & 0 
\end{array}\right]\in\mathbb{R}^{m_y\times d_y}$, and orthogonal matrix  $U=[U_1~ U_2]\in\mathbb{R}^{m_y\times m_y}$ and $V=[V_1~ V_2]\in\mathbb{R}^{d_y\times d_y}$. Also, by assuming $\operatorname{Rank}(A)=r$, we know that $U_1\in\mathbb{R}^{m_y\times r}, V_1\in\mathbb{R}^{d_y\times r}$ and $\Sigma_1\in\mathbb{R}^{r\times r}$ are full rank submatrix. Therefore, $A$ can be decomposed by
\begin{align*}
    A=[U_1~U_2]\left[\begin{array}{ll}
\Sigma_1 & 0 \\
0 & 0 
\end{array}\right]\left[\begin{array}{l}
V_1^\top \\
V_2^\top  
\end{array}\right]=[U_1\Sigma_1~~~0]\left[\begin{array}{l}
V_1^\top \\
V_2^\top  
\end{array}\right]=U_1\Sigma_1V_1^\top 
\end{align*}
and $V_2$ is the orthogonal basis of $\operatorname{Ker}(A)$. 

Next, if we define $y_0(x):=A^\dagger(c-h(x))$, we can prove  
$
    y_0(x)\in\mathcal{Y}(x)
$
since 
\begin{align*}
    Ay_0(x)=A A^\dagger(c-h(x))=c-h(x)
\end{align*}
where the last equality holds due to Assumption \ref{ash}. From the definition of $y_0(x)$ and \eqref{hy*}, we know 
\begin{align}\label{y0_grad}
    \nabla_x y_0(x)=-A^\dagger \nabla h(x).
\end{align}


Moreover, since $V_2$ is the orthogonal basis of $\operatorname{Ker}(A)$, we know $\mathcal{Y}(x)=y_0(x)+\operatorname{Ran}(V_2)$. Thus, let $z^*(x)=\arg\min_{z}g(x,y_0(x)+V_2z)$, then we have $y^*(x)=y_0(x)+V_2z^*(x)$. 

Since $z^*(x)$ satisfies $$\nabla_z g(x,y_0(x)+V_2z^*(x))=V_2^\top \nabla_y g(x,y_0(x)+V_2z^*(x))=0$$ 
then taking the gradient with respect to $x$ of both sides, we get
\begin{align*}
    0&=\nabla_x(V_2^\top\nabla_y g(x,y_0(x)+V_2z^*(x)))\\
    &=\nabla_{xy}g(x,y_0(x)+V_2z^*(x))V_2+\left(\nabla_x z^*(x)^\top V_2^\top+\nabla_x y_0(x)^\top\right)\nabla_{yy}g(x,y_0(x)+V_2z^*(x))V_2\\
    &=\nabla_{xy}g(x,y_0(x)+V_2z^*(x))V_2+\left(\nabla_x z^*(x)^\top V_2^\top-\nabla^\top h(x)A^{\dagger \top}\right)\nabla_{yy}g(x,y_0(x)+V_2z^*(x))V_2\\
    &=\nabla_{xy}g(x,y^*(x))V_2+\left(\nabla_x z^*(x)^\top V_2^\top-\nabla^\top h(x)A^{\dagger \top}\right)\nabla_{yy}g(x,y^*(x))V_2\numberthis\label{deriv}
\end{align*}
where the third equality holds from \eqref{y0_grad}. Then, rearranging \eqref{deriv}, we get 
\begin{equation}
  \nabla z^*(x)=-\left(V_2^\top\nabla_{yy}g(x,y^*(x))V_2\right)^{-1}V_2^\top\left(\nabla_{yx}g(x,y^*(x))-\nabla_{yy}g(x,y^*(x)) A^\dagger \nabla h(x)\right)  
\end{equation}
and as a result of $y^*(x)=y_0(x)+V_2z^*(x)$, we have 
\begin{align*}
    \nabla y^*(x)&=\nabla y_0(x)+V_2\nabla z^*(x)\\
    &=-A^\dagger\nabla h(x)-V_2\left(V_2^\top\nabla_{yy}g(x,y^*(x))V_2\right)^{-1}V_2^\top\left(\nabla_{yx}g(x,y^*(x))-\nabla_{yy}g(x,y^*(x)) A^\dagger \nabla h(x)\right)\numberthis\label{grad_form_y*}.
\end{align*}

Next, utilizing the fact that $V_2$ is the orthogonal matrix, we know $\mu_g I_{d_y-r}\preceq V_2^\top\nabla_{yy}g(x,y)V_2$. Therefore, we have for any $x,y$, 
\begin{align}\label{bound_v}
    V_2\left(V_2^\top\nabla_{yy}g(x,y)V_2\right)^{-1}V_2^\top\preceq\frac{1}{\mu_g} I.
\end{align}

Besides, we have for any $x$, it follows that
\begin{align*}
    \|\nabla_{yx}g(x,y^*(x))-\nabla_{yy}g(x,y^*(x)) A^\dagger \nabla h(x)\|&\leq(1+\|A^\dagger\|\ell_{h,0})\|\nabla^2 g(x,y^*(x))\|\\
    &\leq (1+\|A^\dagger\|\ell_{h,0})\ell_{g,1}.\numberthis\label{bound-jac}
\end{align*}

As a result of \eqref{grad_form_y*}, \eqref{bound_v} and \eqref{bound-jac}, $\nabla y^*(x)$ is bounded by
\begin{align*}
    &~~~~~\|\nabla y^*(x)\|\\
    &\leq \Big\|V_2\left(V_2^\top\nabla_{yy}g(x,y^*(x))V_2\right)^{-1}V_2^\top\Big\|\|\nabla_{yx}g(x,y^*(x))-\nabla_{yy}g(x,y^*(x)) A^\dagger \nabla h(x)\|+\|A^\dagger\nabla h(x)\|\\
    &\leq \frac{\ell_{g,1}+(\ell_{g,1}+\mu_g)\|A^\dagger\|\ell_{h,0}}{\mu_g}=L_y
\end{align*}
which implies $y^*(x)$ is $L_y$ Lipschitz continuous. 

Finally, we aim to prove the smoothness of $y^*(x)$. Defining $B_1=V_2^\top\nabla_{yy}g(x_1,y^*(x_1))V_2$ and $B_2=V_2^\top\nabla_{yy}g(x_2,y^*(x_2))V_2$, for any $x_1$ and $x_2$, we have
\begin{align*}
    &~~~~~\|\nabla y^*(x_1)-\nabla y^*(x_2)\|\\
    &=\Big \|V_2\left(V_2^\top\nabla_{yy}g(x_1,y^*(x_1))V_2\right)^{-1}V_2^\top\left(\nabla_{yx}g(x_1,y^*(x_1))-\nabla_{yy}g(x_1,y^*(x_1)) A^\dagger \nabla h(x_1)\right)\\
    &~~~~~-V_2\left(V_2^\top\nabla_{yy}g(x_2,y^*(x_2))V_2\right)^{-1}V_2^\top\left(\nabla_{yx}g(x_2,y^*(x_2))-\nabla_{yy}g(x_2,y^*(x_2)) A^\dagger \nabla h(x_2)\right)\Big\|\\
    &~~~~~+\|A^\dagger(\nabla h(x_1)-\nabla h(x_2))\|\\
    &\leq \|V_2B_1^{-1}V_2^\top\|\|\nabla_{yx}g(x_1,y^*(x_1))-\nabla_{yx}g(x_2,y^*(x_2)))\|\\
    &~~~~~+\|V_2B_1^{-1}V_2^\top\|\|\nabla_{yy}g(x_1,y^*(x_1)) A^\dagger \nabla h(x_1)-\nabla_{yy}g(x_2,y^*(x_2)) A^\dagger \nabla h(x_2)\|\\
    &~~~~~+\|V_2(B_1^{-1}-B_2^{-1})V_2^\top\|\|\nabla_{yx}g(x_2,y^*(x_2))-\nabla_{yy}g(x_2,y^*(x_2)) A^\dagger \nabla h(x_2)\|+\ell_{h,1}\|A^\dagger\|\|x_1-x_2\|\\
    &\stackrel{(a)}{\leq} \frac{1}{\mu_g}\|\nabla_{yx}g(x_1,y^*(x_1))-\nabla_{yx}g(x_2,y^*(x_2))\|\\
    &~~~~~+\frac{1}{\mu_g}\|\nabla_{yy}g(x_1,y^*(x_1))-\nabla_{yy}g(x_2,y^*(x_2))\|\|A^\dagger\|\|\nabla h(x_1)\|\\
    &~~~~~+\frac{1}{\mu_g}\|\nabla_{yy}g(x_2,y^*(x_2))\|\|A^\dagger\|\|\nabla h(x_1)-\nabla h(x_2)\|\\
    &~~~~~+\frac{\ell_{g,1}\left(1+\|A^\dagger\|\ell_{h,0}\right)}{\mu_g^2}\|\nabla_{yy}g(x_1,y^*(x_1))-\nabla_{yy}g(x_2,y^*(x_2))\|+\ell_{h,1}\|A^\dagger\|\|x_1-x_2\|\\
    &\stackrel{(b)}{\leq}\frac{\ell_{g,2}\left(1+\|A^\dagger\|\ell_{h,0}\right)(1+L_y)(1+\frac{\ell_{g,1}}{\mu_g})+(\ell_{g,1}+\mu_g)\|A^\dagger\|\ell_{h,1}}{\mu_g}\|x_1-x_2\|\numberthis\label{refer}
\end{align*}
where (a) comes from \eqref{bound_v}, \eqref{bound-jac} and the following fact 
\begin{align*}
    &~~~~~V_2\left(B_1^{-1}-B_2^{-1}\right)V_2^\top\\
    &=V_2B_1^{-1}\left(B_2-B_1\right)B_2^{-1}V_2^\top\\
    &=V_2B_1^{-1}\left(\left(V_2^\top\nabla_{yy}g(x_2,y^*(x_2))V_2\right)-\left(V_2^\top\nabla_{yy}g(x_1,y^*(x_1))V_2\right)\right)B_2^{-1}V_2^\top\\
    &=V_2B_1^{-1}V_2^\top \left(\nabla_{yy}g(x_2,y^*(x_2))-\nabla_{yy}g(x_1,y^*(x_1))\right)V_2B_2^{-1}V_2^\top
\end{align*}
so that 
\begin{align*}
    \|V_2\left(B_1^{-1}-B_2^{-1}\right)V_2^\top\|&\leq \|V_2B_1^{-1}V_2^\top\|\|\nabla_{yy}g(x_2,y^*(x_2))-\nabla_{yy}g(x_1,y^*(x_1))\|\|V_2B_2^{-1}V_2^\top\|\\
    &\leq \frac{1}{\mu_g^2}\|\nabla_{yy}g(x_2,y^*(x_2))-\nabla_{yy}g(x_1,y^*(x_1))\|\numberthis\label{inverse}
\end{align*}
and (b) comes from 
\begin{align*}
    \|\nabla^2 g(x_1,y^*(x_1))-\nabla^2 g(x_2,y^*(x_2))\|&\leq \ell_{g,2}\left[\|x_1-x_2\|+\|y^*(x_1)-y^*(x_2)\|\right]\\
    &\leq \ell_{g,2}\left(1+L_y\right)\|x_1-x_2\|\numberthis\label{Hessian-y*}
\end{align*}
from which the proof is complete. 
\end{proof}

\subsection{Supporting lemmas}
\begin{lemma}\label{F-smooth}
Under Assumption \ref{as1}--\ref{as2} and \ref{ash}, $F(x)$ is smooth with constant $L_F$ which is defined as
\begin{align}
    L_F:=\ell_{f,1}\left(1+L_y\right)^2+\ell_{f,0}L_{yx}.
\end{align}
\end{lemma}
\begin{proof}
For any $x_1$ and $x_2$, we have that
\begin{align*}
    \|\nabla F(x_1)-\nabla F(x_2)\|&=\|\nabla_x f(x_1,y^*(x_1))+\nabla^\top y^*(x_1)\nabla_y f(x_1,y^*(x_1))\\
    &~~~~~-\nabla_x f(x_2,y^*(x_2))+\nabla^\top y^*(x_2)\nabla_y f(x_2,y^*(x_2))\|\\
    &\leq \|\nabla_x f(x_1,y^*(x_1))-\nabla_x f(x_2,y^*(x_2))\|\\
    &~~~~~+\|\nabla^\top y^*(x_1)\nabla_y f(x_1,y^*(x_1))-\nabla^\top y^*(x_2)\nabla_y f(x_2,y^*(x_2))\|\\
    &\leq \ell_{f,1}\left(\|x_1-x_2\|+\|y^*(x_1)-y^*(x_2)\|\right)\\
    &~~~~~+\|\nabla y^*(x_1)\|\|\nabla_y f(x_1,y^*(x_1))-\nabla_y f(x_2,y^*(x_2))\|\\
    &~~~~~+\|\nabla_y f(x_2,y^*(x_2))\|\|\nabla y^*(x_1)-\nabla y^*(x_2)\|\\
    &\stackrel{(a)}{\leq} \left(\ell_{f,1}\left(1+L_y\right)^2+\ell_{f,0}L_{yx}\right)\|x_1-x_2\|=L_F\|x_1-x_2\|
\end{align*}
where (a) comes frome the Lipschitz continuity of $y^*(x),\nabla y^*(x)$ in Lemma \ref{y-smooth} and the Lipschitz continuity of $\nabla f$ and $f$ in Assumption \ref{as1}. 
\end{proof}

For simplicity, we denote 
\begin{align}
    &\overline\nabla f(x,y)=\nabla_x f(x,y)+\left[\left(\nabla h(x)^\top A^{\dagger\top}\nabla_{yy}g(x,y)-\nabla_{xy}g(x,y)\right)\nonumber\right.\\
    &\left.\qquad\qquad\quad\times  V_2(V_2^\top \nabla_{yy}g(x,y)V_2)^{-1} V_2^\top-\nabla h(x)^\top A^{\dagger\top}\right]\nabla_y f(x,y).
\end{align}

\begin{lemma}[\bf Boundness of $\overline\nabla f(x,y)$]\label{lm-bound}
Under Assumption \ref{as1}--\ref{as2}, for any $x,y$, $
    \|\overline{\nabla}f(x,y)\|\leq \ell_{f,0}\left(1+L_y\right)
$.

\end{lemma}
\begin{proof}
Based on \eqref{bound_v}, we have that  
\begin{align}\label{sc}
    \left\|V_2\left(V_2^\top\nabla_{yy}g(x,y)V_2\right)^{-1}V_2^\top\right\|\leq\frac{1}{\mu_g}.
\end{align}
Then we can obtain the bound for $\overline{\nabla}f(x,y)$ since
\begin{align*}
    \|\overline\nabla f(x,y)\|&\leq\|\nabla_x f(x,y)\|+\|\nabla h(x)^\top A^{\dagger\top}\nabla_{yy}g(x,y)-\nabla_{xy}g(x,y)\|\\
    &~~~~~\times\| V_2(V_2^\top \nabla_{yy}g(x,y)V_2)^{-1} V_2^\top\|\|\nabla_y f(x,y)\|+\| \nabla h(x)^\top A^{\dagger\top}\|\|\nabla_y f(x,y)\|\\
    &\leq \ell_{f,0}\left(1+\frac{\ell_{g,1}\left(1+\ell_{h,0}\|A^\dagger\|\right)}{\mu_g}+\ell_{h,0}\|A^\dagger\|\right)=\ell_{f,0}\left(1+L_y\right)
\end{align*}
from which the proof is complete. 
\end{proof}

\begin{lemma}[\textbf{Lipschitz continuity of $\overline{\nabla}f(x,y)$}]\label{smooth-barf}
Under Assumption \ref{as1}--\ref{as2} and \ref{ash}, $\overline{\nabla}f(x,y)$ is $L_f$ Lipschitz continuous with respect to $y$, where the constant is defined as  
\begin{align}
    L_f:=\left(1+\ell_{h,0}\|A^\dagger\|\right)\left(\ell_{f,1}+\frac{\ell_{g,1}\ell_{f,1}+\ell_{f,0}\ell_{g,2}}{\mu_g}+\frac{\ell_{f,0}\ell_{g,1}\ell_{g,2}}{\mu_{g,1}^2}\right).
    \label{lip-bar}
\end{align}
\end{lemma}
\begin{proof}\allowdisplaybreaks
For simplicity, we define some notations first. 
\begin{align*}
    &B_1=V_2^\top \nabla_{yy}g(x,y_1)V_2, ~~~B_2=V_2^\top \nabla_{yy}g(x,y_2)V_2,\\
    &C_1=\nabla h(x)^\top A^{\dagger\top}\nabla_{yy}g(x,y_1)-\nabla_{xy}g(x,y_1),\\
    &C_2=\nabla h(x)^\top A^{\dagger\top}\nabla_{yy}g(x,y_2)-\nabla_{xy}g(x,y_2).
\end{align*}

For $i=1,2$, according to \eqref{bound_v}, we have the following bounds.
\begin{align}\label{bounds}
    &\|C_i\|\leq \left(1+\ell_{h,0}\|A^\dagger\|\right)\ell_{g,1}, ~~~~~\|V_2B_i^{-1}V_2^\top\|\leq\frac{1}{\mu_g}, ~~~\|\nabla_y f(x,y_i)\|\leq \ell_{f,0}.
\end{align}
Besides,  we can also bound their differences as
\begin{align*}
    &\|C_1-C_2\|\leq\left(1+\|A^\dagger\|\ell_{h,0}\right)\|\nabla^2 g(x,y_1)-\nabla^2 g(x,y_2)\|\leq \left(1+\|A^\dagger\|\ell_{h,0}\right)\ell_{g,2}\|x_1-x_2\|
\end{align*}
and 
\begin{align*}
        &\|V_2B_1^{-1}V_2^\top-V_2B_2^{-1}V_2^\top\|\stackrel{(a)}{\leq} \|V_2B_1^{-1}V_2^\top\|\|V_2B_2^{-1}V_2^\top\|\|\nabla_{yy}g(x,y_1)-\nabla_{yy}g(x,y_2)\|\\
    &~~~~~~~~~~~~~~~~~~~~~~~~~~~~~~~~~~~~~~~~~~~~~\leq\frac{\ell_{g,2}}{\mu_g^2}\|y_1-y_2\|
\end{align*}
where (a) is due to 
\begin{align*}
    &~~~~~V_2\left(B_1^{-1}-B_2^{-1}\right)V_2^\top\\
    &=V_2B_1^{-1}\left(B_2-B_1\right)B_2^{-1}V_2^\top\\
    &=V_2B_1^{-1}\left(\left(V_2^\top\nabla_{yy}g(x,y_2)V_2\right)-\left(V_2^\top\nabla_{yy}g(x,y_1)V_2\right)\right)B_2^{-1}V_2^\top\\
    &=V_2B_1^{-1}V_2^\top \left(\nabla_{yy}g(x,y_2)-\nabla_{yy}g(x,y_1)\right)V_2B_2^{-1}V_2^\top.\numberthis\label{lip-B} 
\end{align*}
Likewise, we have 
\begin{equation}
 \|\nabla_y f(x,y_1)-\nabla_y f(x,y_2)\|\leq\ell_{f,1}\|x_1-x_2\|. \label{bounds_2}
\end{equation}

Thus, for any $x,y_1,y_2$, based on \eqref{bounds} and \eqref{bounds_2}, we have  
\begin{align*}
    &\quad~\|\overline{\nabla}f(x,y_1)-\overline{\nabla}f(x,y_2)\|\\
    &\leq \|\nabla_x f(x,y_1)-\nabla_x f(x,y_2)\|+\|C_1V_2B_1^{-1}V_2^\top \nabla_y f(x,y_1)-C_2V_2B_2^{-1}V_2^\top \nabla_y f(x,y_2)\|\\
    &~~~~~+\|\nabla h(x)\|\|A^\dagger\|\|\nabla_y f(x,y_1)-\nabla_y f(x,y_2)\|\\
    &\stackrel{(a)}{\leq} (1+\ell_{h,0}\|A^\dagger\|)\ell_{f,1}\|y_1-y_2\|+\|C_1\|\|V_2B_1^{-1}V_2^\top\|\|\nabla_x f(x,y_1)-\nabla_x f(x,y_2)\|\\
    &~~~~~+\|C_1\|\|\nabla_x f(x,y_2)\|\|V_2B_1^{-1}V_2^\top-V_2B_2^{-1}V_2^\top\|+\|V_2B_2^{-1}V_2^\top\|\|\nabla_x f(x,y_2)\|\|C_1-C_2\|\\
    &\stackrel{(b)}{\leq}\left(1+\ell_{h,0}\|A^\dagger\|\right)\left(\ell_{f,1}+\frac{\ell_{g,1}\ell_{f,1}+\ell_{f,0}\ell_{g,2}}{\mu_g}+\frac{\ell_{f,0}\ell_{g,1}\ell_{g,2}}{\mu_{g,1}^2}\right)\|y_1-y_2\| \numberthis
\end{align*}
where (a) is due to 
\begin{align*}
    &C_1D_1E_1-C_2D_2E_2\\
    =&C_1D_1E_1-C_1D_1E_2+C_1D_1E_2-C_1D_2E_2+C_1D_2E_2-C_2D_2E_2\\
    =& C_1D_1(E_1-E_2)+C_1E_2(D_1-D_2)+D_2E_2(C_1-C_2)\numberthis\label{CDE}
\end{align*}
(b) comes from \eqref{bounds} and \eqref{bounds_2}, 
from which the proof is complete. 
\end{proof}

\section{Theoretical Analysis for {\sf AiPOD}}
In this section, we present the proof of Algorithm \ref{stoc-alg}.
We define
\begin{equation}
  \mathcal{F}_{k}^s:=\sigma\{y^0,x^0,\cdots, y^k,x^k,y^{k,1},\cdots,y^{k,s}\}  
\end{equation}
where $\sigma\{\cdot\}$ denotes the $\sigma$-algebra generated by the random variables. Then it follows that $\mathcal{F}_{k}^S=\sigma\{y^0,x^0,\cdots, y^{k+1}\}$. 

\subsection{Supporting lemmas of Theorem \ref{thm}} 

To prove the bias and variance of gradient estimator $h_f^k$, we leverage the following fact. 

\begin{lemma}[{\citep[Lemma 12]{hong2020two}}]\label{cite}
Let $Z_i$ be a sequence of stochastic matrices  defined recursively as $Z_{i}=Y_{i}Z_{i-1}, i\geq 0$ with $Z_{-1}=I\in\mathbb{R}^{d\times d}$, $Y_i$ are independent, symmetric random matrix satisfying  
\begin{align*}
    \|\EE\left[Y_i\right]\|\leq 1-\mu,~~ \EE\left[\|Y_i-\EE\left[Y_i\right]\|^2\right]\leq\sigma^2.
\end{align*}
If $(1-\mu)^2+\sigma^2<1$, then for any $i>0$, it holds that
\begin{align*}
    \EE\left[\|Z_i\|^2\right]\leq d\left((1-\mu)^2+\sigma^2\right)^i.
\end{align*}
\end{lemma}
Based on this lemma, we can bound the second moment bound of Hessian inverse estimator. 
\begin{lemma}\label{lm:J_yy}
Under Assumption \ref{as1}--\ref{ash}, let $\tilde c=\frac{\mu_g}{\mu_g^2+\sigma_{g,2}^2}$ and for any $k$, denote the Hessian inverse estimator as
\begin{align*}
    H_{yy}^k=\frac{\tilde cN}{\ell_{g, 1}} \prod_{n=0}^{N^{\prime}}\left(I-\frac{\tilde c}{\ell_{g, 1}} V_2^\top\nabla_{y y}^{2} g\left(x^k, y^{k+1} ; \phi_{(n)}^k\right)V_2\right).
\end{align*}
With $r=\operatorname{Rank}(A)$, the second moment bound of $H_{yy}^k$ can be bounded as
\begin{align*}
    \EE\left[\|H_{yy}^k\|^2|\mathcal{F}_k^S\right]\leq \frac{N(d_y-r)}{\ell_{g,1}(\mu_g^2+\sigma_{g,2}^2)}.
\end{align*}
\end{lemma}
\begin{proof}
Letting $Y_n=I-\frac{\tilde c}{\ell_{g, 1}} V_2^\top\nabla_{y y}^{2} g\left(x, y ; \phi_{(n)}\right)V_2$, it follows that 
\begin{align*}
    \|\EE\left[Y_n\right]\|\leq \left(1-\frac{\tilde c\mu_g}{\ell_{g,1}}\right), ~~\EE\left[\|Y_n-\EE\left[Y_n\right]\|^2\right]\leq\frac{\tilde c^2\sigma_{g,2}^2}{\ell_{g,1}^2}
\end{align*}
Moreover, since 
$$\left(1-\frac{\tilde c\mu_g}{\ell_{g,1}}\right)^2+\frac{\tilde c^2\sigma_{g,2}^2}{\ell_{g,1}^2}=1-\frac{2\tilde c\mu_g}{\ell_{g,1}}+\frac{\tilde c^2(\mu_g^2+\sigma_{g,2}^2)}{\ell_{g,1}^2}=1-\frac{\mu_g^2}{\ell_{g,1}\left(\mu_g^2+\sigma_{g,2}^2\right)}<1
$$
which satisfies the condition in Lemma \ref{cite}, we can then plug $Y_n$ into Lemma \ref{cite} and achieves the second moment bound for $H_{yy}^k$.
\begin{align*}
    \EE\left[\|H_{yy}^k\|^2|\mathcal{F}_k^S\right]&=\EE\left[\EE\left[\|H_{yy}^k\|^2|N^\prime,\mathcal{F}_k^S\right]|\mathcal{F}_k^S\right]\leq\EE\left[\frac{\tilde c^2N^2(d_y-r)}{\ell_{g,1}^2}\left(1-\frac{\mu_g^2}{\ell_{g,1}\left(\mu_g^2+\sigma_{g,2}^2\right)}\right)^{N^\prime}\Big|N^\prime\right]\\
    &\leq \frac{\tilde c^2N(d_y-r)}{\ell_{g,1}^2}\sum_{n=0}^{N-1}\left(1-\frac{\mu_g^2}{\ell_{g,1}\left(\mu_g^2+\sigma_{g,2}^2\right)}\right)^{n}\\
    &\leq \frac{\tilde c^2N(d_y-r)}{\ell_{g,1}^2}\frac{\ell_{g,1}\left(\mu_g^2+\sigma_{g,2}^2\right)}{\mu_g^2}\stackrel{(a)}{\leq} \frac{N(d_y-r)}{\ell_{g,1}\left(\mu_g^2+\sigma_{g,2}^2\right)}.
\end{align*}
where $r$ is rank of $A$ and (a) comes from the choice of $\tilde c$. 
\end{proof}

\begin{lemma}[\bf Bias and variance of gradient estimator]
Let $\tilde c=\frac{\mu_g}{\mu_g^2+\sigma_{g,2}^2},r=\operatorname{rank}(A)$ and define $$\bar h_f^k=\EE\left[h_f^k\right|\mathcal{F}_k^S], $$then $h_f^k$ is a biased estimator of UL gradient which satisfies that
\begin{align}
    &\|\bar h_f^k-\overline\nabla f(x^k,y^{k+1})\|\leq L_y\ell_{f,0}\left(1-\frac{\mu_g^2}{\ell_{g,1}(\mu_g^2+\sigma_{g,2}^2)}\right)^{N}=: b_{k}\label{bias-1} \\
    &\EE\left[\|h_f^k-\bar h_f^k\|^2|\mathcal{F}_k^S\right]\leq (1+\ell_{h,0}\|A^\dagger\|)\sigma_f^2+\frac{4N(1+\ell_{h,0}^2\|A^\dagger\|^2)(d_y-r)(\ell_{g,1}^2+\sigma_{g,2}^2)\left(2\sigma_f^2+\ell_{f,0}^2\right)}{\ell_{g,1}(\mu_g^2+\sigma_{g,2}^2)}\nonumber\\
    &~~~~~~~~~~~~~~~~~~~~~~~~~~~~~~~~=:\tilde\sigma_f^2=\mathcal{O}\left(N\right).\label{var-2}
\end{align}
\label{lm:var_stoc}
\end{lemma}
\begin{proof}\allowdisplaybreaks
We first prove \eqref{bias-1} by noticing that the error by finite updates can be bounded by
\begin{align}
    \left\|V_2\left((I-D)^{-1}-\sum_{n=0}^{N-1} D^n\right)V_2^\top\right\|&=\left\|V_2\left(\sum_{n=N}^\infty D^n\right)V_2^\top\right\|\nonumber\\
    &=\left\|\sum_{n=N}^\infty (V_2DV_2^\top)^n\right\|\nonumber\\
    &\leq\sum_{n=N}^\infty \|V_2DV_2^\top\|^n=\frac{\|V_2DV_2^\top\|^{N}}{1-\|V_2DV_2^\top\|}
    \label{D_error}
\end{align}
Thus, letting $D=I-\frac{\tilde c}{\ell_{g,1}}V_2^\top\nabla_{yy}g(x,y)V_2$ and multiplying each side by $\frac{\tilde c}{\ell_{g,1}}$, we obtain that
\begin{align*}
    &~~~~\left\|V_2\left(\left(V_2^\top\nabla_{yy}g(x,y)V_2 \right)^{-1}-\frac{\tilde c}{\ell_{g,1}}\sum_{n=0}^{N-1} (I-\frac{\tilde c}{\ell_{g,1}}V_2^\top\nabla_{yy}g(x,y)V_2)^n\right)V_2^\top\right\|\\
    &\leq \frac{\tilde c\|V_2\left(I-\frac{\tilde c}{\ell_{g,1}}\nabla_{yy}g(x,y)\right)V_2^\top\|^{N}}{\ell_{g,1}\left(1-\|V_2\left(I-\frac{\tilde c}{\ell_{g,1}}\nabla_{yy}g(x,y)\right)V_2^\top\|\right)}\\
    &\leq \frac{\left(1-\frac{\tilde c\mu_g}{\ell_{g,1}}\right)^N}{\mu_g}\numberthis\label{Hessian-error}
\end{align*}
where the second inequality holds according to $\mu_g I_{d_y-r}\preceq V_2^\top\nabla_{yy}g(x,y)V_2$. Then we have 
\begin{align*}
    &~~~~~\|\overline{\nabla} f(x^k,y^{k+1})-\bar h_f^k\|\\
    &\leq\left\|\nabla h(x^k)^\top A^{\dagger\top}\nabla_{yy}g(x^k,y^{k+1})-\nabla_{xy}g(x^k,y^{k+1})\right\|\|\nabla_y f(x^k,y^{k+1})\|\\
    &~~~~~\times  \left\|V_2\left(\left(V_2^\top\nabla_{yy}g(x^k,y^{k+1})V_2 \right)^{-1}-\frac{\tilde c}{\ell_{g,1}}\sum_{n=0}^{N-1} (I-\frac{\tilde c}{\ell_{g,1}}V_2^\top\nabla_{yy}g(x,y)V_2)^n\right)V_2^\top\right\|\\
    &\leq \frac{\left(1+\ell_{h,0}\|A^\dagger\|\right)\ell_{g,1}\ell_{f,0}\left(1-\frac{\tilde c\mu_g}{\ell_{g,1}}\right)^N}{\mu_g}\leq L_y\ell_{f,0}\left(1-\frac{\tilde c\mu_g}{\ell_{g,1}}\right)^N\numberthis\label{subgrad-error}
\end{align*}
where the second term of (a) is derived from $\|X^{-1}-Y^{-1}\|\leq\|X^{-1}\|\|X-Y\|\|Y^{-1}\|$. Then plugging in the choice of $\tilde c$ to \eqref{subgrad-error} results in \eqref{bias-1}.

The proof of \eqref{var-2} is based on Lemma \ref{lm:J_yy}. For ease of narration, we denote
\begin{align*}
    &\nabla_{xy}^h g(x^{k}, y^{k+1} ; \phi_{(0)}^{k}):=\nabla h(x^k)^\top A^{\dagger\top}\nabla_{yy} g(x^{k}, y^{k+1} ; \phi_{(0)}^{k})-\nabla_{x y} g(x^{k}, y^{k+1} ; \phi_{(0)}^{k})\\
    &\nabla_{xy}^h g(x^{k}, y^{k+1}):=\nabla h(x^k)^\top A^{\dagger\top}\nabla_{yy}^{2} g(x^{k}, y^{k+1})-\nabla_{x y}^{2} g(x^{k}, y^{k+1}). 
\end{align*}
We notice that 
\begin{align*}
    &\quad~\EE\left[\nabla_{xy}^h g(x^{k}, y^{k+1} ; \phi_{(0)}^{k})|\mathcal{F}_k^S\right]=\nabla_{xy}^h g(x^{k}, y^{k+1})
\end{align*}
and then the bias and variance of $\nabla_{xy}^h g(x^{k}, y^{k+1} ; \phi_{(0)}^{k})$ can be bounded by
\begin{align*}
    &\quad~\|\nabla_{xy}^h g(x^{k}, y^{k+1})\|^2\leq 2(1+\ell_{h,0}^2\|A^\dagger\|^2)\ell_{g,1}^2\numberthis\label{bias-Hk}\\
    &\quad~\EE\left[\|\nabla_{xy}^h g(x^{k}, y^{k+1} ; \phi_{(0)}^{k})-\nabla_{xy}^h g(x^{k}, y^{k+1})\|^2|\mathcal{F}_k^S\right]\\
    &\leq\|\nabla h(x^k)^\top A^{\dagger\top}\|^2\EE\left[\|\nabla_{yy} g(x^{k}, y^{k+1}; \phi_{(0)}^{k})-\nabla_{yy} g(x^{k}, y^{k+1})\|^2|\mathcal{F}_k^S\right]\\
    &\quad~+\EE\left[\|\nabla_{xy} g(x^{k}, y^{k+1} \phi_{(0)}^{k})-\nabla_{xy} g(x^{k}, y^{k+1})\|^2|\mathcal{F}_k^S\right]\\
    &\leq(1+\ell_{h,0}^2\|A^\dagger\|^2)\sigma_{g,2}^2.\numberthis\label{variance-Hk}
\end{align*}
Thus adding \eqref{bias-Hk} and \eqref{variance-Hk}, we arrive at the second moment bound for $\nabla_{xy}^h g(x^{k}, y^{k+1} ; \phi_{(0)}^{k})$ as
\begin{align}
    &~~~~~\EE\left[\|\nabla_{xy}^h g(x^{k}, y^{k+1} ; \phi_{(0)}^{k})\|^2|\mathcal{F}_k^S\right]\nonumber\\
    &=\|\EE\left[\nabla_{xy}^h g(x^{k}, y^{k+1} ; \phi_{(0)}^{k})|\mathcal{F}_k^S\right]\|^2+\EE\left[\|\nabla_{xy}^h g(x^{k}, y^{k+1} ; \phi_{(0)}^{k})-\nabla_{xy}^h g(x^{k}, y^{k+1})\|^2|\mathcal{F}_k^S\right]\nonumber\\
    &\leq (1+\ell_{h,0}^2\|A^\dagger\|^2)(2\ell_{g,1}^2+\sigma_{g,2}^2). \label{second=Hk} 
\end{align}
Then the variance of $h_f^k$ can be decomposed and bounded as
{\small
\begin{align*}
    &~~~~~\EE\left[\|h_f^k-\bar h_f^k\|^2|\mathcal{F}_k^S\right]\\
    &\leq\EE\left[\|\nabla_x f(x^k,y^{k+1};\xi^k)-\nabla_x f(x^k,y^{k+1})\|^2|\mathcal{F}_k^S\right]\\
    &~~~~~~~+\|\nabla h(x^k) A^\dagger\|\EE\left[\|\nabla_y f(x^k,y^{k+1};\xi^k)-\nabla_y f(x^k,y^{k+1})\|^2|\mathcal{F}_k^S\right]\\
    &~~~~~~~+\EE\Big[\|\nabla_{xy}^h g(x^{k}, y^{k+1} ; \phi_{(0)}^{k}) V_2 H_{yy}^k V_2^\top\nabla_y f(x^k,y^{k+1};\xi^k)\\
    &~~~~~~~-\nabla_{xy}^h g(x^{k}, y^{k+1})V_2\EE[H_{yy}^k|\mathcal{F}_k^S]V_2^\top\nabla_y f(x^k,y^{k+1})\|^2|\mathcal{F}_k^S\Big]\\
    &\stackrel{(a)}{\leq}(1+\ell_{h,0}\|A^\dagger\|)\sigma_f^2+3\EE[\|\nabla_{xy}^h g(x^{k}, y^{k+1} ; \phi_{(0)}^{k})\|^2|\mathcal{F}_k^S]\EE[\|H_{yy}^k\|^2|\mathcal{F}_k^S]\EE[\|\nabla_y f(x^k,y^{k+1};\xi^k)-\nabla_y f(x^k,y^{k+1})\|^2|\mathcal{F}_k^S]\\
    &~~~~~~+3\EE[\|\nabla_{xy}^h g(x^{k}, y^{k+1} ; \phi_{(0)}^{k})\|^2|\mathcal{F}_k^S]\EE[\|\nabla_y f(x^k,y^{k+1};\xi^k)\|^2|\mathcal{F}_k^S]\EE[\|H_{yy}^k-\EE[H_{yy}^k|\mathcal{F}_k^S]\|^2|\mathcal{F}_k^S]\\
    &~~~~~~+3\EE[\|H_{yy}^k\|^2|\mathcal{F}_k^S]\EE[\|\nabla_y f(x^k,y^{k+1};\xi^k)\|^2|\mathcal{F}_k^S]\EE[\|\nabla_{xy}^h g(x^{k}, y^{k+1} ; \phi_{(0)}^{k})-\nabla_{xy}^h g(x^{k}, y^{k+1})\|^2|\mathcal{F}_k^S]\\
    &\stackrel{(b)}{\leq}(1+\ell_{h,0}\|A^\dagger\|)\sigma_f^2+\frac{3N(1+\ell_{h,0}^2\|A^\dagger\|^2)(d_y-r)}{\ell_{g,1}(\mu_g^2+\sigma_{g,2}^2)}\left[(2\ell_{g,1}^2+\sigma_{g,2}^2)\left(2\sigma_f^2+\ell_{f,0}^2\right)+(\sigma_f^2+\ell_{f,0}^2)\sigma_{g,2}^2\right]\\
    &\leq(1+\ell_{h,0}\|A^\dagger\|)\sigma_f^2+\frac{6N(1+\ell_{h,0}^2\|A^\dagger\|^2)(d_y-r)(\ell_{g,1}^2+\sigma_{g,2}^2)\left(2\sigma_f^2+\ell_{f,0}^2\right)}{\ell_{g,1}(\mu_g^2+\sigma_{g,2}^2)}=:\tilde\sigma_f^2
\end{align*}}
where (a) comes from \eqref{CDE} and $(A+B+C)^2\leq 3(A^2+B^2+C^2)$, and (b) comes from the second moment bound and variance of $\nabla_{xy}^h g(x^{k}, y^{k+1} ; \phi_{(0)}^{k}),H_{yy}$ and $\nabla_y f(x^k,y^{k+1};\xi^k)$. 
\end{proof}

\subsection{Descent of upper and lower levels}
\begin{lemma}[\bf{Descent of upper level}]\label{lm-up-stoc} Suppose Assumption \ref{as1}--\ref{ash} hold, then the sequence of $x_k$ generated by Algorithm \ref{stoc-alg} satisfies
\begin{align*}
    \EE[F(x^{k+1})]-\EE[F(x^k)]&\leq-\frac{\alpha}{2}\EE\left[\|\nabla F(x^k)\|_{P_x}^2\right]+\alpha L_f^2\EE\left[\|y^*(x^k)-y^{k+1}\|^2\right]+\alpha b_k^2\\
   &~~~~~ -\left(\frac{\alpha}{2}-\frac{L_F\alpha^2}{2}\right) \EE\left[\|\bar h_f^k\|_{P_x}^2\right]+\frac{L_F\alpha^2\tilde\sigma_f^2}{2}\numberthis\label{stoc:up}
\end{align*}
where $P_x=I-B^\dagger B$ is the projection matrix of $B$ and $B^\dagger$ is the Moore-Penrose inverse of $B$. 
\end{lemma}
\allowdisplaybreaks
\begin{proof}
Since $\mathcal{X}=\{x~|~Bx=e\}$, for any $x$, we have that $\operatorname{Proj}_{\mathcal{X}}(x)=(I-B^\dagger B)x+B^\dagger e$ is a linear operator of $x$ according to \eqref{closed-form-proj}. Thus, we have
\begin{align*}
    x^{k+1}&=\operatorname{Proj}_{\mathcal{X}}(x^k-\alpha h_f^k)=(I-B^\dagger B)(x^k-\alpha h_f^k)+B^\dagger e\\
    &=(I-B^\dagger B)x^k+B^\dagger e-(I-B^\dagger B)(\alpha h_f^k)\\
    &=\operatorname{Proj}_{\mathcal{X}}(x^k)-\alpha(I-B^\dagger B)h_f^k\\
    &=x^k-\alpha(I-B^\dagger B)h_f^k\numberthis\label{linear-update}
\end{align*}
where the last equality is due to $x^k\in\mathcal{X}$. 

Taking the expectation of $F(x^{k+1})$ conditioned on $\mathcal{F}_k^S$, we get
\begin{align*}
    \EE\left[F(x^{k+1})|\mathcal{F}_k^S\right]
    &\stackrel{(a)}{\leq} F(x^k)+\langle \nabla F(x^k),\EE\left[x^{k+1}-x^k|\mathcal{F}_k^S\right]\rangle+\frac{L_F}{2}\EE\left[\|x^{k+1}-x^k\|^2|\mathcal{F}_k^S\right]\\
    &= F(x^k)-\alpha\langle \nabla F(x^k),(I-B^\dagger B)\bar h_f^k\rangle+\frac{L_F}{2}\alpha^2 \EE\left[\|(I-B^\dagger B)h_f^k\|^2|\mathcal{F}_k^S\right]\\
    &\stackrel{(b)}{\leq} F(x^k)-\frac{\alpha}{2}\|(I-B^\dagger B)\nabla F(x^k)\|^2+\frac{\alpha}{2}\|(I-B^\dagger B)(\nabla F(x^k)-\bar h_f^k)\|^2\\
    &~~~~~-\left(\frac{\alpha}{2}-\frac{L_F\alpha^2}{2}\right) \|(I-B^\dagger B)\bar h_f^k\|^2+\frac{L_F\alpha^2\tilde\sigma_f^2}{2}\\
    &\stackrel{(c)}{\leq} F(x^k)-\frac{\alpha}{2}\|\nabla F(x^k)\|_{P_x}^2+\frac{\alpha}{2}\|\nabla F(x^k)-\bar h_f^k\|^2\\
    &~~~~~-\left(\frac{\alpha}{2}-\frac{L_F\alpha^2}{2}\right) \|\bar h_f^k\|_{P_x}^2+\frac{L_F\alpha^2\tilde\sigma_f^2}{2}
    \numberthis\label{stoc:upper}
\end{align*}
where (a) comes from the smoothness of $F$, (b) is derived from $2a^\top b= \|a\|^2+\|b\|^2-\|a-b\|^2$, $\EE[\|X\|^2|Y]=\|\EE[X|Y]\|^2+\EE[\|X-\EE[X|Y]\|^2|Y]$, $(I-B^\dagger B)^2=I-B^\dagger B$ and Lemma \ref{lm:var_stoc}, (c) is due to the definition of $\|\cdot\|_{P_x}$ and $\|I-B^\dagger B\|\leq 1$. 

Besides, we decompose the gradient bias term as follows 
\begin{align*}
    \|\nabla F(x^k)-\bar h_f^k\|^2&\leq 2\|\nabla F(x^k)-\overline\nabla f(x^k,y^{k+1})\|^2+2\|\overline\nabla f(x^k,y^{k+1})-\bar h_f^k\|^2\\
    &\leq2\|\overline\nabla f(x^k,y^*(x^k))-\overline\nabla f(x^k,y^{k+1})\|^2+2b_k^2\\
    &\leq 2L_f^2\|y^*(x^k)-y^{k+1}\|+2b_k^2.
    \numberthis\label{grad_bias_stoc}
\end{align*}

Plugging \eqref{grad_bias_stoc} to \eqref{stoc:upper} and taking expectation, we get that
\begin{align*}
    \EE[F(x^{k+1})]-\EE[F(x^k)]&\leq-\frac{\alpha}{2}\EE\left[\|\nabla F(x^k)\|_{P_x}^2\right]+\alpha L_f^2\EE\left[\|y^*(x^k)-y^{k+1}\|^2\right]+\alpha b_k^2\\
   &~~~~~ -\left(\frac{\alpha}{2}-\frac{L_F\alpha^2}{2}\right) \EE\left[\|\bar h_f^k\|_{P_x}^2\right]+\frac{L_F\alpha^2\tilde\sigma_f^2}{2}.
\end{align*}
This completes the proof.
\end{proof}

\begin{lemma}[\bf{Error of lower-level update}]\label{lm-low-stoc}
Suppose that Assumption \ref{as1}--\ref{ash} hold and $\beta\leq\frac{1}{\ell_{g,1}}$, then the error of lower-level variable can be bounded by
\begin{subequations}
\begin{align}
    &\EE[\|y^{k+1}-y^*(x^k)\|^2]\leq\left(1-\beta\mu_g\right)^S\EE[\|y^k-y^*(x^k)\|^2]+S\beta^2\sigma_{g,1}^2\label{40a}\\
    &\EE[\|y^{k+1}-y^*(x^{k+1})\|^2]\leq \left(1+\gamma+ L_{yx}\widetilde C_f^2\alpha^2\right)\EE[\|y^{k+1}-y^*(x^{k})\|^2]\nonumber\\
    &\qquad\qquad\qquad\qquad\qquad\quad+\left(L_y^2+L_{yx}\right)\alpha^2\tilde\sigma_f^2+\left(L_y^2+L_{yx}+\frac{L_y^2}{\gamma}\right)\alpha^2\EE\left[\|\bar h_f^k\|_{P_x}^2\right]\label{40b}
\end{align}
\end{subequations}
where $\widetilde C_f^2:=2b_k^2+2\ell_{f,0}^2\left(1+\frac{\ell_{g,1}}{\mu_g}\right)^2+\tilde\sigma_f^2$, $\gamma$ is parameter which will be chosen in the final theorem. 
\end{lemma}
\begin{proof}
First, since the lower-level objective function is strongly-convex and smooth, when $0\leq\beta\leq\frac{1}{\ell_{g,1}}$, we have the following fact 
\begin{align*}
    &~~~~~\|y_1-\beta\nabla_y g(x,y_1)-(y_2-\beta\nabla_y g(x,y_2))\|^2\\
    &= \|y_1-y_2\|^2+\beta^2\|\nabla_y g(x,y_1)-\nabla_y g(x,y_2))\|^2-2\beta\langle y_1-y_2,\nabla_y g(x,y_1)-\nabla_y g(x,y_2)\rangle\\
    &\leq (1-\beta\mu_g)\|y_1-y_2\|^2-2\beta D_g((x,y_1),(x,y_2))+\beta^2\|\nabla_y g(x,y_1)-\nabla_y g(x,y_2))\|^2\\
    &\leq (1-\beta\mu_g)\|y_1-y_2\|^2\numberthis\label{contract}
\end{align*}
where the first inequality is according to \eqref{point_diver} and the last inequality is due to \eqref{grad_diver} and $\beta\leq\frac{1}{\ell_{g,1}}$.  

Then, for each lower-level update, we obtain that
\begin{align*}
    &~~~~~\EE[\|y^{k,s+1}-y^*(x^k)\|^2|\mathcal{F}_{k,s}]\\
    &=\EE[\|\operatorname{Proj}_{{\cal Y}(x^k)}(y^{k,s}-\beta\nabla_y g(x^k,y^{k,s};\phi^{k,s}))-\operatorname{Proj}_{{\cal Y}(x^k)}(y^*(x^k)-\beta\nabla_y g(x^k,y^*(x^k)))\|^2|\mathcal{F}_{k}^s]\\
    &\leq\EE[\|y^{k,s}-\beta\nabla_y g(x^k,y^{k,s};\phi^{k,s})-y^*(x^k)+\beta\nabla_y g(x^k,y^*(x^k))\|^2|\mathcal{F}_{k}^s]\\
    &\leq\EE[\|y^{k,s}-\beta\nabla_y g(x^k,y^{k,s})-y^*(x^k)+\beta\nabla_y g(x^k,y^*(x^k))\|^2|\mathcal{F}_{k}^s]\\
    &~~~~~+\beta^2\EE[\|\nabla_y g(x^k,y^{k,s};\phi^{k,s})-\nabla_y g(x^k,y^{k,s})\|^2|\mathcal{F}_{k}^s]\\
    &\leq (1-\beta\mu_g)\|y^{k,s}-y^*(x^k)\|^2+\beta^2\sigma_{g,1}^2\numberthis \label{contra-stoc}
\end{align*}
where the first inequality is due to $y^*(x^k)=\operatorname{Proj}_{\mathcal{Y}(x^k)}(y^*(x^k)-\beta\nabla_y g(x^k,y^*(x^k)))$, the second inequality is due to Lemma \ref{lm:matrix} and the last inequality is obtained by \eqref{contract} with $x=x^k,y_2=y^*(x^k),y_1=y^{k,s}$, and Assumption \ref{as-stoc}. Taking the expectation of both sides in \eqref{contra-stoc}, one has
\begin{align}
    \EE[\|y^{k,s+1}-y^*(x^k)\|^2]\leq(1-\beta\mu_g)\EE[\|y^{k,s}-y^*(x^k)\|^2]+\beta^2\sigma_{g,1}^2.\label{42}
\end{align}
Thus, \eqref{40a} can be obtained by telescoping \eqref{42}. 

On the other hand, we have
\begin{align*}
     \|y^{k+1}-y^*(x^{k+1})\|^2&= \|y^{k+1}-y^*(x^{k})\|^2+ \underbrace{\|y^{*}(x^{k})-y^*(x^{k+1})\|^2}_{J_1}\\
     &~~~~~+2\underbrace{\langle y^{k+1}-y^*(x^{k}),y^{*}(x^{k})-y^*(x^{k+1})\rangle}_{J_2}.
\end{align*}

Since $y^*(x)$ is $L_y$ Lipschitz continuous, $J_1$ can be bounded by
\begin{align*}
    \EE\left[J_1\right]&\leq L_{y}^2\EE\left[\|x^{k+1}-x^k\|^2\right]\stackrel{(a)}{=}  \alpha^2L_{y}^2\EE\left[\EE\left[\|(I-B^\dagger B)h_f^k\|^2\right]|\mathcal{F}_k^S\right]\\
    &\stackrel{(b)}{\leq} \alpha^2L_{y}^2\left(\EE\left[\|\bar h_f^k\|_{P_x}^2\right]+\tilde\sigma_f^2\right)\numberthis\label{J1}
\end{align*}
where (a) comes from \eqref{linear-update}, (b) holds since $\EE[\|C\|^2|D]=\|\EE[C|D]\|^2+\EE[\|C-\EE[C|D]\|^2|D]$ and Lemma \ref{lm:var_stoc}. 

Moreover, we can decompose $J_2$ by two terms as follows.
\begin{align*}
    J_2&=\underbrace{-\langle y^{k+1}-y^*(x^{k}), \nabla y^*(x^k)^\top (x^{k+1}-x^k) \rangle}_{J_{2,1}}\\
    &~~~~\underbrace{-\langle y^{k+1}-y^*(x^{k}), y^*(x^{k+1})-y^*(x^k)-\nabla y^*(x^k)^\top(x^{k+1}-x^k) \rangle}_{J_{2,2}}.
\end{align*}
Moreover, the conditional expectation of $J_{2,1}$ can be bounded by
\begin{align*}
    \EE[J_{2,1}|\mathcal{F}_k^S]&=-\langle y^{k+1}-y^*(x^{k}),\EE[\nabla y^*(x^k)^\top (x^{k+1}-x^k)|\mathcal{F}_k^S]\rangle\\
    &\leq -\alpha\langle y^{k+1}-y^*(x^{k}),\nabla y^*(x^k)^\top (I-B^\dagger B)\bar h_f^k\rangle\\
    &\stackrel{(a)}{\leq} \frac{\gamma}{2}\|y^{k+1}-y^*(x^{k})\|^2+\frac{\alpha^2L_y^2}{2\gamma}\|\bar h_f^k\|_{P_x}^2\numberthis\label{J21-condi}
\end{align*}
where (a) comes form Young's inequality and the boundedness of $\nabla y^*(x^k)$. Then taking expectation of \eqref{J21-condi}, we obtain that
\begin{align}
    \EE[J_{2,1}]\leq \frac{\gamma}{2}\EE[\|y^{k+1}-y^*(x^k)\|^2]+\frac{\alpha^2L_y^2}{2\gamma}\EE[\|\bar h_f^k\|_{P_x}^2]. \label{J21}
\end{align}

Based on the smoothness of $y^*(x)$ and Jensen inequality, $J_{2,2}$ can be bounded by
\begin{align*}
    \EE[J_{2,2}]&\leq \EE\left[\|y^{k+1}-y^*(x^k)\|\|y^*(x^{k+1})-y^*(x^k)-\nabla y^*(x^k)^\top(x^{k+1}-x^k)\|^2\right]\\
    &\leq \frac{L_{yx}}{2}\EE\left[\|y^{k+1}-y^*(x^k)\|\|x^{k+1}-x^k\|^2\right]\\
    &\stackrel{(a)}{\leq} \frac{ L_{yx}\alpha^2}{2}\EE[\|y^{k+1}-y^*(x^k)\|^2\| h_f^k\|^2]+\frac{L_{yx}\alpha^2}{2}\EE\left[\|h_f^k\|_{P_x}^2\right]\\
    &\leq\frac{ L_{yx}\alpha^2}{2}\EE\left[\|y^{k+1}-y^*(x^k)\|^2\EE[\| h_f^k\|^2|\mathcal{F}_k^S]\right]+\frac{L_{yx}\alpha^2}{2}\EE\left[\EE\left[\|h_f^k\|_{P_x}^2|\mathcal{F}_k^S\right]\right]\\
    &\leq \frac{ L_{yx}\alpha^2}{2}\EE\left[\|y^{k+1}-y^*(x^k)\|^2\left(\|\bar h_f^k\|^2+\tilde\sigma_f^2\right)\right]+\frac{L_{yx}\alpha^2}{2}\left(\EE[\|\bar h_f^k\|_{P_x}^2]+\tilde\sigma_f^2\right)\\
    &\leq \frac{ L_{yx}\alpha^2}{2}\EE\left[\|y^{k+1}-y^*(x^k)\|^2\left(2\|\bar h_f^k-\overline{\nabla}f(x^k,y^{k+1})\|^2+2\|\overline{\nabla}f(x^k,y^{k+1})\|^2+\tilde\sigma_f^2\right)\right]\\
    &~~~~~+\frac{L_{yx}\alpha^2}{2}\left(\EE[\|\bar h_f^k\|_{P_x}^2]+\tilde\sigma_f^2\right)\\
    &\stackrel{(b)}{\leq} \frac{ L_{yx}\alpha^2}{2}\left(2b_k^2+2\ell_{f,0}^2\left(1+L_y\right)^2+\tilde\sigma_f^2\right)\EE\left[\|y^{k+1}-y^*(x^k)\|^2\right]\\
    &~~~~~+\frac{L_{yx}\alpha^2}{2}\left(\EE[\|\bar h_f^k\|_{P_x}^2]+\tilde\sigma_f^2\right) \numberthis\label{J22}
\end{align*}
where (a) comes from the update \eqref{linear-update}, Young's inequality and $\|h_f^k\|_{P_x}=\|(I-B^\dagger B)h_f^k\|\leq\|h_f^k\|$ and (b) holds from Lemma \ref{lm-bound} and Lemma \ref{lm:var_stoc}.
Then denoting $\widetilde C_f^2:=2b_k^2+2\ell_{f,0}^2\left(1+L_y\right)^2+\tilde\sigma_f^2$ and combining \eqref{J1}, \eqref{J21} and \eqref{J22}, we get 
\begin{align*}
    \EE[\|y^{k+1}-y^*(x^{k+1})\|^2]&\leq \left(1+\gamma+ L_{yx}\widetilde C_f^2\alpha^2\right)\EE[\|y^{k+1}-y^*(x^{k})\|^2]+\left(L_y^2+L_{yx}\right)\alpha^2\tilde\sigma_f^2\\
    &~~~~~~+\left(L_y^2+L_{yx}+\frac{L_y^2}{\gamma}\right)\alpha^2\EE\left[\|\bar h_f^k\|_{P_x}^2\right].
\end{align*}
This completes the proof. 
\end{proof}

\subsection{Proof of Theorem \ref{thm}}\label{sec:thm-alg1-conv}
We first restate a formal version of Theorem \ref{thm} as follows. 
\begin{customthm}{1}
Under Assumption \ref{as1}--\ref{ash}, defining the constants as 
\begin{align}
    \bar\alpha_1=\frac{1}{2L_F+4L_fL_y+\frac{4L_fL_{yx}}{ L_y}},\qquad \bar\alpha_2=\frac{\mu_g}{\ell_{g,1}(5L_fL_y+ L_{yx}\tilde C_f^2)}\label{constant}
\end{align}
and choosing $$\alpha=\min\left(\bar\alpha_1,\bar\alpha_2,\frac{\bar\alpha}{\sqrt{K}}\right),\qquad \beta=\frac{5L_fL_y+ L_{yx}\tilde C_f^2}{\mu_g}\alpha, \qquad N={\cal O}(\log K)$$
then for any $S\geq 1$ in Algorithm \ref{stoc-alg}, we have
\begin{align*}
    \frac{1}{K}\sum_{k=0}^{K-1} \EE\left[\|\nabla F(x^k)\|_{P_x}^2\right]={\cal O}\left(\frac{\log(K)}{\sqrt{K}}\right)
\end{align*}
where $P_x=I-B^\dagger B$ is the projection matrix on $\operatorname{Ker}(B)$ and $\|x\|_{P_x}=\sqrt{x^\top P_x x}$ is the weighted Euclidean norm associated with $P_x$. 
\end{customthm}

\begin{proof}
According to Lemma \ref{lm-low-stoc} and plugging \eqref{40b} into \eqref{40a}, we get that
\begin{align*}
    \EE[\|y^{k+1}-y^*(x^{k+1})\|^2]&\leq \left(1+\gamma+ L_{yx}\widetilde C_f^2\alpha^2\right)\left(1-\beta\mu_g\right)^S\EE[\|y^k-y^*(x^k)\|^2]\\
    &~~~~~+\left(1+\gamma+ L_{yx}\widetilde C_f^2\alpha^2\right)S\beta^2\sigma_{g,1}^2+\left(L_y^2+L_{yx}\right)\alpha^2\tilde\sigma_f^2\\
    &~~~~~+\left(L_y^2+L_{yx}+\frac{L_y^2}{\gamma}\right)\alpha^2\EE\left[\|\bar h_f^k\|_{P_x}^2\right].\numberthis\label{45}
\end{align*}

We can define Lyapunov function as $$\mathbb{V}^k:= F(x^k)+\frac{ L_f}{L_y}\|y^*(x^k)-y^k\|^2.$$



Using Lemma \ref{lm:var_stoc}--\ref{lm-low-stoc}, we get
\begin{align*}
    \EE\left[\mathbb{V}^{k+1}\right]-\EE\left[\mathbb{V}^k\right]&\leq-\frac{\alpha}{2}\EE\left[\|\nabla F(x^k)\|_{P_x}^2\right]+\alpha L_f^2\left(1-\beta\mu_g\right)^S\EE\left[\|y^k-y^*(x^k)\|^2\right]+\alpha b_k^2\\
   &~~~~~ -\left(\frac{\alpha}{2}-\frac{L_F\alpha^2}{2}\right) \EE\left[\|\bar h_f^k\|_{P_x}^2\right]+\frac{L_F\alpha^2_k \tilde\sigma_f^2}{2}\\
   &~~~~~ +\frac{L_f}{L_y}\left[\left(1+\gamma+ L_{yx}\widetilde C_f^2\alpha^2\right)\left(1-\beta\mu_g\right)^S-1\right]\EE\left[\|y^k-y^*(x^k)\|^2\right]\\
   &~~~~~+\frac{L_f}{L_y}\left(1+\gamma+L_yL_f\alpha+ L_{yx}\widetilde C_f^2\alpha^2\right)S\beta^2\sigma_{g,1}^2\\
   &~~~~~ +\frac{L_f}{L_y}\left(L_y^2+L_{yx}\right)\alpha^2\tilde\sigma_f^2+\frac{L_f}{L_y}\left(L_y^2+L_{yx}+\frac{L_y^2}{\gamma}\right)\alpha^2\EE\left[\|\bar h_f^k\|_{P_x}^2\right]\\
   &\leq-\frac{\alpha}{2}\EE\left[\|\nabla F(x^k)\|_{P_x}^2\right]+\frac{L_f}{L_y}\left(1+\gamma+L_yL_f\alpha+ L_{yx}\widetilde C_f^2\alpha^2\right)\beta^2  S\sigma_{g,1}^2\\
   &~~~~~+\alpha b_k^2+\left[\frac{L_F}{2}+\frac{L_f}{L_y}\left(L_y^2+L_{yx}\right)\right]\alpha^2\tilde\sigma_f^2\\
   &~~~~~-\left[\frac{\alpha}{2}-\left(\frac{L_F}{2}+L_fL_y\left(1+\frac{1}{\gamma}\right)+\frac{L_fL_{yx}}{ L_y}\right)\alpha^2\right]\EE\left[\|\bar h_f^k\|_{P_x}^2\right]\\
   &~~~~~-\left(\frac{L_f\mu_g\beta}{L_y}-\alpha L_f^2-\frac{L_f\gamma}{L_y}-\frac{ L_fL_{yx}\tilde{C}_f^2\alpha^2}{L_y}\right)\EE\left[\|y^k-y^*(x^k)\|^2\right]. \numberthis\label{lay-minus-stoc}
\end{align*}
Selecting $\gamma=4L_fL_y\alpha$, \eqref{lay-minus-stoc} can be simplified by
\begin{align*}
    \EE[\mathbb{V}^{k+1}]-\EE[\mathbb{V}^k]&\leq-\frac{\alpha}{2}\EE\left[\|\nabla F(x^k)\|_{P_x}^2\right]+\frac{L_f}{L_y}\left(1+5L_fL_y\alpha+ L_{yx}\widetilde C_f^2\alpha^2\right)\beta^2  S\sigma_{g,1}^2\\
    &~~~~~+\alpha b_k^2+\left[\frac{L_F}{2}+\frac{L_f}{L_y}\left(L_y^2+L_{yx}\right)\right]\alpha^2\tilde\sigma_f^2\\
   &~~~~~-\left[\frac{\alpha}{4}-\left(\frac{L_F}{2}+L_fL_y+\frac{L_fL_{yx}}{ L_y}\right)\alpha^2\right]\EE\left[\|\bar h_f^k\|_{P_x}^2\right]\\
   &~~~~~-\left(\frac{L_f\mu_g\beta}{L_y}-5\alpha L_f^2-\frac{ L_fL_{yx}\tilde{C}_f^2\alpha^2}{L_y}\right)\EE\left[\|y^k-y^*(x^k)\|^2\right].\numberthis\label{minus-stoc}
\end{align*}

Choosing $\alpha\leq 1$,   the sufficient condition of making the last two terms in \eqref{minus-stoc} negative becomes
\begin{align}
    \alpha\leq\min\left(\frac{1}{2L_F+4L_fL_y+\frac{4L_f\mu_g\ell_{g,1}}{ L_y}}, \frac{\mu_g\beta}{5L_fL_y+ L_{yx}\tilde C_f^2}\right).\label{suff} 
\end{align}
Since we also need $\beta\leq\frac{1}{\ell_{g,1}}$, then the sufficient condition for \eqref{suff} becomes
\begin{align*}
    \alpha\leq\min\left(\frac{1}{2L_F+4L_fL_y+\frac{4L_f\mu_g\ell_{g,1}}{ L_y}}, \frac{\mu_g}{\ell_{g,1}(5L_fL_y+ L_{yx}\tilde C_f^2)}\right)~ ,~~~~ \beta=\frac{5L_fL_y+ L_{yx}\tilde C_f^2}{\mu_g}\alpha.
\end{align*}
Denoting 
\begin{align*}
    \bar\alpha_1=\frac{1}{2L_F+4L_fL_y+\frac{4L_fL_{yx}}{ L_y}},\qquad \bar\alpha_2=\frac{\mu_g}{\ell_{g,1}(5L_fL_y+ L_{yx}\tilde C_f^2)}
\end{align*}
and choosing $\alpha=\min\left(\bar\alpha_1,\bar\alpha_2,\frac{\bar\alpha}{\sqrt{K}}\right), \beta=\frac{5L_fL_y+ L_{yx}\tilde C_f^2}{\mu_g}\alpha$, then \eqref{minus-stoc} becomes
\begin{align}
    \frac{\alpha}{2}\EE[\|\nabla F(x^k)\|_{P_x}^2]\leq\left(\EE[\mathbb{V}_0^{k}]-\EE[\mathbb{V}_0^{k+1}]\right)+c_1S\alpha^2\sigma_{g,1}^2+c_2\alpha^2\tilde\sigma_f^2+\alpha b_k^2\label{51}
\end{align}
where $c_1$ and $c_2$ are defined as
\begin{align*}
    &c_1=\frac{L_f}{L_y}\left(1+5L_fL_y\alpha+ L_{yx}\widetilde C_f^2\alpha^2\right)\left(\frac{5L_fL_y+ L_{yx}\tilde C_f^2}{\mu_g\beta}\right)^2\\
    &c_2=\frac{L_F}{2}+\frac{L_f}{L_y}\left(L_y^2+L_{yx}\right). 
\end{align*}

Telescoping \eqref{51} and dividing both sides by $\frac{1}{2}\sum_{k=0}^{K-1}\alpha$ leads to
\begin{align*}
    \frac{\sum_{k=0}^{K-1}\alpha \EE\left[\|\nabla F(x^k)\|_{P_x}^2\right]}{\sum_{k=0}^{K-1}\alpha}\leq \frac{\mathbb{V}^{0}+\sum_{k=0}^{K-1} (\alpha b_{k}^{2}+c_{1} S\alpha \sigma_{g, 1}^{2}+c_{2}  \alpha \tilde{\sigma}_{f}^{2})}{\frac{1}{2} \sum_{k=0}^{K-1} \alpha K}.
\end{align*}

We then let $\bar\alpha, S={\cal O}(1)$ and $N={\cal O}(\log K)$ so that $b_k^2\leq {\cal O}(1/\sqrt{K})$ and $\tilde{\sigma}_f^2={\cal O}(N)={\cal O}(\log K)$, and thus, \begin{align*}
    \frac{1}{K}\sum_{k=0}^{K-1} \EE\left[\|\nabla F(x^k)\|_{P_x}^2\right]\leq \frac{\mathbb{V}^{0}+\sum_{k=0}^{K-1} (\alpha b_{k}^{2}+c_{1} S\alpha \sigma_{g, 1}^{2}+c_{2}  \alpha \tilde{\sigma}_{f}^{2})}{\frac{1}{2} \sum_{k=0}^{K-1} \alpha}=\tilde{\cal O}\left(\frac{1}{\sqrt{K}}\right).
\end{align*}

This implies that Algorithm \ref{stoc-alg} achieves an $\epsilon$-stationary point by $K=\tilde{\mathcal O}(\epsilon^{-2})$ iterations.
\end{proof} 

\section{Theoretical Analysis for {\sf E-AiPOD}}
In this section, we present the proof of Algorithm \ref{stoc-alg-skip}.
We define 
\begin{equation}
    \widetilde{\mathcal{F}}_{k}^t:=\sigma\{y^0,x^0,\cdots, y^{k+1},x^{k,1},\cdots,x^{k,t}\}
\end{equation}
where $\sigma\{\cdot\}$ denotes the $\sigma$-algebra generated by the random variables. Then it follows that $\widetilde{\mathcal{F}}_{k}^0=\sigma\{y^0,x^0,\cdots, y^{k+1}\}$.

\subsection{Descent of upper level}
First, we notice that the update of {\sf E-AiPOD} in \eqref{delta-x} can be written as
\begin{align*}
    x^{k+1}&\stackrel{(a)}{=}(1-\delta)x^k+\delta\operatorname{Proj}_{\mathcal{X}}\left(x^{k}-\alpha\sum_{t=0}^{T-1}h_f^{k,t}\right)\\
    &\stackrel{(b)}{=}(1-\delta)x^k+\delta(I-B^\dagger B)\left(x^{k}-\alpha\sum_{t=0}^{T-1}h_f^{k,t}\right)+\delta B^\dagger e\\
    &=(1-\delta)x^k+\delta\operatorname{Proj}_{\mathcal{X}}(x^k)-\alpha\delta\sum_{t=0}^{T-1}(I-B^\dagger B)h_f^{k,t}\\
    &\stackrel{(c)}{=}x^k-\alpha\delta(I-B^\dagger B)\left(\sum_{t=0}^{T-1}h_f^{k,t}\right)\numberthis\label{update-up-T}
\end{align*}
where $(a)$ comes from the update rule in \eqref{update-xkt} and \eqref{delta-x} , $(b)$ is derived from the closed form of $\operatorname{Proj}(\cdot)$ on the linear space $\mathcal{X}=\{x\mid Bx=e\}$, and  $(c)$ holds since $x^k\in\mathcal{X}$. 




We first quantify the bias induced by using the Hessian inverse vector product at a different point. 

\begin{lemma}[\bf Error of using delayed Hessian vector product]\label{hess-delay}
Define
\begin{align*} 
G(x,y,\tilde x)&:=\nabla_x f(x,y)+\left[\left(\nabla h(\tilde x)^\top A^{\dagger\top}\nabla_{yy}g(\tilde x,y)-\nabla_{xy}g(\tilde x,y)\right)\nonumber\right.\\
    &\left.\qquad\qquad\quad\times V_2(V_2^\top \nabla_{yy}g(\tilde x,y)V_2)^{-1} V_2^\top-\nabla h(\tilde x)^\top A^{\dagger\top}\right]\nabla_y f(\tilde x,y)
\end{align*}
as the gradient estimator using Hessian vector product at point $\tilde x$. We have
\begin{align*}
    &\|G(x,y,\tilde x)\|\leq\ell_{f,0}(1+L_y)\\
    &\|G(x,y,\tilde x)-\overline\nabla f(x,y)\|\leq L_G\|x-\tilde x\|
\end{align*}
where $L_G:=\frac{\left(1+\ell_{h,0}\|A^\dagger\|\right)\ell_{g,1}}{\mu_g}\left(\ell_{f,1}+\frac{\ell_{f,0}\ell_{g,2}}{\mu_g}+\frac{\mu_g\ell_{g,2}}{\ell_{g,1}}\right)+\left(\ell_{h,0}\ell_{f,1}+\ell_{h,1}(\ell_{g,1}+\ell_{f,0})\right)\|A^\dagger\|$.
\end{lemma}
\begin{proof}
First, since $G(x,y,\tilde x)$ only differs from  $\overline{\nabla}f(x,y)$ at the evaluation point $\tilde x$ of Hessian vector product, the bound for $\|G(x,y,\tilde x)\|$ can derived the same way as $\overline{\nabla}f(x,y)$ following Lemma \ref{lm-bound}, that is
\begin{align*}
    \|G(x,y,\tilde x)\|\leq\ell_{f,0}(1+L_y).
\end{align*}

Next, for any $x,y,\tilde x$, we have that
\begin{align*}
    \|G(x,y,\tilde x) &-\overline\nabla f(x,y)\|\\
    &\leq \|\left(\nabla h(\tilde x)^\top A^{\dagger\top}\nabla_{yy}g(\tilde x,y)-\nabla_{xy}g(\tilde x,y)\right)V_2(V_2^\top \nabla_{yy}g(\tilde x,y)V_2)^{-1} V_2^\top\nabla_y f(\tilde x,y)\\
    &~~~~~-\left(\nabla h( x)^\top A^{\dagger\top}\nabla_{yy}g( x,y)-\nabla_{xy}g( x,y)\right)V_2(V_2^\top \nabla_{yy}g( x,y)V_2)^{-1} V_2^\top\nabla_y f( x,y)\|\\
    &~~~~~+\|\nabla h(\tilde x)^\top A^{\dagger\top}\nabla_y f(\tilde x,y)-\nabla h(x)^\top A^{\dagger\top}\nabla_y f(x,y)\|\\
    &\leq \|\nabla h(x)^\top A^{\dagger\top}\nabla_{yy}g(x,y)-\nabla_{xy}g(x,y)\|\\
    &~~~~~\times\|V_2(V_2^\top \nabla_{yy}g(\tilde x,y)V_2)^{-1} V_2^\top\nabla_y f(\tilde x,y)-V_2(V_2^\top \nabla_{yy}g(x,y)V_2)^{-1} V_2^\top\nabla_y f(x,y)\|\\
    &~~~~~+\|V_2(V_2^\top \nabla_{yy}g(\tilde x,y)V_2)^{-1} V_2^\top\nabla_y f(\tilde x,y)\|\\
    &~~~~~\times\|\nabla h(\tilde x)^\top A^{\dagger\top}\nabla_{yy}g(\tilde x,y)-\nabla_{xy}g(\tilde x,y)-\nabla h(x)^\top A^{\dagger\top}\nabla_{yy}g(x,y)+\nabla_{xy}g(x,y)\|\\
    &~~~~~+\|\nabla h(\tilde x)^\top A^{\dagger\top}\|\|\nabla_y f(\tilde x,y)-\nabla_y f(x,y)\|+\|\nabla h(\tilde x)^\top A^{\dagger\top}-\nabla h( x)^\top A^{\dagger\top}\|\|\nabla_y f(x,y)\|\\
    &\stackrel{(a)}{\leq}\ell_{g,1}\left(1+\ell_{h,0}\|A^\dagger\|\right)\left(\|V_2(V_2^\top \nabla_{yy}g(\tilde x,y)V_2)^{-1} V_2^\top\|\|\nabla_y f(\tilde x,y)-\nabla_y f(x,y)\|\right.\\
    &~~~~~+\left.\|\nabla_y f(x,y)\|\|V_2((V_2^\top \nabla_{yy}g(\tilde x,y)V_2)^{-1}-(V_2^\top \nabla_{yy}g(x,y)V_2)^{-1})V_2^\top\|\right)\\
    &~~~~~+\|\nabla_{xy}g(\tilde x,y)-\nabla_{xy}g(x,y)\|+\|\nabla h(\tilde x)^\top A^{\dagger\top}\nabla_{yy}g(\tilde x,y)-\nabla h(x)^\top A^{\dagger\top}\nabla_{yy}g(x,y)\|\\
    &~~~~~+\ell_{h,0}\ell_{f,1}\|A^\dagger\|\|\tilde x-x\|+\ell_{h,1}\ell_{f,0}\|A^\dagger\|\|x-\tilde x\|\\
    &\stackrel{(b)}{\leq} \left[\frac{\left(1+\ell_{h,0}\|A^\dagger\|\right)\ell_{g,1}}{\mu_g}\left(\ell_{f,1}+\frac{\ell_{f,0}\ell_{g,2}}{\mu_g}+\frac{\mu_g\ell_{g,2}}{\ell_{g,1}}\right)+\left(\ell_{h,0}\ell_{f,1}+\ell_{h,1}(\ell_{g,1}+\ell_{f,0})\right)\|A^\dagger\|\right]\|x-\tilde x\|
\end{align*}
where (a) and (b) hold similarly with the derivation of \eqref{refer}. 
\end{proof}

Lemma \ref{hess-delay} shows the bias induced by evaluating the Hessian inverse vector product at a different point can be controlled by the point difference. 
We have the  counterpart of Lemma \ref{lm:var_stoc} as below. 

\begin{lemma}[\bf Bias and variance of gradient estimator]
Let $\tilde c=\frac{\mu_g}{\mu_g^2+\sigma_{g,2}^2}$ and define $$\bar h_f^{k,t}:=\EE[h_f^{k,t}| \widetilde{\mathcal{F}}_{k}^t], $$then $h_f^{k,t}$ is a biased estimator of upper-level gradient which satisfies that
\begin{align}
    &\|\bar h_f^{k,t}-G (x^{k,t},y^{k+1},x^{k})\|\leq b_{k};~~~\EE\left[\|h_f^{k,t}-\bar h_f^{k,t}\|^2|\widetilde{\mathcal{F}}_{k}^t\right]\leq \tilde\sigma_f^2=\mathcal{O}\left(N\kappa^{2}\right)\label{var-2-G}\\
    &\EE\left[\left\|\frac{1}{T}\sum_{t=0}^{T-1}(h_f^{k,t}-\bar h_f^{k,t})\right\|^2|\widetilde{\mathcal{F}}_{k}^0\right]\leq\frac{\tilde\sigma_f^2}{T}.\label{var-avg-G}
\end{align}
\label{lm:var_stoc-fed}
\end{lemma}
\begin{proof}
We omit the proof of  \eqref{var-2-G} since it is almost the same as the proof of Lemma \ref{lm:var_stoc}, and only prove \eqref{var-avg-G}. For \eqref{var-avg-G}, we have
\begin{align*}
    \EE\left[\left\|\frac{1}{T}\sum_{t=0}^{T-1}(h_f^{k,t}-\bar h_f^{k,t})\right\|^2|\widetilde{\mathcal{F}}_{k}^0\right]&=\frac{1}{T^2}\EE\left[\EE\left[\left\|\sum_{t=0}^{T-1}h_f^{k,t}-\bar h_f^{k,t}\right\|^2|\widetilde{\mathcal{F}}_{k}^{T-1}\right]|\widetilde{\mathcal{F}}_{k}^0\right]\\
    &\stackrel{(a)}{=}\frac{1}{T^2}\EE\left[\EE\left[\|h_f^{k,T-1}-\bar h_f^{k,T-1}\|^2+\left \|\sum_{t=0}^{T-2}h_f^{k,t}-\bar h_f^{k,t}\right\|^2|\widetilde{\mathcal{F}}_{k}^{T-1}\right]|\widetilde{\mathcal{F}}_{k}^0\right]\\
    &=\frac{1}{T^2}\EE\left[\|h_f^{k,T-1}-\bar h_f^{k,T-1}\|^2|\widetilde{\mathcal{F}}_{k}^0\right]+\frac{1}{T^2}\EE\left[\left\|\sum_{t=0}^{T-2}h_f^{k,t}-\bar h_f^{k,t}\right\|^2|\widetilde{\mathcal{F}}_{k}^0\right]\\
    &\stackrel{(b)}{=}\frac{1}{T^2}\sum_{t=0}^{T-1}\EE\left[\|h_f^{k,t}-\bar h_f^{k,t}\|^2|\widetilde{\mathcal{F}}_{k}^0\right]\leq\frac{\tilde\sigma_f^2}{T}
\end{align*}
where $(a)$ holds since $h_f^{k,T-1}-\bar h_f^{k,T-1}$ is independent from $\sum_{t=0}^{T-2}h_f^{k,t}-\bar h_f^{k,t}$ when given $\mathcal{F}^\star_{k,T-1}$; and (b) follows from applying the previous procedure $T-1$ times. 
\end{proof}

We then state a lemma controlling the drifting error of lazy projections. 

\begin{lemma}[\bf Drifting error of upper level]\label{drift-error}
Under Assumption \ref{as1}--\ref{ash}, it holds that for any $t$, 
\begin{align*}
    \EE\left[\|x^{k,t}-x^{k}\|^2\right]&\leq \alpha^2t^2\tilde C_f^2
\end{align*}
where $\widetilde C_f^2=2b_k^2+2\ell_{f,0}^2\left(1+L_y\right)^2+\tilde\sigma_f^2$. 
\end{lemma}
\begin{proof}
For any $t$, we know that
\begin{align*}
    \EE\left[\|x^{k,t+1}-x^{k,t}\|^2\right]&=\EE\left[\|x^{k,t+1}-x^{k,t}\|^2\right]=\EE\left[\|x^{k,t}-\alpha h_f^{k,t}-x^{k,t}\|^2\right]\\
    &=\alpha^2\EE\left[\|h_f^{k,t}\|^2\right]\stackrel{(a)}{\leq}\alpha^2\tilde C_f^2 \numberthis \label{C_f}
\end{align*}
where (a) is derived similarly from the bound for $\EE[\|h_f^k\|^2]$ in \eqref{J22} based on Lemma \ref{hess-delay} and Lemma \ref{lm:var_stoc-fed} and the constant is defined as $\widetilde C_f^2=2b_k^2+2\ell_{f,0}^2\left(1+L_y\right)^2+\tilde\sigma_f^2$. 

Then for any $t$, it holds that
\begin{align*}
    \EE\left[\|x^{k,t}-x^{k}\|^2\right]&=\EE\left[\|x^{k,t}-x^{k,t-1}+x^{k,t-1}-x^{k,t-2}+\cdots-x^k\|^2\right]\\
    &\leq t\sum_{\tau=0}^{t-1}\EE\left[\|x^{k,\tau+1}-x^{k,\tau}\|^2\right]\leq \alpha^2t^2\tilde C_f^2
\end{align*}
which completes the proof.
\end{proof}

\begin{lemma}\label{h_f^kt_error}
Under Assumption \ref{as1}--\ref{ash} and let $N={\cal O}(\log \alpha^{-1})$, it holds that 
\begin{align}
    \EE\left[\left\|\frac{1}{T}\sum_{t=0}^{T-1}\bar h_f^{k,t}-\overline\nabla f(x^k,y^{k+1})\right\|^2\right]={\cal O}\left(\alpha^2T^2\right).\label{reduction}
\end{align}
\end{lemma}
\begin{proof}
For any $t$, we have that
\begin{align*}
 \EE\left[\|\bar h_f^{k,t}-\overline{\nabla}f(x^{k},y^{k+1})\|^2\right] 
    &\leq 3\EE[\|\bar h_f^{k,t}-G(x^{k,t},y^{k+1},x^{k})\|^2]+3\EE[\|\overline\nabla f(x^{k,t},y^{k+1})\!-\!\overline\nabla f(x^k,y^{k+1})\|^2]\\
    &~~~~+3\EE[\|G(x^{k,t},y^{k+1},x^{k})-\overline\nabla f(x^{k,t},y^{k+1})\|^2]\\
    &\stackrel{(a)}{\leq} 3b_k^2+3L_G^2\EE[\|x^{k,t}-x^k\|^2]+3L_f^2\EE[\|x^{k,t}-x^k\|^2]\\
    &\stackrel{(b)}{\leq} 3b_k^2+3\alpha^2(L_G^2+L_f^2)t^2\tilde C_f^2\stackrel{(c)}{=}{\cal O}(\alpha^2 t^2)\numberthis\label{123}
\end{align*}
where $(a)$ is due to Lemma \ref{hess-delay}, $(b)$ comes from Lemma \ref{drift-error} and $(c)$ holds by \eqref{bias-1} and $N={\cal O}(\log\alpha^{-1})$. 

Using the fact that $\left\|\frac{1}{T}\sum_{t=0}^{T-1}z_t \right\|^2=\frac{1}{T^2}\left\|\sum_{t=0}^{T-1}z_t\right\|^2\leq\frac{1}{T}\sum_{t=0}^{T-1}\|z_t\|^2$, from \eqref{123}, we have 
\begin{align}
    \EE\left[\left\|\frac{1}{T}\sum_{t=0}^{T-1}\bar h_f^{k,t}-\overline\nabla f(x^k,y^{k+1})\right\|^2\right]= {\cal O}\left(\alpha^2T^2\right) 
\end{align}
from which the proof is complete. 
\end{proof}

\begin{lemma}[\bf Descent of upper level]\label{up-new}
Under Assumption \ref{as1}--\ref{ash}, if we define $\bar{H}_f^k:=\frac{1}{T}\sum_{t=0}^{T-1} \bar h_f^{k,t}$,
it holds that
\begin{align*}
    \EE[F(x^{k+1})]&\leq\EE[F(x^{k})]-\frac{\alpha T}{2}\EE[\|\nabla F(x^k)\|^2_{P_x}]-\left(\frac{\alpha T}{2}-\frac{\alpha^2L_F T^2}{2}\right)\EE[\|\bar{H}_f^k\|^2_{P_x}]+\frac{\alpha^2L_F T \tilde \sigma_f^2 }{2}\\
    &~~~~+\alpha TL_y^2\EE\left[\|y^{k+1}-y^*(x^k)\|^2\right]+{\cal O}(\alpha^3 T^3).
\end{align*}
\end{lemma}
\begin{proof}
First, we have 
\begin{align*}
    \EE\left[\left\|\frac{1}{T}\sum_{t=0}^{T-1} h_f^{k,t}\right\|_{P_x}^2\right]&=\EE\left[\left\|\frac{1}{T}\sum_{t=0}^{T-1} h_f^{k,t}-\bar h_f^{k,t}+\bar h_f^{k,t}\right\|_{P_x}^2\right]\\
    &\stackrel{(a)}{=}\EE\left[\left\|\frac{1}{T}\sum_{t=0}^{T-1} \bar h_f^{k,t}\right\|_{P_x}^2\right]+\EE\left[\left\|\frac{1}{T}\sum_{t=0}^{T-1} h_f^{k,t}-\bar h_f^{k,t}\right\|_{P_x}^2\right]\\
    &\stackrel{(b)}{\leq} \EE\left[\|\bar{H}_f^k\|_{P_x}^2\right]+\frac{\tilde \sigma_f^2}{T}\numberthis\label{var--de}
\end{align*}
where (a) follows from Lemma \ref{lm:matrix} and (b) results from \eqref{var-avg-G}. 

Moreover, it follows that
\begin{align*}
    \EE\left[\langle\nabla F(x^k),(I-B^\dagger B)\frac{1}{T}\sum_{t=0}^{T-1}h_f^{k,t}\rangle\right]&=
    \EE\left[\frac{1}{T}\sum_{t=0}^{T-1}\EE\left[\langle\nabla F(x^k),(I-B^\dagger B)h_f^{k,t}\rangle|\widetilde{\mathcal{F}}_{k}^t\right]\right]\\
    &=\EE\left[\langle\nabla F(x^k),(I-B^\dagger B)\frac{1}{T}\sum_{t=0}^{T-1}\bar h_f^{k,t}\rangle\right]\\
    &=\EE\left[\langle\nabla F(x^k),(I-B^\dagger B)\bar{H}_f^{k}\rangle\right].\numberthis\label{expec}
\end{align*}

Taking the expectation of $F(x^{k+1})$, we get
\begin{align*}
    \EE\left[F(x^{k+1})\right]
    &\stackrel{(a)}{\leq} \EE\left[F(x^k)\right]+\EE\left[\langle \nabla F(x^k),x^{k+1}-x^k\rangle\right]+\frac{L_F}{2}\EE\left[\|x^{k+1}-x^k\|^2\right]\\
    &\stackrel{(b)}{\leq} \EE[F(x^k)]-\alpha \delta T\EE\left[\langle \nabla F(x^k),(I-B^\dagger B)\frac{1}{T}\sum_{t=0}^{T-1}h_f^{k,t}\rangle\right]+\frac{L_F}{2}\alpha^2T^2\delta^2 \EE\left[\|\frac{1}{T}\sum_{t=0}^T h_f^{k,t}\|_{P_x}^2\right]\\
    &\stackrel{(c)}{\leq} \EE[F(x^k)]-\alpha \delta T\EE\left[\langle \nabla F(x^k),(I-B^\dagger B)\bar{H}_f^k\rangle\right]+\frac{L_F}{2}\alpha^2T^2\delta^2 \EE\left[\|\bar{H}_f^k\|_{P_x}^2\right]+\frac{L_F}{2}\alpha^2T\delta^2\tilde\sigma_f^2\\
    &\stackrel{(d)}{\leq} \EE[F(x^k)]-\frac{\alpha\delta T}{2}\EE[\|\nabla F(x^k)\|^2_{P_x}]-\left(\frac{\alpha\delta T}{2}-\frac{\alpha^2L_F T^2\delta^2}{2}\right)\EE[\|\bar{H}_f^k\|^2_{P_x}]\\
    &~~~~+\frac{\alpha\delta T}{2}\EE\left[\|\nabla F(x^k)-\bar{H}_f^k\|^2\right]+\frac{\alpha^2\delta^2L_F T \tilde \sigma_f^2 }{2}\\
    &\leq \EE[F(x^{k})]-\frac{\alpha\delta T}{2}\EE[\|\nabla F(x^k)\|^2_{P_x}]-\left(\frac{\alpha\delta T}{2}-\frac{\alpha^2\delta^2L_F T^2}{2}\right)\EE[\|\bar{H}_f^k\|^2_{P_x}]+\frac{\alpha^2\delta^2L_F T \tilde \sigma_f^2 }{2}\\
    &~~~~+\alpha\delta T\EE\left[\|\nabla F(x^k)-\overline\nabla f(x^k,y^{k+1})\|^2\right]+\alpha\delta T\EE\left[\|\overline\nabla f(x^k,y^{k+1})-\bar{H}_f^k\|^2\right]\\
    &\stackrel{(e)}{\leq} \EE[F(x^{k})]-\frac{\alpha\delta T}{2}\EE[\|\nabla F(x^k)\|^2_{P_x}]-\left(\frac{\alpha\delta T}{2}-\frac{\alpha^2\delta^2 L_F T^2}{2}\right)\EE[\|\bar{H}_f^k\|^2_{P_x}]+\frac{\alpha^2\delta^2 L_F T \tilde \sigma_f^2 }{2}\\
    &~~~~+\alpha T\delta L_y^2\EE\left[\|y^{k+1}-y^*(x^k)\|^2\right]+{\cal O}(\alpha^3 T^3\delta)
\end{align*}
where (a) comes from the smoothness of $F$ and the update \eqref{update-up-T}, (b) is derived from \eqref{update-up-T}, and (c) results from \eqref{var--de} and \eqref{expec}, (d) comes from $2a^\top b= \|a\|^2+\|b\|^2-\|a-b\|^2$ and $\|I-B^\dagger B\|\leq 1$, (e) is derived from Lipschitz continuity of $\overline\nabla f(x,y)$, $F(x)=\overline\nabla f(x,y^*(x))$ and Lemma \ref{h_f^kt_error}. This completes the proof. 
\end{proof}

\subsection{Error of lower level}
\begin{lemma}[\bf Lipschitz continuity and smoothness of the $r^*(x)$]\label{smooth-r}
Under Assumption \ref{as1}--\ref{as2}, the projection offset $r^*(x)$ is $L_r$-Lipschitz continuous and $L_{rx}$-smooth with constants defined as 
\begin{align*}
    L_r:=\ell_{g,1}\left(1+L_y\right),~~~~{\rm and}~~~~ L_{rx}:=\ell_{g,2}(1+L_y)^2+\ell_{g,1}L_{yx}.
\end{align*}
\end{lemma}
\begin{proof}
Recall the definition of $r^*(x):=\nabla_y g(x,y^*(x))$, then for any $x_1,x_2$,
\begin{align*}
    \|r^*(x_1)-r^*(x_2)\|&\leq\|\nabla_y g(x_1,y^*(x_1))-\nabla_y g(x_2,y^*(x_2))\|\\
    &\leq \ell_{g,1}\left(\|x_1-x_2\|+\|y^*(x_1)-y^*(x_2)\|\right)\\
    &\leq \ell_{g,1}\left(1+L_y\right)\|x_1-x_2\|=L_r\|x_1-x_2\|. 
\end{align*}
Using the chain rule, we can obtain the gradient of $r^*(x)$ as
\begin{align*}
    \nabla r^*(x)=\nabla_{yx}g(x,y^*(x))+\nabla_{yy}g(x,y^*(x))\nabla y^*(x). 
\end{align*}
According to the Lipschitz continuity of $\nabla y^*(x)$ and $\nabla^2 g$, we get for any $x_1,x_2$
\begin{align*}
    \|\nabla r^*(x_1)-\nabla r^*(x_2)\|&\leq \|\nabla_{yx}g(x_1,y^*(x_1))-\nabla_{yx}g(x_1,y^*(x_1))\|\\
    &~~~~~+\|\nabla_{yy}g(x_1,y^*(x_1))\nabla y^*(x_1)-\nabla_{yy}g(x_2,y^*(x_2))\nabla y^*(x_2)\|\\
    &\stackrel{(a)}{\leq} \ell_{g,2}(1+L_y)\|x_1-x_2\|+\|\nabla_{yy}g(x_1,y^*(x_1))\|\|\nabla y^*(x_1)-\nabla y^*(x_2)\|\\
    &~~~~~+\|\nabla y^*(x_2)\|\|\nabla_{yy}g(x_1,y^*(x_1))-\nabla_{yy}g(x_2,y^*(x_2))\|\\
    &\leq \left(\ell_{g,2}(1+L_y)^2+\ell_{g,1}L_{yx}\right)\|x_1-x_2\|=L_{rx}\|x_1-x_2\|
\end{align*}
where (a) is derived from \eqref{Hessian-y*} and Lemma \ref{y-smooth}. 
\end{proof}

\begin{lemma}[\bf{Error of lower-level update}]\label{lm-low-stoc-skip}
Suppose that Assumption \ref{as1}--\ref{ash} hold and $\beta\leq\frac{1}{\ell_{g,1}}$, then the error of lower-level update can be bounded by
\begin{subequations}
\small
\begin{align}
    &\EE\left[\|y^{k+1}-y^*(x^k)\|^2+\frac{\beta^2}{p^2}\|r^{k+1}-r^*(x^k)\|^2\right]\nonumber\\
    &~~~~~~~~~~~~~~~~~~~~~~~~~~~~~~~~~~~~~~~\leq\left(1-\nu\right)^S\EE\left[\|y^{k}-y^*(x^k)\|^2+\frac{\beta^2}{p^2}\|r^{k}-r^*(x^k)\|^2\right]+S\beta^2\sigma_{g,1}^2\label{46a-new}\\
    &\EE[\|y^{k+1}-y^*(x^{k+1})\|^2]\leq \left(1+\gamma +2 L_{yx}\widetilde C_f^2T^2\alpha^2\delta^2\right)\EE[\|y^{k+1}-y^*(x^{k})\|^2]\nonumber\\
    &\qquad\qquad\qquad\qquad\qquad\quad+\left(L_y^2+L_{yx}\right)\alpha^2\delta^2T\tilde\sigma_f^2+\left(L_y^2+L_{yx}+\frac{L_y^2}{\gamma}\right)\alpha^2\delta^2T^2\EE\left[\|\bar{H}_f^k\|_{P_x}^2\right]\label{46b-new}\\
    &\EE[\|r^{k+1}-r^*(x^{k+1})\|^2]\leq \left(1+\gamma +2 L_{rx}\widetilde C_f^2T^2\alpha^2\delta^2\right)\EE[\|r^{k+1}-r^*(x^{k})\|^2]\nonumber\\
    &\qquad\qquad\qquad\qquad\qquad\quad+\left(L_r^2+L_{rx}\right)\alpha^2\delta^2 T\tilde\sigma_f^2+\left(L_r^2+L_{rx}+\frac{L_r^2}{\gamma}\right)\alpha^2\delta^2T^2\EE\left[\|\bar{H}_f^k\|_{P_x}^2\right]\label{46c-new}
\end{align}
\end{subequations}
where $\widetilde C_f^2$ is defined in Lemma \ref{lm-low-stoc}, $\nu:=\min\left\{\beta\mu_g,p^2\right\}$ , $\gamma$ is the balancing parameter that will be chosen in the final theorem. 
\end{lemma}

\begin{proof}
First, for a given $x^k$, defining $\nu:=\min\left(\beta\mu_g,p^2\right)$ and applying Lemma C.1 and Lemma C.2 in \citep{mishchenko2022proxskip}, we can obtain that
\begin{align*}
    &\EE\left[\|y^{k,s+1}-y^*(x^k)\|^2+\frac{\beta^2}{p^2}\|r^{k,s+1}-r^*(x^k)\|^2\mid \mathcal{F}_{k,s}\right]\\
    \leq &(1-\nu) \left[\|y^{k,s}-y^*(x^k)\|^2+\frac{\beta^2}{p^2}\|r^{k,s}-r^*(x^k)\|^2\right]+\beta^2\sigma_{g,1}^2. \numberthis\label{linear}
\end{align*}
Then taking the expectation of the both sides of \eqref{linear} and telescoping it, we can arrive at \eqref{46a-new}. 


Next, proof of \eqref{46b-new} and \eqref{46c-new} are similar with only difference on Lipschitz constant, so that we only prove \eqref{46c-new}. For \eqref{46c-new}, we have
\begin{align*}
     \|r^{k+1}-r^*(x^{k+1})\|^2&= \|r^{k+1}-r^*(x^{k})\|^2+ \underbrace{\|r^{*}(x^{k})-r^*(x^{k+1})\|^2}_{J_1}\\
     &~~~~~+2\underbrace{\langle r^{k+1}-r^*(x^{k}),r^{*}(x^{k})-r^*(x^{k+1})\rangle}_{J_2}.
\end{align*}

Since $r^*(x)$ is $L_r$ Lipschitz continuous according to Lemma \ref{smooth-r}, $J_1$ can be bounded by
\begin{align*}
    \EE\left[J_1\right]&\leq L_{r}^2\EE\left[\|x^{k+1}-x^k\|^2\right]\stackrel{(a)}{=}  \alpha^2\delta^2L_{r}^2\EE\left[\|\sum_{t=0}^{T-1}h_f^k\|_{P_x}^2\right]\\
    &\stackrel{(b)}{\leq} \alpha^2\delta^2L_{r}^2\left(T^2\EE\left[\|\bar{H}_f^k\|_{P_x}^2\right]+T\tilde\sigma_f^2\right)\numberthis\label{J1-new}
\end{align*}
where (a) comes from \eqref{linear-update}, (b) is attained by \eqref{var--de}. 

On the other hand, we can decompose $J_2$ by two terms as follows.
\begin{align*}
    J_2&=\underbrace{-\langle r^{k+1}-r^*(x^{k}), \nabla r^*(x^k)^\top (x^{k+1}-x^k) \rangle}_{J_{2,1}}\\
    &~~~~\underbrace{-\langle r^{k+1}-r^*(x^{k}), r^*(x^{k+1})-r^*(x^k)-\nabla r^*(x^k)^\top(x^{k+1}-x^k) \rangle}_{J_{2,2}}.
\end{align*}
Moreover, the conditional expectation of $J_{2,1}$ can be bounded by
\begin{align*}
    \EE[J_{2,1}]&=-\EE[\langle r^{k+1}-r^*(x^{k}),\nabla r^*(x^k)^\top (x^{k+1}-x^k)]\rangle\\
    &=-\alpha\delta\EE[\langle r^{k+1}-r^*(x^{k}),\nabla r^*(x^k)^\top(I-B^\dagger B) (\sum_{t=0}^{T-1}h_f^{k,t})\rangle]\\
    &=-\alpha\delta\EE\left[\sum_{t=0}^{T-1}\EE[\langle r^{k+1}-r^*(x^{k}),\nabla r^*(x^k)^\top(I-B^\dagger B) h_f^{k,t}\rangle|\widetilde{\mathcal{F}}_{k}^t]\right]\\
    &\leq -\alpha\delta T\EE[\langle r^{k+1}-r^*(x^{k}),\nabla r^*(x^k)^\top (I-B^\dagger B)\bar{H}_f^k\rangle]\\
    &\stackrel{(a)}{\leq} \frac{\gamma }{2}\EE[\|r^{k+1}-r^*(x^{k})\|^2]+\frac{\alpha^2\delta^2T^2 L_r^2}{2\gamma}\EE[\|\bar{H}_f^k\|_{P_x}^2].\numberthis\label{J21-new}
\end{align*}

Based on the smoothness of $r^*(x)$ in Lemma \ref{smooth-r} and Young's inequality, $J_{2,2}$ can be bounded by
\begin{align*}
    \EE[J_{2,2}]&\leq \EE\left[\|r^{k+1}-r^*(x^k)\|\|r^*(x^{k+1})-r^*(x^k)-\nabla r^*(x^k)^\top(x^{k+1}-x^k)\|^2\right]\\
    &\leq \frac{L_{rx}}{2}\EE\left[\|r^{k+1}-r^*(x^k)\|\|x^{k+1}-x^k\|^2\right]\\
    &\leq \frac{ L_{rx}\alpha^2\delta^2}{2}\EE\left[\|r^{k+1}-r^*(x^k)\|^2\| \sum_{t=0}^{T-1}h_f^{k,t}\|_{P_x}^2\right]+\frac{L_{rx}\alpha^2\delta^2}{2}\EE\left[\|\sum_{t=0}^{T-1}h_f^{k,t}\|_{P_x}^2\right]\\
    &\leq \frac{ L_{rx}\alpha^2\delta^2 T}{2}\EE\left[\|r^{k+1}-r^*(x^k)\|^2\sum_{t=0}^{T-1}\EE[\|  h_f^{k,t}\|^2|\widetilde{\mathcal{F}}_{k}^t]\right]+\frac{L_{rx}\alpha^2\delta^2T^2}{2}\left(\EE\left[\|\bar{H}_f^k\|^2_{P_x}\right]+\frac{\tilde\sigma_f^2}{T}\right)\\
    &\leq  L_{rx}\alpha^2\delta^2T\EE\left[\|r^{k+1}-r^*(x^k)\|^2\sum_{t=0}^{T-1}(\EE[\|\bar h_f^{k,t}\|^2|\widetilde{\mathcal{F}}_{k}^t]+\tilde\sigma_f^2)\right]+\frac{L_{rx}\alpha^2\delta^2T^2}{2}\left(\EE[\|\bar{H}_f^k\|_{P_x}^2]+\frac{\tilde\sigma_f^2}{T}\right)\\
    &\stackrel{(a)}{\leq}  L_{rx}\alpha^2\delta^2\tilde C_f^2T^2\EE\left[\|r^{k+1}-r^*(x^k)\|^2\right]+\frac{L_{rx}\alpha^2\delta^2T^2}{2}\left(\EE[\|\bar{H}_f^k\|_{P_x}^2]+\frac{\tilde\sigma_f^2}{T}\right)\numberthis\label{J22-new}
\end{align*}
where $(a)$ comes from
\begin{align*}
    \EE[\|\bar h_f^{k,t}\|^2|\widetilde{\mathcal{F}}_{k}^t]&\leq 2\EE[\|\bar h_f^{k,t}-G(x^{k,t},y^{k+1},x^k)\|^2|\widetilde{\mathcal{F}}_{k}^t]+2\EE[\|G(x^{k,t},y^{k+1},x^k)\|^2|\widetilde{\mathcal{F}}_{k}^t]\\
    &\leq 2b_k^2+2\ell_{f,0}^2(1+L_y)^2
\end{align*}
and $\tilde C_f^2=2b_k^2+2\ell_{f,0}^2(1+L_y)^2+\tilde\sigma_f^2$. Then combining \eqref{J1-new}, \eqref{J21-new} and \eqref{J22-new}, we get 
\begin{align*}
    \EE[\|r^{k+1}-r^*(x^{k+1})\|^2]&\leq \left(1+\gamma +2 L_{rx}\widetilde C_f^2T^2\alpha^2\delta^2\right)\EE[\|r^{k+1}-r^*(x^{k})\|^2]+\left(L_r^2+L_{rx}\right)\alpha^2\delta^2T\tilde\sigma_f^2\\
    &~~~~~~+\left(L_r^2+L_{rx}+\frac{L_r^2}{\gamma}\right)\alpha^2\delta^2T^2\EE\left[\|\bar{H}_f^k\|_{P_x}^2\right]
\end{align*}
which completes the proof for \eqref{46c-new}. With similar proof for \eqref{46c-new}, we can prove \eqref{46b-new}.
\end{proof}

\subsection{Proof of Theorem \ref{alskip-conv}}\label{sec:proxskip-conv}



\begin{proof}
First, without loss of generality, we can assume that $\ell_{g,1}\geq 1$ so that $L_r\geq L_y, L_{rx}\geq L_{yx}$ and plugging \eqref{46b-new}, \eqref{46c-new} into \eqref{46a-new} in Lemma \ref{lm-low-stoc-skip}, we get that
\begin{align*}
    &~~~~\EE\left[\|y^{k+1}-y^*(x^{k+1})+\frac{\beta^2}{p^2}\|r^{k+1}-r^*(x^{k+1})\|^2\|^2\right]\\
    &\leq \left(1+\gamma +2 L_{rx}\widetilde C_f^2\alpha^2\delta^2T^2\right)\left(1-\nu\right)^S\EE\left[\|y^{k}-y^*(x^{k})+\frac{\beta^2}{p^2}\|r^{k}-r^*(x^{k})\|^2\|^2\right]\\
    &~~~~~+\left(1+\gamma +2 L_{rx}\widetilde C_f^2\alpha^2\delta^2T^2\right)S\beta^2\sigma_{g,1}^2+\left(L_r^2+L_{rx}\right)\left(1+\frac{\beta^2}{p^2}\right)\alpha^2\delta^2T\tilde\sigma_f^2\\
    &~~~~~+\left(L_r^2+L_{rx}+\frac{L_r^2}{\gamma}\right)\left(1+\frac{\beta^2}{p^2}\right)\alpha^2\delta^2T^2\EE\left[\|\bar H_f^k\|_{P_x}^2\right].\numberthis\label{2yr*}
\end{align*}
Then using Lyapunov function defined in \eqref{lyapunov-skip} and applying \eqref{2yr*}, Lemma \ref{up-new} and Lemma \ref{lm-low-stoc-skip}, we get
\begin{align*}
   & \EE\left[\mathbb{V}_1^{k+1}\right]-\EE\left[\mathbb{V}_1^k\right]\\
    &\leq-\frac{\alpha\delta T}{2}\EE\left[\|\nabla F(x^k)\|_{P_x}^2\right]+\alpha\delta T L_f^2\EE\left[\|y^{k+1}-y^*(x^k)\|^2\right]+{\cal O}(\alpha^3 T^3\delta)\\
   &~~~~~ -\left(\frac{\alpha\delta T}{2}-\frac{L_F\alpha^2\delta^2 T^2}{2}\right) \EE\left[\|\bar{H}_f^k\|_{P_x}^2\right]+\frac{L_F\alpha^2\delta^2 T \tilde\sigma_f^2}{2}\\
   &~~~~~ +\frac{L_f}{L_r}\left[\left(1+\gamma +2 L_{rx}\widetilde C_f^2\alpha^2\delta^2T^2\right)\left(1-\nu\right)^S-1\right]\EE\left[\|y^k-y^*(x^k)\|^2+\frac{\beta^2}{p^2}\|r^k-r^*(x^k)\|^2\right]\\
   &~~~~~+\frac{L_f}{L_r}\left(1+\gamma +2 L_{rx}\widetilde C_f^2\alpha^2\delta^2T^2\right)S\beta^2\sigma_{g,1}^2  +\frac{L_f}{L_r}\left(L_r^2+L_{rx}\right)\left(1+\frac{\beta^2}{p^2}\right)\alpha^2\delta^2T\tilde\sigma_f^2\\
   &~~~~~~+\frac{L_f}{L_r}\left(L_r^2+L_{rx}+\frac{L_r^2}{\gamma}\right)\left(1+\frac{\beta^2}{p^2}\right)\alpha^2\delta^2T^2\EE\left[\|\bar{H}_f^k\|_{P_x}^2\right]\\
   &\leq-\frac{\alpha\delta T}{2}\EE\left[\|\nabla F(x^k)\|_{P_x}^2\right]+\frac{L_f}{L_r}\left(1+\gamma +L_rL_f\alpha\delta T+2 L_{rx}\widetilde C_f^2\alpha^2\delta^2T^2\right)\beta^2 S\sigma_{g,1}^2+{\cal O}(\alpha^3T^3\delta)\\
   &~~~~~+\left[\frac{L_F}{2}+\frac{L_f}{L_r}\left(L_r^2+L_{rx}\right)\left(1+\frac{\beta^2}{p^2}\right)\right]\alpha^2T\delta^2\tilde\sigma_f^2\\
   &~~~~~-\left[\frac{\alpha\delta T}{2}-\left(\frac{L_F}{2}+L_fL_r\left(1+\frac{1}{\gamma}\right)\left(1+\frac{\beta^2}{p^2}\right)+\frac{L_fL_{rx}}{ L_r}\left(1+\frac{\beta^2}{p^2}\right)\right)\alpha^2\delta^2T^2\right]\EE\left[\|\bar H_f^k\|_{P_x}^2\right]\\
   &~~~~~-\left(\frac{L_f\nu}{L_r}-\alpha\delta T L_f^2-\frac{L_f\gamma }{L_r}-\frac{2 L_fL_{rx}\tilde{C}_f^2\alpha^2\delta^2T^2}{L_r}\right)\EE\left[\|y^k-y^*(x^k)\|^2+\frac{\beta^2}{p^2}\|r^k-r^*(x^k)\|^2\right]. \numberthis\label{lay-minus-stoc-r2}
\end{align*}
Selecting $\gamma=4L_fL_r\alpha\delta T, p=\sqrt{\beta\mu_g}$, \eqref{lay-minus-stoc-r2} can be simplified by
\begin{align*}
&    \EE[\mathbb{V}_1^{k+1}]-\EE[\mathbb{V}_1^k]\\
    &\leq-\frac{\alpha\delta T}{2}\EE\left[\|\nabla F(x^k)\|_{P_x}^2\right]+\frac{L_f}{L_r}\left(1+5L_fL_r\alpha\delta T+2  L_{rx}\widetilde C_f^2\alpha^2\delta^2 T^2\right)\beta^2  S\sigma_{g,1}^2+{\cal O}(\alpha^3 T^3\delta)\\
    &~~~~+\left[\frac{L_F}{2}+\frac{L_f}{L_r}\left(L_r^2+L_{rx}\right)\left(1+\frac{\beta}{\mu_g}\right)\right]\alpha^2\delta^2T\tilde\sigma_f^2\\
   &~~~~-\left[\frac{\alpha\delta T}{4}-\left(\frac{L_F}{2}+L_fL_r+\frac{L_fL_{rx}}{  L_r}\right)\alpha^2\delta^2T^2-\frac{\beta\alpha\delta T}{4\mu_g}-\left(L_fL_r+\frac{L_fL_{rx}}{  L_r}\right)\frac{\beta\alpha^2\delta^2T^2}{\mu_g}\right]\EE\left[\|\bar H_f^k\|_{P_x}^2\right]\\
   &~~~~-\left(\frac{L_f\mu_g\beta}{L_r}-5\alpha\delta L_f^2T-\frac{2  L_fL_{rx}\tilde{C}_f^2\alpha^2\delta^2T^2}{L_r}\right)\EE\left[\|y^k-y^*(x^k)\|^2+\|r^k-r^*(x^k)\|^2\right].\!\!\numberthis\label{minus-stoc-r}
\end{align*}

Let $\alpha\delta T\leq 1$. Since we also need $\beta\leq\frac{1}{\ell_{g,1}}$, the sufficient condition of making the last two terms in \eqref{minus-stoc-r} negative becomes
\begin{align*}
    &\alpha\delta\leq\frac{\bar\alpha}{T}~ ,~~~~~~~\beta=\frac{5L_fL_r+2  L_{rx}\tilde C_f^2}{\mu_g}\alpha\delta T
\end{align*}
where $\bar\alpha=\min\left(\bar\alpha_1,\bar\alpha_2\right)$ and $\bar\alpha_1,\bar\alpha_2$ are defined as
\begin{align*}
    &\bar\alpha_1=\frac{1}{2L_F+4L_fL_r+\frac{4L_fL_{rx}}{  L_r}+\frac{\left(5L_fL_y+  L_{yx}\tilde C_f^2\right)\left(1+4L_fL_r+\frac{4L_fL_{rx}}{  L_r}\right)}{\mu_g^2}},~\bar\alpha_2=\frac{\mu_g}{\ell_{g,1}(5L_fL_r+  L_{rx}\tilde C_f^2)}\numberthis\label{baralpha}.
\end{align*}
Then \eqref{minus-stoc-r} becomes
\begin{align}
\!\!\!     \frac{\alpha\delta T}{2}\EE[\|\nabla F(x^k)\|_{P_x}^2]\leq\left(\EE[\mathbb{V}_1^{k}]-\EE[\mathbb{V}_1^{k+1}]\right)+c_1S\alpha^2\delta^2T^2\sigma_{g,1}^2+c_2\alpha^2\delta^2T\tilde\sigma_f^2+{\cal O}(\alpha^3\delta^3T^2)\label{51-r}
\end{align}
where $c_1$ and $c_2$ are defined as
\begin{align}
    c_1=\frac{L_f S}{L_r}\left(1+  L_{rx}\widetilde C_f^2\bar\alpha^2\right)\left(\frac{5L_fL_r+2  L_{rx}\tilde C_f^2}{\mu_g}\right)^2~~~{\rm and}~~c_2=\frac{L_F}{2}+\frac{L_f}{L_r}\left(L_r^2+L_{rx}\right)\numberthis\label{c1c2}
\end{align}
Telescoping \eqref{51-r} and dividing both sides by $\alpha\delta TK$ leads to
\begin{align*}
    \frac{1}{K}\sum_{k=0}^{K-1}\EE\left[\|\nabla F(x^k)\|_{P_x}^2\right]\leq \frac{2(\mathbb{V}_1^{0}-F^*)}{\alpha \delta T K}+2c_1\sigma_{g,1}^2\alpha\delta T+2c_2\tilde\sigma_f^2\alpha\delta+{\cal O}(\alpha^2\delta^2 T).
\end{align*}
This completes the proof.
\end{proof}
\vspace{-0.2cm}

\section{Theoretical Analysis for {\sf E2-AiPOD}}
This section presents the proof of {\sf E2-AiPOD}. 
Since we introduce a new sequence $u^k$, we define
\begin{align*}
    &\overline{\mathcal{F}}_{k}^n:=\sigma\{y^0,u^0,x^0,\cdots, y^{k+1},u^{k,0},\cdots,u^{k,n}\}
\end{align*}
where $\sigma\{\cdot\}$ denotes the $\sigma$-algebra generated by the random variables. For simplicity, we also denote $$\bar d_f^k:=\EE[d_f^k|\overline{\mathcal{F}}_{k}^{N}]=\nabla_x f(x^k,y^{k+1})-\nabla_{xy}g(x^k,y^{k+1})u^{k+1}.$$

\subsection{Supporting lemmas of Theorem \ref{thm:conv-E2}}
First, we prove Lemma \ref{lm:u*-strong} which states $u^*(x,y)$ is the solution of a strongly convex minimization problem with linear constraints. 
\begin{customlm}{4}
$u^*(x,y)$ is the minimizer of a $\mu_g$- strongly convex problem, i.e.
$$u^*(x,y)=\argmin_{u\in\{u|V_1^\top u=0\}}\frac{1}{2}\|\nabla_{yy}^{-\frac{1}{2}}g(x,y)\nabla_y f(x,y)+\nabla_{yy}^{\frac{1}{2}}g(x,y)u\|^2.$$
\end{customlm}
\begin{proof}
We denote the optimal solution mapping as 
$$\tilde u(x,y)=\argmin_{u\in\{u|V_1^\top u=0\}}\frac{1}{2}\|\nabla_{yy}^{-\frac{1}{2}}g(x,y)\nabla_y f(x,y)+\nabla_{yy}^{\frac{1}{2}}g(x,y)u\|^2.$$
According to the optimality condition, we know that $\tilde u(x,y)$ satisfies
\begin{align*}
    \tilde u(x,y)&=\operatorname{Proj}_{\{u|V_1^\top u=0\}}\left(\tilde u(x,y)-(\nabla_y f(x,y)+\nabla_{yy}g(x,y)\tilde u(x,y))\right)\\
    &\stackrel{(a)}{=}(I-V_1V_1^\top)\left(\tilde u(x,y)-(\nabla_y f(x,y)+\nabla_{yy}g(x,y)\tilde u(x,y))\right)\\
    &\stackrel{(b)}{=}\tilde u(x,y)-V_2V_2^\top\left(\nabla_y f(x,y)+\nabla_{yy}g(x,y)\tilde u(x,y)\right)\numberthis\label{!!!}
\end{align*}
where (a) comes from Lemma \ref{lm:proj_closed} and $(V_1^\top)^\dagger=V_1$, (b) is derived from $\tilde u(x,y)\in\{u|V_1^\top u=0\}$ and $I-V_1V_1^\top=V_2V_2^\top$. According to \eqref{!!!}, we know that 
\begin{align}\label{!!-1}
    V_2V_2^\top\left(\nabla_y f(x,y)+\nabla_{yy}g(x,y)\tilde u(x,y)\right)=0.
\end{align}
By the definition of $u^*(x,y)$ in \eqref{lbda*}, we can verify it satisfies \eqref{!!-1} since
\begin{align*}
    &~~~~~V_2V_2^\top(\nabla_yf(x,y)-\nabla_{yy}g(x,y)V_2(V_2^\top\nabla_{yy}g(x,y)V_2)^{-1}V_2^\top\nabla_y f(x,y))\\
    &=V_2\left(I-V_2^\top\nabla_{yy}g(x,y)V_2(V_2^\top\nabla_{yy}g(x,y)V_2)^{-1}\right)V_2^\top\nabla_y f(x,y))\\
    &=V_2\left(I-I\right)V_2^\top\nabla_y f(x,y))=0.
\end{align*}
Then due to uniqueness of $\tilde u(x,y)$, we know $u^*(x,y)=\tilde u(x,y)$. 
\end{proof}

Next, we will prove the boundedness of $u^*(x,y)$ and $e^*(x,y)$ where the latter is  defined as
\begin{align}
    e^*(x,y):=\nabla_y f(x,y)+\nabla_{yy}g(x,y)u^*(x,y).\label{estar}
\end{align}

\begin{lemma}[Boundedness of $u^*(x,y)$ and $e^*(x,y)$] 
Under Assumption \ref{as1}--\ref{as2}, for any $x$ and $y$, $u^*(x,y)$ is bounded by $\frac{\ell_{f,0}}{\mu_g}$, and $e^*(x,y)$ is bounded by $\ell_{f,0}\left(1+\ell_{g,1}/\mu_g\right)$. 
\label{ue-bound}
\end{lemma}
\begin{proof}
According to the definition of $u^*(x,y)$, 
we have $\|u^*(x,y)\|$ is bounded by $\frac{\ell_{f,0}}{\mu_g}$ since 
\begin{align*}
    \|u^*(x,y)\|\leq \|V_2(V_2^\top\nabla_{yy}g(x,y))^{-1}V_2^\top\|\|\nabla_y f(x,y)\|\leq \frac{\ell_{f,0}}{\mu_g}. 
\end{align*}
On the other hand, since $e^*(x,y)=\nabla_y f(x,y)+\nabla_{yy}g(x,y)u^*(x,y)$, we have $e^*(x,y)$ is bounded 
\begin{align*}
    \|e^*(x,y)\|\leq\|\nabla_y f(x,y)\|+\|\nabla_{yy}g(x,y)\|\|u^*(x,y)\|\leq\ell_{f,0}\left(1+\ell_{g,1}/\mu_g\right)
\end{align*}
which completes the proof. 
\end{proof}
\vspace{-0.2cm}
\subsection{Error of medium level}
Unfortunately, we cannot obtain the descent lemma for $u^k$ using the similar procedure of Lemma \ref{lm-low-stoc-skip} or applying the results in \citep{shen2022single} since $u^*(x,y)$ is not smooth with respect to $x$ and $y$. Instead, we can utilize the boundedness of $u^*(x,y)$ and $e^*(x,y)$. 


\begin{lemma}
Suppose that Assumption \ref{as1}--\ref{ash} hold,  $\rho\leq\min\left\{\frac{1}{\ell_{g,1}},\frac{\mu_g}{4\sigma_{g,2}^2}\right\}$ and $q=\sqrt{\rho\mu_g/2}$, then we have the following inequalities
\begin{align*}
    &\EE\left[\|u^{k+1}-u^*(x^k,y^{k+1})\|^2\Big|\overline{\mathcal{F}}_{k}^0\right]\leq (1-\rho\mu_g/2)^NC_u^2+\frac{2\rho\sigma_u^2}{\mu_g}\numberthis\label{uk-a}
\end{align*}
where $C_u^2:=\ell_{f,0}^2\left(1/\mu_g^2+(1+\ell_{g,1}/\mu_g)^2\right)$ and $\sigma_u^2:=\sigma_f^2+\frac{\sigma_{g,2}^2\ell_{f,0}^2}{\mu_g^2}$. 
\label{descent-uk}
\end{lemma}

\begin{proof}
First, we notice that $u^*(x,y)$ is the minimizer of a $\mu_g$- strongly convex function over a linear space according to Lemma \ref{lm:u*-strong}, which enables us to apply the convergence results of Proxskip at our medium level. Besides, the variance of gradient estimator at $u^{k,n}$ can be bounded by
\begin{align*}
    &~~~~\EE\left[\|\nabla_y f(x^k,y^{k+1};\xi^{k}_{(n)})+\nabla_{yy}g(x^k,y^{k+1};\phi^k_{(n)})u^{k,n}-\nabla_y f(x^k,y^{k+1})-\nabla_{yy}g(x^k,y^{k+1})u^{k,n}\|^2|\overline{\mathcal{F}}_{k}^n\right]\\
    &\stackrel{(a)}{\leq}\EE\left[\|\nabla_y f(x^k,y^{k+1};\phi^k_{(n)})-\nabla_y f(x^k,y^{k+1})\|^2|\overline{\mathcal{F}}_{k}^n\right]\\
    &~~~~~+\EE\left[\|\left(\nabla_{yy}g(x^k,y^{k+1};\phi^k_{(n)})-\nabla_{yy}g(x^k,y^{k+1})\right)u^{k,n}\|^2|\overline{\mathcal{F}}_{k}^n\right]\\
    &\stackrel{(b)}{\leq}\sigma_{f}^2+2\sigma_{g,2}^2\EE[\|u^{k,n}-u^*(x^k,y^{k+1})\|^2|\overline{\mathcal{F}}_{k}^n]+2\sigma_{g,2}^2\EE[\|u^*(x^k,y^{k+1})\|^2|\overline{\mathcal{F}}_{k}^n]\\
    &\stackrel{(c)}{\leq}\sigma_{f}^2+2\sigma_{g,2}^2\EE[\|u^{k,n}-u^*(x^k,y^{k+1})\|^2|\overline{\mathcal{F}}_{k}^n]+\frac{2\sigma_{g,2}^2\ell_{f,0}^2}{\mu_g^2}
\end{align*}
where (a) is derived from Lemma \ref{lm:matrix}, (b) comes from Assumption \ref{as-stoc} and (c) is due to the boundedness of $u^*(x,y)$ in Lemma \ref{ue-bound}. 

Then, defining $\sigma_u^2:=\sigma_{f}^2+\frac{2\sigma_{g,2}^2\ell_{f,0}^2}{\mu_g^2}$ and leveraging Lemma C.1 and Lemma C.2 in Proxskip \citep{mishchenko2022proxskip} with the above variance, we can prove the one-step contraction by 
\begin{align*}
    &\quad~\EE\left[\|u^{k,n+1}-u^*(x^k,y^{k+1})\|^2+\frac{\rho^2}{q^2}\|e^{k,n+1}-e^*(x^k,y^{k+1})\|^2|\overline{\mathcal{F}}_{k}^n\right]\nonumber\\
    &\leq\EE\left[(1-\rho\mu_g+2\rho^2\sigma_{g,2}^2)\| u^{k,n}-u^*(x^k,y^{k+1})\|^2+(1-q^2)\frac{\rho^2}{q^2}\|e^{k,n}-e^*(x^k,y^{k+1})\|^2|\overline{\mathcal{F}}_{k}^n\right]+\rho^2\sigma_u^2\\
    &\stackrel{(a)}{\leq}\EE\left[(1-\rho\mu_g/2)\| u^{k,n}-u^*(x^k,y^{k+1})\|^2+(1-q^2)\frac{\rho^2}{q^2}\|e^{k,n}-e^*(x^{k},y^{k+1})\|^2|\overline{\mathcal{F}}_{k}^n\right]+\rho^2\sigma_u^2\\
    &\stackrel{(b)}{=}(1-\rho\mu_g/2)\EE\left[\| u^{k,n}-u^*(x^k,y^{k+1})\|^2+\frac{\rho^2}{q^2}\|e^{k,n}-e^*(x^{k},y^{k+1})\|^2|\overline{\mathcal{F}}_{k}^n\right]+\rho^2\sigma_u^2\numberthis\label{contract-medium}
\end{align*}
where (a) comes from $\rho\leq\frac{\mu_g}{4\sigma_{g,2}^2}$ and (b) is derived from the choice of $q$. 

Then telescoping \eqref{contract-medium} and since $u^{k,0}=e^{k,0}=0$, we have
\begin{align*}
    &\quad~\EE\left[\|u^{k+1}-u^*(x^k,y^{k+1})\|^2|\overline{\mathcal{F}}_{k}^0\right]\\
    &\leq\EE\left[\|u^{k+1}-u^*(x^k,y^{k+1})\|^2+\frac{\rho^2}{q^2}\|e^{k+1}-e^*(x^k,y^{k+1})\|^2|\overline{\mathcal{F}}_{k}^0\right]\nonumber\\
    &\leq(1-\rho\mu_g/2)^N\EE\left[\| u^*(x^k,y^{k+1})\|^2+\frac{\rho^2}{q^2}\|e^*(x^{k},y^{k+1})\|^2|\overline{\mathcal{F}}_{k}^0\right]+\rho^2\sigma_u^2\sum_{n=0}^{N}(1-\rho\mu_g/2)^n\\
    &\stackrel{(a)}{\leq}(1-\rho\mu_g/2)^N\ell_{f,0}^2\left(1/\mu_g^2+(1+\ell_{g,1}/\mu_g)^2\right)+\frac{\rho^2\sigma_u^2}{\rho\mu_g/2}
\end{align*}
where (a) is derived from the boundedness of $u^*(x,y)$ and $e^*(x,y)$, and $\rho\leq q$. Then by the definition of $C_u^2$, we obtain \eqref{uk-a}. 
\end{proof}

\subsection{Descent of upper level}
\begin{lemma}
Suppose that Assumption \ref{as1}--\ref{ash} hold and choose $N={\cal O}(1/\alpha), \rho\leq\min\left\{\frac{1}{\ell_{g,1}},\frac{\mu_g}{4\sigma_{g,2}^2}\right\}$ and $q=\sqrt{\rho\mu_g/2}$, then we can obtain the following bounds
\begin{subequations}
\begin{align}
    &\EE[\|\bar d_f^k\|^2|\overline{\mathcal{F}}_k^0]\leq 2\ell_{f,0}^2\left(1+\frac{2\ell_{g,1}^2}{\mu_g^2}\right)+4\ell_{g,1}^2\left((1-\rho\mu_g/2)^NC_u^2+\frac{2\rho\sigma_u^2}{\mu_g}\right)=:C_{f,2}^2\label{108-a}\\
    &\EE[\|d_f^k-\bar d_f^k\|^2|\overline{\mathcal{F}}_k^0]\leq 4\sigma_u^2+4\sigma_{g,2}^2\left((1-\rho\mu_g/2)^NC_u^2+\frac{2\rho\sigma_u^2}{\mu_g}\right)=:\tilde\sigma_{f,2}^2\label{108-b}
\end{align}
\end{subequations}
where $\sigma_u^2:=\sigma_{f}^2+\frac{\sigma_{g,2}^2\ell_{f,0}^2}{\mu_g^2}$. 
\label{incre-bound}
\end{lemma}
\begin{proof}\allowdisplaybreaks
For $x$- sequence, we have
\begin{align*}
    \|\bar d_f^k\|^2&=\|\nabla_x f(x^k,y^{k+1})+\nabla_{xy}g(x^k,y^{k+1})u^{k+1}\|^2\\
    &\leq2\ell_{f,0}^2+2\ell_{g,1}^2\|u^{k+1}\|^2\\
    &\leq 2\ell_{f,0}^2+4\ell_{g,1}^2\left(\|u^{k+1}-u^*(x^k,y^{k+1})\|^2+\|u^*(x^k,y^{k+1})\|^2\right)\\
    &\leq 2\ell_{f,0}^2+4\ell_{g,1}^2\left(\|u^{k+1}-u^*(x^k,y^{k+1})\|^2+\frac{\ell_{f,0}^2}{\mu_g^2}\right)\\
    &=2\ell_{f,0}^2\left(1+\frac{2\ell_{g,1}^2}{\mu_g^2}\right)+4\ell_{g,1}^2\|u^{k+1}-u^*(x^k,y^{k+1})\|^2
\end{align*}
where the first inequality comes from Assumption \ref{as1} and the third inequality is derived from the bound for $u^*(x,y)$ in Lemma \ref{ue-bound}. Then taking expectation and using Lemma \ref{descent-uk}, we get \eqref{108-a}.

Also, we can bound the variance of $x$-update as
\begin{align*}
    &~~~~~\EE[\|d_f^k-\bar d_f^k\|^2|\overline{\mathcal{F}}_k^0]\\
    &=\EE\left[\|\nabla_x f(x^k,y^{k+1};\xi^k)+\nabla_{xy}g(x^k,y^{k+1};\xi^k)u^{k+1}-\nabla_x f(x^k,y^{k+1})-\nabla_{xy}g(x^k,y^{k+1})u^{k+1}\|^2|\overline{\mathcal{F}}_k^0\right]\\
    &\leq2\EE\left[\|\nabla_x f(x^k,y^{k+1};\xi^k)-\nabla_x f(x^k,y^{k+1})\|^2|\overline{\mathcal{F}}_k^0\right]\\
    &~~~~~+2\EE\left[\|\left(\nabla_{xy}g(x^k,y^{k+1};\xi^k)-\nabla_{xy}g(x^k,y^{k+1})\right)u^{k+1}\|^2|\overline{\mathcal{F}}_k^0\right]\\
    &\leq2\sigma_f^2+2\sigma_{g,2}^2\EE\left[\|u^{k+1}\|^2|\overline{\mathcal{F}}_k^0\right]\\
    &\leq2\sigma_f^2+4\sigma_{g,2}^2\EE\left[\|u^{k+1}-u^*(x^k,y^{k+1})\|^2|\overline{\mathcal{F}}_k^0\right]+4\sigma_{g,2}^2\EE\left[\|u^*(x^k,y^{k+1})\|^2|\overline{\mathcal{F}}_k^0\right]\\
    &\leq2\sigma_f^2+4\sigma_{g,2}^2\EE\left[\|u^{k+1}-u^*(x^k,y^{k+1})\|^2|\overline{\mathcal{F}}_k^0\right]+\frac{4\sigma_{g,2}^2\ell_{f,0}^2}{\mu_g^2}\\
    &\leq4\sigma_u^2+4\sigma_{g,2}^2\EE\left[\|u^{k+1}-u^*(x^k,y^{k+1})\|^2|\overline{\mathcal{F}}_k^0\right]\\
    &\leq 4\sigma_u^2+4\sigma_{g,2}^2\left((1-\rho\mu_g/2)^NC_u^2+\frac{2\rho\sigma_u^2}{\mu_g}\right)
\end{align*}
where the second inequality is from Assumption \ref{as-stoc} and the last inequality is from Lemma \ref{descent-uk}. 
\end{proof}

The update of UL in {\sf E2-AiPOD} can still be viewed as biased SGD. For simplicity, we define a virtual sequence $\bar x^k:=\operatorname{Proj}_{\mathcal{X}}(x^k)$. Then we have 
\begin{align*}
    \bar x^{k+1}
    &\stackrel{(a)}{=}\operatorname{Proj}_{\mathcal{X}}(x^{k}-\alpha d_f^k)\\
    &\stackrel{(b)}{=}(I-B^\dagger B)(x^{k}-\alpha d_f^{k})+B^\dagger Be\\
    &=\operatorname{Proj}_{\mathcal{X}}(x^{k})-\alpha(I-B^\dagger B)d_f^{k}\\
    &=\bar x^{k}-\alpha(I-B^\dagger B)d_f^{k}\numberthis\label{update-2}
\end{align*}
where $(a)$ results from the update of $x$ and $(b)$ is derived from the definition of $\operatorname{Proj}(\cdot)$ on the linear space $\mathcal{X}=\{x\mid Bx=e\}$. We can prove the following lemma. 

\begin{lemma}\label{lm:diff-x}
Under Assumption \ref{as1}--\ref{ash}, if we choose $N={\cal O}(1/\alpha), \rho\leq\min\left\{\frac{1}{\ell_{g,1}},\frac{\mu_g}{4\sigma_{g,2}^2}\right\}$ and $q=\sqrt{\rho\mu_g/2}$, then the sequence of $x^k$ generated by Algorithm \ref{stoc-alg-skip-WS} satisfies
\begin{align}
    \EE\left[\|x^k-\bar x^k\|^2\right]= {\cal O}(\alpha^2 T^2).
\end{align}
\end{lemma}
\begin{proof}
We prove this by induction. For any $k$ s.t. $k\operatorname{mod} T=0$, it holds that $x^k=\bar x^k$. Besides, for $(k+1)\operatorname{mod} T=0$, we have
\begin{align*}
    \bar x^{k+1}-x^{k+1}&=\bar x^k-\alpha(I-B^\dagger B)d_f^k-(x^k-\alpha d_f^k)\\
    &=\bar x^k-x^k+\alpha B^\dagger Bd_f^k. \numberthis\label{!!}
\end{align*}
For any $k$, let $i(k)\leq k$ be the largest value such that $i(k)\operatorname{mod} T=0$, then telescoping \eqref{!!} and using $\bar x^{i(k)}-x^{i(k)}=0$, we know
$
    \bar x^k-x^k=\alpha B^\dagger B \sum_{j=i(k)}^k d_f^j
$. 
Therefore, we obtain 
\begin{align*}
    \EE[\|\bar x^k-x^k\|^2]&=\EE\left[\left\|\alpha B^\dagger B \sum_{j=i(k)}^k d_f^j\right\|^2\right]\stackrel{(a)}{\leq}\alpha^2\EE[\|\sum_{j=i(k)}^k d_f^j\|^2]\stackrel{(b)}{\leq}\alpha^2 T\sum_{j=i(k)}^k\EE[\|d_f^j\|^2]\\
    &\leq \alpha^2 T\sum_{j=i(k)}^k\left\{\EE[\|\bar d_f^j\|^2]+\EE[\|d_f^j-\bar d_f^j\|^2]\right\}\\
    &\stackrel{(c)}{\leq} \alpha^2 T^2\left(C_{f,2}^2+4(\ell_{g,1}^2+\sigma_{g,2}^2)\EE[\|u^{k+1}-u^*(x^k,y^{k+1})\|^2]+4\sigma_u^2\right)\numberthis\label{medium--1}
\end{align*}
where (a) holds since $\|B^\dagger B\|\leq 1$, (b) holds since $\EE[\|\sum_{i=1}^I X_i\|^2]\leq I\sum_{i=1}^I\EE[\| X_i\|^2]$ and $k-i(k)\leq T$, and (c) is derived from Lemma \ref{incre-bound}. Then by plugging the bound in Lemma \ref{descent-uk} into \eqref{medium--1} and choose $N={\cal O}(1/\alpha)$, we obtain
\begin{align*}
    \EE\left[\|x^k-\bar x^k\|^2\right]\leq\alpha^2 T^2\left(C_{f,2}^2+4(\ell_{g,1}^2+\sigma_{g,2}^2)\left((1-\rho\mu_g/2)^NC_u^2+\frac{2\rho\sigma_u^2}{\mu_g}\right)+4\sigma_u^2\right)={\cal O}(\alpha^2 T^2)
\end{align*}
which completes the proof. 
\end{proof}

\begin{lemma}[Descent of UL]
Under Assumption \ref{as1}--\ref{ash} and setting $N={\cal O}(1/\alpha), \rho\leq\min\left\{\frac{1}{\ell_{g,1}},\frac{\mu_g}{4\sigma_{g,2}^2}\right\}$ and $q=\sqrt{\rho\mu_g/2}$, the virtual sequence $\bar x^k$ generated by Algorithm \ref{stoc-alg-skip-WS} satisfies
\begin{align*}
   \!\! \EE[F(\bar x^{k+1})]-\EE[F(\bar x^k)]&\leq -\frac{\alpha}{2}\EE[\|\nabla F(\bar x^k)\|_{P_x}^2]-\left(\frac{\alpha}{2}-\frac{L_F\alpha^2}{2}\right)\EE[\|\bar d_f^k\|_{P_x}^2]+\frac{L_F\alpha^2\tilde\sigma_{f,2}^2}{2}+{\cal O}(\alpha^3T^2)\\
    &~~~~+\frac{3\ell_{g,1}^2\alpha}{2}\!\Big((1-\rho\mu_g/2)^NC_u^2+\frac{2\rho\sigma_u^2}{\mu_g}\Big)\!+\!\frac{3\alpha}{2}\EE[\|y^{k+1}-y^*(x^k)\|^2]\!\numberthis\label{stoc:up-2}
\end{align*}
where $P_x=I-B^\dagger B$ is the projection matrix of $B$ and $B^\dagger$ is the Moore-Penrose inverse of $B$. \label{lm:UL-E2aipod}
\end{lemma}
\begin{proof}\allowdisplaybreaks
Taking expectation given $\overline{\mathcal{F}}_k^{N}$ and using smoothness, we obtain
\begin{align*}
    \EE[F(\bar x^{k+1})|\overline{\mathcal{F}}_k^{N}]&\leq F(\bar x^k)+\EE\langle\nabla F(\bar x^k),\bar x^{k+1}-\bar x^k\rangle|\overline{\mathcal{F}}_k^{N}]+\frac{L_F}{2}\EE[\|\bar x^{k+1}-\bar x^k\|^2|\overline{\mathcal{F}}_k^{N}]\\
    &\stackrel{(a)}{\leq} F(\bar x^k)-\langle\nabla F(\bar x^k),\alpha(I-B^\dagger B)\bar d_f^k\rangle+\frac{L_F\alpha^2}{2}\EE[\|d_f^k\|_{P_x}^2|\overline{\mathcal{F}}_k^{N}]\\
    &\stackrel{(b)}{=} F(\bar x^k)-\alpha\langle\nabla F(\bar x^k),(I-B^\dagger B)\bar d_f^k\rangle+\frac{L_F\alpha^2}{2}\left\{\EE[\|d_f^k-\bar d_f^k\|_{P_x}^2|\overline{\mathcal{F}}_k^{N}]+\|\bar d_f^k\|_{P_x}^2\right\}\\
    &\stackrel{(c)}{\leq}  F(\bar x^k)-\frac{\alpha}{2}\|\nabla F(\bar x^k)\|_{P_x}^2-\left(\frac{\alpha}{2}-\frac{L_F\alpha^2}{2}\right)\|\bar d_f^k\|_{P_x}^2\\
    &~~~~~+\frac{\alpha}{2}\|\nabla F(\bar x^k)-\bar d_f^k\|^2+\frac{L_F\alpha^2\tilde\sigma_{f,2}^2}{2}\numberthis\label{smooth-medium}
\end{align*}
where (a) results from \eqref{update-2}, (b) is derived from  $\EE[\|X\|^2|Y]=\|\EE[X|Y]\|^2+\EE[\|X-\EE[X|Y]\|^2|Y]$, and (c) holds since $2a^\top b= \|a\|^2+\|b\|^2-\|a-b\|^2$, $\|I-B^\dagger B\|\leq 1$ and Lemma \ref{incre-bound}. 

Besides, we can decompose the bias of UL gradient estimator as
\begin{align*}
    \|\nabla F(\bar x^k)-\bar d_f^k\|^2&=\|\nabla F(\bar x^k)-\overline\nabla f(\bar x^k,y^{k+1})+\overline\nabla f(\bar x^k,y^{k+1})-\overline\nabla f(x^k,y^{k+1})+\overline\nabla f(x^k,y^{k+1})-\bar d_f^k\|^2\\
    &\stackrel{(a)}{\leq} 3 \|\nabla F(\bar x^k)-\overline\nabla f(\bar x^k,y^{k+1})\|^2+3\|\overline\nabla f(\bar x^k,y^{k+1})-\overline\nabla f(x^k,y^{k+1})\|^2\\
    &~~~~~+3\|\nabla_{xy} g(x^k,y^{k+1})(u^*(x^k,y^{k+1})-u^{k+1})\|^2\\
    &\stackrel{(b)}{\leq} 3 L_f^2\|y^{k+1}-y^*(\bar x^k)\|^2+3L_f^2\|\bar x^k-x^k\|^2+3\ell_{g,1}^2\|u^{k+1}-u^*(x^k,y^{k+1})\|^2\numberthis\label{!!medium}
\end{align*}
where (a) comes from $\|X+Y+Z\|^2\leq 3\|X\|^2+3\|Y\|^2+3\|Z\|^2$, $\nabla F(x)=\overline\nabla f(x,y^*(x))$ and $\overline\nabla f(x,y)=\nabla_x f(x,y)+\nabla_{xy}g(x,y)u^*(x,y)$; and (b) holds since $F(x),\overline\nabla f(x,y)$ is Lipschitz continuous and $\nabla_{xy}g(x,y)$ is bounded. 

Plugging \eqref{!!medium} into \eqref{smooth-medium} and taking expectation, we get
\begin{align*}
    \EE[F(\bar x^{k+1})]&\leq\EE[F(\bar x^k)]-\frac{\alpha}{2}\EE[\|\nabla F(\bar x^k)\|_{P_x}^2]-\left(\frac{\alpha}{2}-\frac{L_F\alpha^2}{2}\right)\EE[\|\bar d_f^k\|_{P_x}^2]+\frac{L_F\alpha^2\tilde\sigma_{f,2}^2}{2}\\
    &~~~~~~+\frac{3\ell_{g,1}^2\alpha}{2}\EE\left[\|u^{k+1}-u^*(x^k,y^{k+1})\|^2\right]\\
    &~~~~~~+\frac{3\alpha L_f^2}{2}\EE[\|y^{k+1}-y^*(\bar x^k)\|^2]+\frac{3\alpha L_f^2}{2}\EE[\|x^k-\bar x^k\|^2]\\
    &\stackrel{(a)}{\leq}\EE[F(\bar x^k)]-\frac{\alpha}{2}\EE[\|\nabla F(\bar x^k)\|_{P_x}^2]-\left(\frac{\alpha}{2}-\frac{L_F\alpha^2}{2}\right)\EE[\|\bar d_f^k\|_{P_x}^2]+\frac{L_F\alpha^2\tilde\sigma_{f,2}^2}{2}\\
    &~~~~~~+\frac{3\ell_{g,1}^2\alpha}{2}\left((1-\rho\mu_g/2)^NC_u^2+\frac{2\rho\sigma_u^2}{\mu_g}\right) +\frac{3\alpha L_f^2}{2}\EE[\|y^{k+1}-y^*(\bar x^k)\|^2]+{\cal O}(\alpha^3T^2)
\end{align*}
where (a) results from Lemma \ref{descent-uk} and Lemma \ref{lm:diff-x}. This completes the proof. 
\end{proof}

\subsection{Error of lower level}
In this section, we focus on the LL update of {\sf E2-AiPOD} and will prove the counterparts of Lemmas \ref{lm-low-stoc} and \ref{lm-low-stoc-skip}. The alternative of one-step contraction in \eqref{46a-new} is easy to obtain by the nature of Proxskip. However, establishing the alternative of \eqref{46b-new} requires the bound for $\EE[\|x^{k+1}-x^k\|^2]$. Unlike the update rule for {\sf AiPOD} \eqref{linear-update} and {\sf E-AiPOD} \eqref{update-up-T}, the UL update of {\sf E2-AiPOD} can not be expressed as
\begin{align*}
    x^{k+1}=x^k-\alpha (I-B^\dagger B)d_f^k
\end{align*}
unless $k\operatorname{mod}T=0$. However, only this type of error $\|\alpha(I-B^\dagger B)d_f^k\|^2=\alpha^2\|d_f^k\|^2_{P_x}$ can be mitigated by the descent of UL. We then observe that the update of virtual sequence in \eqref{update-2} is preferable since $\EE[\|\bar x^{k+1}-\bar x^k\|^2]=\alpha^2\EE[\| d_f^k\|_{P_x}^2]$. Thus, a corresponding modification is made on the one-step contraction part that we change the reference point from $y^*(x^k)$ to $y^*(\bar x^k)$. Fortunately, the error induced by treating $y^{k+1}$ as the output of biased Proxskip with respect to $y^*(\bar x^k)$ is negligible in the final convergence (i.e. ${\cal O}(\beta \alpha^2T^2S)$ in \eqref{46a-new2}). Formally, we have the following lemma. 
\begin{lemma}[\bf{Error of LL update}]\label{lm-low-stoc-skip2}
Suppose that Assumption \ref{as1}--\ref{ash} hold and $\beta\leq\frac{1}{\ell_{g,1}}$, if we choose $N={\cal O}(1/\alpha), \rho\leq\min\left\{\frac{1}{\ell_{g,1}},\frac{\mu_g}{4\sigma_{g,2}^2}\right\}$ and $q=\sqrt{\rho\mu_g/2}$,   the error of LL update can be bounded by
\begin{subequations}
\small
\begin{align}
    &\EE\left[\|y^{k+1}-y^*(\bar x^k)\|^2+\frac{\beta^2}{p^2}\|r^{k+1}-r^*(\bar x^k)\|^2\right]\nonumber\\
    &~~~~~~~~~~~~~~~~~~~~\leq\left(1-\nu\right)^S\EE\left[\|y^{k}-y^*(\bar x^k)\|^2+\frac{\beta^2}{p^2}\|r^{k}-r^*(\bar x^k)\|^2\right]+S\beta^2\sigma_{g,1}^2+{\cal O}(\beta \alpha^2T^2S)\label{46a-new2}\\
    &\EE[\|y^{k+1}-y^*(\bar x^{k+1})\|^2]\leq \left(1+\gamma +  L_{yx}(C_{f,2}^2+\tilde\sigma_{f,2}^2)\alpha^2\right)\EE[\|y^{k+1}-y^*(\bar x^{k})\|^2]\nonumber\\
    &\qquad\qquad\qquad\qquad\qquad\quad+\left(L_y^2+L_{yx}\right)\alpha^2\tilde\sigma_{f,2}^2+\left(L_y^2+L_{yx}+\frac{L_y^2}{\gamma}\right)\alpha^2\EE\left[\|{\bar d}_f^k\|_{P_x}^2\right]\label{46b-new2}\\
    &\EE[\|r^{k+1}-r^*(\bar x^{k+1})\|^2]\leq \left(1+\gamma +  L_{rx}(C_{f,2}^2+\tilde\sigma_{f,2}^2)\alpha^2\right)\EE[\|r^{k+1}-r^*(\bar x^{k})\|^2]\nonumber\\
    &\qquad\qquad\qquad\qquad\qquad\quad+\left(L_r^2+L_{rx}\right)\alpha^2 \tilde\sigma_{f,2}^2+\left(L_r^2+L_{rx}+\frac{L_r^2}{\gamma}\right)\alpha^2\EE\left[\|{\bar d}_f^k\|_{P_x}^2\right]\label{46c-new2}
\end{align}
\end{subequations}
where $C_{f,2}^2,\tilde\sigma_{f,2}^2$ are constants defined in Lemma \ref{incre-bound}, $\nu:=\min\left\{\beta\mu_g/2,p^2\right\}$ , $\gamma$ is the balancing parameter that will be chosen in the final theorem. 
\end{lemma}

\begin{proof}
The proof of \eqref{46a-new2} can be viewed as a biased version of \eqref{46a-new} if we treat $\nabla_y g(x^k,y^{k,s};\phi^{k,s})$ as a biased estimator of $\nabla_y g(\bar x^k,y^{k,s})$. Note that Lemma C.1 in Proxskip holds even if the gradient estimator is biased since it only considers the randomness of $\theta^{k,s}$ and uses the properties of projection. Therefore, leveraging Lemma C.1 from Proxskip \citep{mishchenko2022proxskip}, we get
{\small\begin{align*}
    &~~~~\EE_{\theta^{k,s}}\left[\|y^{k,s+1}-y^*(\bar x^k)\|^2+\frac{\beta^2}{p^2}\|r^{k,s+1}-r^*(\bar x^k)\|^2|\mathcal{F}_{k}^s\right]\\
    &\leq\|y^{k,s}-\beta\nabla_y g(x^k,y^{k,s};\phi^k)-(y^*(\bar x^k)-\beta\nabla_y g(\bar x^k,y^*(\bar x^k)))\|^2+\frac{(1-p^2)\beta^2}{p^2}\|r^{k,s}-r^*(\bar x^k)\|^2. \numberthis\label{proxskip-var}
\end{align*}}
Then taking a conditional expectation over \eqref{proxskip-var} yields
\begin{align*}
    &~~~~~\EE\left[\|y^{k,s}-\beta\nabla_y g(x^k,y^{k,s};\phi^k)-(y^*(\bar x^k)-\beta\nabla_y g(\bar x^k,y^*(\bar x^k)))\|^2\Big|\mathcal{F}_{k}^s\right]\\
    &\stackrel{(a)}{\leq}\|y^{k,s}-\beta\nabla_y g(x^k,y^{k,s})-(y^*(\bar x^k)-\beta\nabla_y g(\bar x^k,y^*(\bar x^k)))\|^2\\
    &~~~~~+\beta^2\EE\left[\|\nabla_y g(x^k,y^{k,s};\phi^{k,s})-\nabla_y g( x^k,y^{k,s})\|^2\Big|\mathcal{F}_{k}^s\right]\\
    &\leq \|y^{k,s}-\beta\nabla_y g(x^k,y^{k,s})-(y^*(\bar x^k)-\beta\nabla_y g(\bar x^k,y^*(\bar x^k)))\|^2+\beta^2\sigma_{g,1}^2\\
    &\stackrel{(b)}{\leq} (1+\frac{\beta\mu_g}{2})\|y^{k,s}-\beta\nabla_y g(\bar x^k,y^{k,s})-(y^*(\bar x^k)-\beta\nabla_y g(\bar x^k,y^*(\bar x^k)))\|^2\\
    &~~~~+(1+\frac{1}{2\beta\mu_g})\beta^2\|\nabla_y g(\bar x^k,y^{k,s})-\nabla_y g(x^k,y^{k,s})\|^2+\beta^2\sigma_{g,1}^2\\
    &\stackrel{(c)}{\leq} (1+\beta\mu_g/2)(1-\beta\mu_g)\|y^{k,s}-y^*(\bar x^k)\|^2+(1+\frac{1}{2\beta\mu_g})\beta^2\ell_{g,1}^2\|x^k-\bar x^k\|^2+\beta^2\sigma_{g,1}^2\numberthis\label{medium-ex}
\end{align*}
where (a) comes from Lemma \ref{lm:matrix}, (b) is obtained by Young's inequality, and (c) is derived similarly to \eqref{contra-stoc} and the Lipschitz continuity of $\nabla_y g(x,y)$.  

Furthermore, taking expectation over \eqref{medium-ex} yields
\begin{align*}
    &~~~~~\EE\left[\|y^{k,s}-\beta\nabla_y g(x^k,y^{k,s};\phi^k)-(y^*(\bar x^k)-\beta\nabla_y g(\bar x^k,y^*(\bar x^k)))\|^2\right]\\
    &\leq(1+\beta\mu_g/2)(1-\beta\mu_g)\EE[\|y^{k,s}-y^*(\bar x^k)\|^2]+(1+\frac{1}{2\beta\mu_g})\beta^2\ell_{g,1}^2\EE[\|x^k-\bar x^k\|^2]+\beta^2\sigma_{g,1}^2\\
    &\leq (1-\beta\mu_g/2)\EE[\|y^{k,s}-y^*(\bar x^k)\|^2]+{\cal O}(\beta \alpha^2T^2)+\beta^2\sigma_{g,1}^2\numberthis\label{medium-tele}
\end{align*}
where the last inequality is derived from Lemma \ref{lm:diff-x}. 

Therefore, combining with \eqref{proxskip-var} and telescoping \eqref{medium-tele}, we get
\begin{align*}
    &\EE\left[\|y^{k+1}-y^*(\bar x^k)\|^2+\frac{\beta^2}{p^2}\|r^{k+1}-r^*(\bar x^k)\|^2\right]\nonumber\\
    &~~~~~~~~~~~~~~~\leq\left(1-\nu\right)^S\EE\left[\|y^{k}-y^*(\bar x^k)\|^2+\frac{\beta^2}{p^2}\|r^{k}-r^*(\bar x^k)\|^2\right]+S\beta^2\sigma_{g,1}^2+{\cal O}(\beta \alpha^2T^2S)\numberthis
\end{align*}
where $\nu:=\min\{\beta\mu_g/2,p^2\}$. 

The proof of \eqref{46b-new2} is identical to the proof of \eqref{40b} by replacing $\bar h_f^k,\tilde\sigma_{f}^2,\tilde C_f^2$ by $\bar d_f^k,\tilde\sigma_{f,2}^2,(C_{f,2}^2+\tilde\sigma_{f,2}^2)$ owing to the update rule in \eqref{update-2}. While the proof of \eqref{46c-new2} can be obtained with the same spirits of \eqref{46b-new} by replacing $L_y, L_{yx}$ in \eqref{46b-new2} with $L_r$ and $L_{rx}$. 
\end{proof}

\subsection{Proof of Theorem \ref{thm:conv-E2}}\label{sec:thm:conv-E2}
 
First, without loss of generality, we can assume that $\ell_{g,1}\geq 1$ so that $L_r\geq L_y, L_{rx}\geq L_{yx}$ and plugging \eqref{46b-new2}, \eqref{46c-new2} into \eqref{46a-new2} in Lemma \ref{lm-low-stoc-skip2}, we get that
\begin{align*}
    &~~~~\EE\left[\|y^{k+1}-y^*(\bar x^{k+1})+\frac{\beta^2}{p^2}\|r^{k+1}-r^*(\bar x^{k+1})\|^2\|^2\right]\\
    &\leq \left(1+\gamma +  L_{rx}(C_{f,2}^2+\tilde\sigma_{f,2}^2)\alpha^2\right)\left(1-\nu\right)^S\EE\left[\|y^{k}-y^*(\bar x^{k})\|^2+\frac{\beta^2}{p^2}\|r^{k}-r^*(\bar x^{k})\|^2\right]\\
    &~~~~~+\left(1+\gamma +  L_{rx}(C_{f,2}^2+\tilde\sigma_{f,2}^2)\alpha^2\right)\left(S\beta^2\sigma_{g,1}^2+{\cal O}(\beta \alpha^2T^2S)\right)\\
    &~~~~~+\left(L_r^2+L_{rx}\right)\left(1+\frac{\beta^2}{p^2}\right)\alpha^2\tilde\sigma_{f,2}^2+\left(L_r^2+L_{rx}+\frac{L_r^2}{\gamma}\right)\left(1+\frac{\beta^2}{p^2}\right)\alpha^2\EE\left[\|\bar d_f^k\|_{P_x}^2\right].\numberthis\label{yr*}
\end{align*}
Then using Lyapunov function defined in \eqref{lyapunov-skip-WS} and applying \eqref{yr*}, and Lemma \ref{lm:UL-E2aipod}, we get
{\small\begin{align*}
   & \EE\left[\mathbb{V}_2^{k+1}\right]-\EE\left[\mathbb{V}_2^k\right]\\
    &\leq-\frac{\alpha}{2}\EE\left[\|\nabla F(\bar x^k)\|_{P_x}^2\right]+\frac{3\alpha L_f^2}{2} \EE\left[\|y^{k+1}-y^*(\bar x^k)\|^2\right]+{\cal O}(\alpha^3 T^2)\\
   &~~~~~ -\left(\frac{\alpha}{2}-\frac{L_F\alpha^2}{2}\right) \EE\left[\|{\bar d}_f^k\|_{P_x}^2\right]+\frac{L_F\alpha^2 \tilde\sigma_{f,2}^2}{2}+\frac{3\ell_{g,1}^2\alpha}{2}\left((1-\rho\mu_g/2)^NC_u^2+\frac{2\rho\sigma_u^2}{\mu_g}\right)\\
   &~~~~~ +\frac{L_f}{L_r}\left[\left(1+\gamma +  L_{rx}(C_{f,2}^2+\tilde\sigma_{f,2}^2)\alpha^2\right)\left(1-\nu\right)^S-1\right]\EE\left[\|y^k-y^*(\bar x^k)\|^2+\frac{\beta^2}{p^2}\|r^k-r^*(\bar x^k)\|^2\right]\\
   &~~~~~+\frac{L_f}{L_r}\left(1+\gamma +  L_{rx}(C_{f,2}^2+\tilde\sigma_{f,2}^2)\alpha^2\right)\left(S\beta^2\sigma_{g,1}^2+{\cal O}(\beta \alpha^2T^2S)\right)  +\frac{L_f}{L_r}\left(L_r^2+L_{rx}\right)\left(1+\frac{\beta^2}{p^2}\right)\alpha^2\tilde\sigma_{f,2}^2\\
   &~~~~~~+\frac{L_f}{L_r}\left(L_r^2+L_{rx}+\frac{L_r^2}{\gamma}\right)\left(1+\frac{\beta^2}{p^2}\right)\alpha^2\EE\left[\|{\bar d}_f^k\|_{P_x}^2\right]\\
   &\leq-\frac{\alpha}{2}\EE\left[\|\nabla F(\bar x^k)\|_{P_x}^2\right]+\frac{L_f}{L_r}\left(1+\gamma +\frac{3L_rL_f\alpha}{2}+  L_{rx}(C_{f,2}^2+\tilde\sigma_{f,2}^2)\alpha^2\right)\left(S\beta^2\sigma_{g,1}^2+{\cal O}(\beta \alpha^2T^2S)\right) \\
   &~~~~~+\left[\frac{L_F}{2}+\frac{L_f}{L_r}\left(L_r^2+L_{rx}\right)\left(1+\frac{\beta^2}{p^2}\right)\right]\alpha^2\tilde\sigma_{f,2}^2+{\cal O}(\alpha^3T^2)+\frac{3\ell_{g,1}^2\alpha}{2}\left((1-\rho\mu_g/2)^NC_u^2+\frac{2\rho\sigma_u^2}{\mu_g}\right)\\
   &~~~~~-\left[\frac{\alpha}{2}-\left(\frac{L_F}{2}+L_fL_r\left(1+\frac{1}{\gamma}\right)\left(1+\frac{\beta^2}{p^2}\right)+\frac{L_fL_{rx}}{  L_r}\left(1+\frac{\beta^2}{p^2}\right)\right)\alpha^2\right]\EE\left[\|\bar d_f^k\|_{P_x}^2\right]\\
   &~~~~~-\left(\frac{L_f\nu}{L_r}-\frac{3\alpha L_f^2}{2}-\frac{L_f\gamma }{L_r}-\frac{  L_fL_{rx}(C_f^2+\tilde\sigma_d^2)\alpha^2}{L_r}\right)\EE\left[\|y^k-y^*(\bar x^k)\|^2+\frac{\beta^2}{p^2}\|r^k-r^*(\bar x^k)\|^2\right]. \numberthis\label{lay-minus-stoc-r}
\end{align*}}
Selecting $\gamma=4L_fL_r\alpha$ and $p=\sqrt{\beta\mu_g/2}$, \eqref{lay-minus-stoc-r} can be simplified by
\begin{align*}
\!\!    &\EE[\mathbb{V}_2^{k+1}]-\EE[\mathbb{V}_2^k]\\
    &\leq-\frac{\alpha}{2}\EE\left[\|\nabla F(\bar x^k)\|_{P_x}^2\right]+\frac{L_f}{L_r}\left(1+\frac{11L_fL_r\alpha}{2}+  L_{rx}(C_{f,2}^2+\tilde\sigma_{f,2}^2)\alpha^2\right)\left(S\beta^2\sigma_{g,1}^2+{\cal O}(\beta \alpha^2T^2S)\right)\\
    &~~+\left[\frac{L_F}{2}+\frac{L_f}{L_r}\left(L_r^2+L_{rx}\right)\left(1+\frac{2\beta}{\mu_g}\right)\right]\alpha^2\tilde\sigma_{f,2}^2+{\cal O}(\alpha^3T^2)+\frac{3\ell_{g,1}^2\alpha}{2}\left((1-\rho\mu_g/2)^NC_u^2+\frac{2\rho\sigma_u^2}{\mu_g}\right)\\
   &~~-\left[\frac{\alpha}{4}-\left(\frac{L_F}{2}+L_fL_r+\frac{L_fL_{rx}}{  L_r}\right)\alpha^2-\frac{\beta\alpha}{2\mu_g}-\left(L_fL_r+\frac{L_fL_{rx}}{  L_r}\right)\frac{2\beta\alpha^2}{\mu_g}\right]\EE\left[\|\bar d_f^k\|_{P_x}^2\right]\\
   &~~-\!\left(\frac{L_f\mu_g\beta}{2L_r}-\frac{11\alpha L_f^2}{2}-\frac{  L_fL_{rx}(C_{f,2}^2+\tilde\sigma_{f,2}^2)\alpha^2}{L_r}\right)\!\EE\left[\|y^k-y^*(\bar x^k)\|^2+\|r^k-r^*(\bar x^k)\|^2\right].\!\!\numberthis\label{minus-stoc-r2}
\end{align*}

As before, we can choose $\beta={\cal O}(\alpha)$ to make the last two terms non-positive. Specifically, a sufficient condition for this is to choose $\beta=c_1\alpha$ and $\alpha\leq\min(\alpha_1,\alpha_2)$ where
\begin{align*}
    &c_1=\frac{11L_fL_r+2  L_{rx}(C_{f,2}^2+\tilde\sigma_{f,2}^2)}{\mu_g}\alpha,~~~~\alpha_1=\frac{1}{c_1\ell_{g,1}}\\
    &\alpha_2=\frac{1}{2L_F+2L_fL_r+\frac{4L_fL_{rx}}{  L_r}+\frac{2c_1}{\mu_g}\left(1+4\left(L_fL_r+\frac{L_fL_{rx}}{  L_r}\right)\right)}.
\end{align*}
Additionally, we set $N={\cal O}(1/\alpha)$ such that $(1-\rho\mu_g/2)^N={\cal O}(\alpha^2)$, then \eqref{minus-stoc-r2} becomes
\begin{align}
    \frac{\alpha}{2}\EE[\|\nabla F(\bar x^k)\|_{P_x}^2]\leq\left(\EE[\mathbb{V}_2^{k}]-\EE[\mathbb{V}_2^{k+1}]\right)+{\cal O}(S\alpha^2\sigma_{g,1}^2+S\alpha^3T^2+\alpha^2\tilde\sigma_{f,2}^2+\rho\alpha\sigma_u^2).\label{51-r2}
\end{align}
Telescoping \eqref{51-r2} and dividing both sides by $\alpha K/2$ yields
\begin{align*}
    \frac{1}{K}\sum_{k=0}^{K-1}\EE\left[\|\nabla F(\bar x^k)\|_{P_x}^2\right]\leq \frac{2(\mathbb{V}_2^{0}-F^*)}{\alpha  K}+{\cal O}(S\alpha\sigma_{g,1}^2+S\alpha^2T^2+\alpha\tilde\sigma_{f,2}^2+\rho\sigma_u^2).
\end{align*}
Choosing $\alpha={\cal O}\left(\frac{1}{\sqrt{K}}\right), S={\cal O}(1), T={\cal O}(K^{1/4})$ and $\rho={\cal O}\left(\frac{1}{\sqrt{K}}\right)$ leads to Corollary \ref{cor-E2}.

\section{Application on Federated Bilevel Learning}
In this section, we present the pseudo-code of {\sf E-AiPOD} and {\sf E2-AiPOD} for federated bilevel learning and compare the metric we used with the state-of-the-art works. 
\subsection{Pseudo-code of {\sf E-AiPOD} and {\sf E2-AiPOD} on federated bilevel learning}
The application of {\sf E2-AiPOD} in the federated bilevel setting is straightforward with the facts in Section \ref{sec:fedbi}. For E-AiPOD, if we plug $V_2$ in \eqref{A-V2} to \eqref{NS-hf}, then we obtain $w^k=[w_1^k,\cdots,w_M^k]$ with 
\begin{align} 
w_m^k=&-\nabla_{xy} g(x^k_m,y^{k+1}_m;\phi_{(0)}^{k})\left[\frac{\tilde cN}{\ell_{g, 1}} \prod_{n=1}^{N^{\prime}}\left(I-\frac{\tilde c}{M\ell_{g, 1}} \sum_{m=1}^M\nabla_{yy}g(x^k_m,y^{k+1}_m;\phi^k_{(n)})\right)\right]\nonumber\\
&\times\left(\frac{1}{M}\sum_{m=1}^M\nabla_y f(x^k_m,y^{k+1}_m;\xi_m^k)\right).\label{wmk}
\end{align}

To avoid transmission of Hessian, which is highly expensive, we can calculate $w_m^k$ by communicating the Hessian-vector product purely like in FedNest \citep{tarzanagh2022fednest}, which is detailed in Algorithm \ref{alg:wk-eff-cal}. We summarize {\sf E-AiPOD} and {\sf E2-AiPOD} on federated bilevel learning together in Algorithm \ref{stoc-alg-fed}.  

\begin{algorithm}[tb]
\small
\caption{\small\blue{\sf E-AiPOD} and \red{\sf E2-AiPOD} in federated bilevel learning: \colorbox{blue!15}{blue part} is run only by \blue{\sf E-AiPOD}; \colorbox{red!15}{red part} is implemented only by \red{\sf E2-AiPOD}; not both at the same time.}
\begin{algorithmic}
\State Initialization: $\{x^0_m,y^0_m\}_{m\in[M]}$, stepsizes $\{\alpha,\beta, \blue{\delta},\red{\rho}\}$, projection probability $\{p,\red{q}\}$,   projection frequency $T$
\For{$k=0$ {\bfseries to} $K-1$}
\State $\{y^{k+1}_m,r^{k+1}_m\}_{m\in[M]}=${\sf E-AiPOD}$_{\rm{low}}^{\rm{fed}}(\{x^k_m,y^k_m,r^k_m\}_{m\in[M]},\beta,p,S)$\\
\colorbox{blue!15}{
\begin{minipage}{0.9\textwidth}
\State call Algorithm \ref{alg:wk-eff-cal} to calculate $\{w_m^k\}_{m\in[M]}$
\For{all workers $m\in[M]$ in parallel}

\State initialize $x_m^{k,0}=x^k_m$
\For{$t=0$ {\bfseries to} $T-1$}
\State update $x^{k,t+1}_m=x^{k,t}_m-\alpha(\nabla_x f_m(x_m^k,y_m^{k+1};\xi^{k,t}_{m})-w^k_m)$
\EndFor
\State set $\Delta_m^k=x_{m}^{k,T}-x_m^{k}$
\EndFor
\State update $x_m^{k+1}=x_m^{k}+\delta\sum_{m=1}^M\Delta_m^k$
\end{minipage}}
\colorbox{red!15}{
        \begin{minipage}{0.9\textwidth}
\State $\{u^{k+1}_m\}_{m\in[M]}=${\sf E2-AiPOD}$_{\rm{med}}^{\rm{fed}}(\{x^k_m,y^{k+1}_m\}_{m\in[M]},\rho,q,N)$
\For{all workers $m\in[M]$ in parallel}
\State update $x^{k+1}_m=x^{k}_m-\alpha(\nabla_x f_m(x_m^k,y_m^{k+1};\xi^{k}_{m})+\nabla_{xy} g_m(x_m^k,y_m^{k+1};\phi^{k}_{m})u^{k+1}_m)$
\EndFor
\If{$K\operatorname{mod} T=0$}
\State update $x^{k+1}_m=\frac{1}{M}\sum_{i=1}^M x^{k+1}_i$
\EndIf
\EndFor
\end{minipage}}
\end{algorithmic}
\label{stoc-alg-fed}
\end{algorithm}

\begin{algorithm}[tb]
\small
\caption{\small{\sf E-AiPOD}$_{\rm{low}}^{\rm{fed}}(\{x^k_m,y^k_m,r^k_m\},\beta,p,S)$: \colorbox{green!15}{green part} denotes the communication round} 
\begin{algorithmic}[1]
\State Initialization:  $\{x^k_m,y^k_m,r^k_m\}_{m\in[M]}$, stepsize $\beta$, skipping probability $p$, 
$y^{k,0}_m=y^{k}_m, r^{k,0}_m=r^{k}_m, ~~\forall m\in[M]$
\For{$s=0$ {\bfseries to} $S-1$}
\For{ all workers $m\in[M]$ in parallel}
\State update $\hat y_m^{k,s+1}=y_m^{k,s}-\beta(\nabla_y g_m(x^{k,s}_m,y^{k,s}_m;\phi^{k,s}_m)-r^{k,s}_m)$
\If{a Bernoulli random variable $\theta^{k,s}=1$}~~~~~~~~~~~~~~~~~~~~\Comment{$\mathbb{P}(\theta^{k,s}=1)=p$}
\State  \colorbox{green!15}{update $y^{k,s+1}_m=\frac{1}{M}\sum_{i=1}^M \left(\hat{y}^{k,s+1}_i-\frac{\beta}{p}r^{k,s}_i\right)$}~~~~~~~~~~~\Comment{Communicate to server and average}
\State  update $r^{k,s+1}_m=r^{k,s}_m+\frac{p}{\beta}(y_m^{k,s+1}-\hat y_m^{k,s+1})$
\Else
\State set $y^{k,s+1}_m=\hat{y}^{k,s+1}_m, ~~~r^{k,s+1}_m=r^{k,s}_m$
\EndIf
\EndFor
\EndFor
\State Outputs: $\{y^{k+1}_m=y^{k,S}_m,r^{k+1}_m=r^{k,S}_m\}_{m\in[M]}$
\end{algorithmic}
\label{stoc-alg-skip-LL-fed}
\end{algorithm}

\begin{algorithm}[htb]
\small
\caption{\small Efficient calculation of $\{w_m^k\}_{m\in[M]}$: \colorbox{green!15}{green part} denotes the communication round}
\begin{algorithmic}[1]
\State Initialization: $\{x^k_m,y^{k+1}_m\}_{m\in[M]}$, constant $\tilde c\leq 1,\ell_{g,1}$, $N$.
\For{ all workers $m\in[M]$ in parallel} 
\State update $v_{m,0}^k=\nabla_y f_m(x_m^k,y_m^{k+1};\xi_{m}^k)$
\EndFor
\State \colorbox{green!15}{update $v_0^k=\frac{\tilde c N}{\ell_{g,1}M}\sum_{m=1}^M v_{m,0}^k$}
\For{$n=1$ {\bfseries to} $N^\prime$} ~~~~~~~~~~~~~~~~~~~~~~~~~~~~~~~~~~~~~~~~~~~~~~~~~~~~~~~~~~~~~\Comment{Draw $N^\prime$ uniformly at random from $\{0,\cdots N-1\}$ }
\For{ all workers $m\in[M]$ in parallel}
\State update $v_{m,n}^k=(I-\frac{\tilde c}{\ell_{g,1}}\nabla_{yy}g_m(x_m^k,y_m^{k+1};\phi_{m,(n)}^k))v_{n-1}^k$
\EndFor
\State \colorbox{green!15}{$v_n^k=\frac{1}{M}\sum_{m=1}^M v_{m,n}^k$}~~~~~~~~~~~~~~~~~~~~~~~~~~~~~~~~~~~~~~~~~~~~~~\Comment{Communicate to server and average}
\EndFor ~~~~~~~~~~~~~~~~~~~~~~~~~~~~~~~~~~~~~~~~~~~~~~~~~~~~~~~~~~~~~~~~~~~~~~~~~~~~\Comment{Set $v^k=v^k_{N^\prime}$}
\For{ all workers $m\in[M]$ in parallel}
\State $w_m^k=-\nabla_{xy}g(x^k_m,y_m^k;\phi_{m,(0)}^k)v^k$. 
\EndFor
\end{algorithmic}
\label{alg:wk-eff-cal}
\end{algorithm}

\begin{algorithm}[tb]
\small
\caption{\small{\sf E2-AiPOD}$_{\rm{med}}^{\rm{fed}}(\{x^k_m,y^{k+1}_m\},\rho,q,N)$: \colorbox{green!15}{green part} denotes the communication round} 
\begin{algorithmic}[1]
\State Initialization: $\{x^k_m,y^{k+1}_m\}_{m\in[M]}$, stepsize $\rho$, skipping probability $q$, 
$u^{k,0}_m=e^{k,0}_m=0,~~\forall m\in[M]$
\For{$n=0$ {\bfseries to} $N-1$}
\For{ all workers $m\in[M]$ in parallel}
\State update $\hat u_m^{k,n+1}=u^{k,n}_m-\rho(\nabla_y f(x^k_m,y^{k+1}_m;\xi^{k,m}_{(n)})+\nabla_{yy}g(x^k_m,y^{k+1}_m;\phi^{k,m}_{(n)})u^{k,n}_m-e^{k,n}_m)$
\If{a Bernoulli random variable $\tilde\theta^{k,n}_m=1$}~~~~~~~~~~~~~~~~~~~~~~~~~~~~~~~~~~~~~~~~~~~~~~~~~~~~~~\Comment{$\mathbb{P}(\tilde\theta_m^{k,n}=1)=q$}
\State \colorbox{green!15}{update $u^{k,n+1}_m=\frac{1}{M}\sum_{i=1}^M \left(\hat{u}^{k,n+1}_i-\frac{\rho}{q}e^{k,n}_i\right)$}~~~~~~~~~~~~~~\Comment{Communicate to server and average}
\State  update $e_m^{k,n+1}=e_m^{k,n}+\frac{q}{\rho}(u_m^{k,n+1}-\hat u_m^{k,n+1})$
\Else
\State set $u^{k,n+1}_m=\hat{u}^{k,n+1}_m, ~~~e^{k,n+1}_m=e^{k,n}_m$
\EndIf
\EndFor
\EndFor ~~~~~~~~~~~~~~~~~~~~~~~~~~~~~~~~~~~~~~~~~~~~~~~~~~~~~~~~~~~~~~~~~~~~~~~~~~~~~~~~~~~~~~~~~~~~~~~~\Comment{Output $\{u^{k+1}_m=u^{k,N}_m\}_{m\in[M]}$}
\end{algorithmic}
\end{algorithm}

\subsection{Equivalence between our metric with metric in federated bilevel learning}\label{measure-fed}
In this section, we prove the equivalence of our measure in \eqref{stationary} with the measure of FedNest in non-consensus federated bilevel setting. 

According to Lemma \ref{lm:proj_closed}, and $\mathcal{X}=\{x\mid Bx=0\}$ in federated bilevel setting, we know
\begin{align}\label{proj_Bx}
    (I-B^\dagger B)\nabla F(x)=\operatorname{Proj}_{\mathcal{X}}(\nabla F(x)).
\end{align}
Then according to \eqref{proj_x_fed} and \eqref{upp-grad}, we obtain
\begin{align}
    (I-B^\dagger B)&\nabla F(x)=\frac{1}{M}\sum_{m=1}^M\nabla_{x_m} f_m(x_m,y_m^*(x_m))+\left(\frac{1}{M}\sum_{m=1}^M\nabla_{x_my_m}g_m(x_m,y_m^*(x_m))\right)\nonumber\\
    &\times\left(\frac{1}{M}\sum_{m=1}^M\nabla_{y_my_m}g_m(x_m,y_m^*(x_m))\right)^{-1}\left(\frac{1}{M}\sum_{m=1}^M\nabla_{y_m}f_m(x_m,y_m^*(x_m))\right).\label{upp-grad-proj}
\end{align}

On the other hand, the gradient of the objective in FedNest is 
\begin{align} 
    \frac{1}{M}\sum_{m=1}^M\nabla_{x} f_m(x,y^*(x))&+\left(\frac{1}{M}\sum_{m=1}^M\nabla_{xy}g_m(x,y^*(x))\right)\nonumber\\
    &\times\left(\frac{1}{M}\sum_{m=1}^M\nabla_{yy}g_m(x,y^*(x))\right)^{-1}\left(\frac{1}{M}\sum_{m=1}^M\nabla_{y_m}f_m(x,y^*(x))\right).\label{upp-grad-fednest}
\end{align}

We find that \eqref{upp-grad-fednest} is the same as \eqref{upp-grad-proj}, if replacing $g_m(x,y^*(x))$ and $f_m(x,y^*(x))$ by $g_m(x_m,y_m^*(x_m))$ and $f_m(x_m,y_m^*(x_m))$. Moreover, since $\EE[\|\nabla F(x)\|^2_{P_x}]=\EE[\|(I-B^\dagger B)\nabla F(x)\|^2]$, the measure $\EE[\|\nabla F(x)\|^2_{P_x}]$ in our analysis coincides with the gradient norm measure in FedNest \citep{tarzanagh2022fednest}.

\section{Additional Details of Experiments}
We will report the detailed settings of the experiments in Section \ref{sec:experiment}.
In the federated bilevel learning experiments, the number of workers is set as $M=50$ and each local network is a 2-layer multilayer perceptron with hidden dimension $200$. 
The hyper-parameters are found by measuring both the convergence speed and the stability of the algorithm via a grid search.

\paragraph{Synthetic task.} {\sf E-AiPOD}: The projection probability is set as $p=0.3$ in the right figures, the  total number of  iterations is  $K=400/p$, the number of UL iterations is $T=2$ in the left figures, the number of LL iterations is $S=5$, the step sizes are set as $\alpha=0.02,\beta=0.01$, the noise has mean $0$ and std $0.1$. {\sf AiPOD} is a special case of {\sf E-AiPOD} with $p=1$.


\paragraph{Federated hyper-representation learning.} {\sf E-AiPOD}: The communication probability is set as $p=0.1$ in Figure \ref{fig:name2} (right), $S=20$, $\alpha=0.01,\beta=0.05$, Neumann iteraion $N'=5$, and the batch size is $256$. FedNest (notations in \cite{tarzanagh2022fednest}): Choose LL iteration number $\tau=10$ and episode $T=1$ so that the communication frequency is $0.1$ per LL iteration, which is the same as the choice of $p$ for {\sf E-AiPOD}. The UL iteration numbers are specified in Figure \ref{fig:name2}, and we set $\alpha=0.01$, $\beta=0.02$ (under $T=1$) or $\beta=0.01$ (under $T=5$), $N'=5$ and batch size as $256$.

\paragraph{Federated learning from imbalanced data-set.} {\sf E-AiPOD}: Communication probability $p=0.3$ in Figure \ref{fig:name2} (right), $S=20$, $\alpha=0.01,\beta=0.04$, $N'=3$, and batch size $256$. FedNest: Choose LL iteration number $\tau=3$ and episode $T=3$ and thus the communication frequency is $0.3$ per LL iteration, which is the same as $p=0.3$. The UL iteration numbers are specified in Figure \ref{fig:name2}. Set $\alpha=0.01$, $\beta=0.02$, $N'=3$, and batch size as $256$.

\end{document}